%% file: Main.tex
\documentclass[11pt,a4paper,oneside,openany]{report}

\usepackage{phd_thesis} 
\usepackage{emptypage} 
\usepackage{afterpage}
\usepackage[pdftex]{graphicx}
\usepackage{amsmath}
\usepackage{paralist}
\usepackage{amssymb}
\usepackage[dvipsnames]{xcolor}\usepackage{booktabs}
\usepackage{fancynum}
\usepackage{subcaption}
\usepackage{setspace}
\usepackage[hang,flushmargin]{footmisc} 
\usepackage{url}
\usepackage{multirow}
\usepackage{makecell}
\usepackage{threeparttable}
\usepackage{rotating}
\usepackage{wrapfig}
\usepackage{xcolor}
\usepackage[ruled]{algorithm2e}
\usepackage{hyperref}\hypersetup{colorlinks=false,linkcolor=blue,urlcolor=blue,citecolor=red} 
\usepackage{cleveref}
\usepackage{enumitem}

\usepackage[labelfont=bf, textfont=bf]{caption}

\usepackage{amsthm}

\usepackage{mathtools}

\usepackage{multicol}
\usepackage{array}
\newcommand{\PreserveBackslash}[1]{\let\temp=\\#1\let\\=\temp}
\newcolumntype{C}[1]{>{\PreserveBackslash\centering}p{#1}}
\newcolumntype{R}[1]{>{\PreserveBackslash\raggedleft}p{#1}}
\newcolumntype{L}[1]{>{\PreserveBackslash\raggedright}p{#1}}
\newcolumntype{?}{!{\vrule width 0.6pt}}

\usepackage{dsfont}
\usepackage{pifont}
\newcommand{\white}[1]{{\color{white}{#1}}}

\theoremstyle{definition}

\theoremstyle{definition}

\usepackage{etoolbox}
\preto\tabular{\setcounter{magicrownumbers}{0}}
\newcounter{magicrownumbers}

\newcommand{\Lapl}{\mathbf{\mathop{\mathcal{L}}}}
\newcommand{\Trans}[1]{{#1}^{\top}}

\newcommand{\Mat}[1]{\mathbf{#1}}

\newcommand{\Space}[1]{\mathbb{#1}}
\newcommand{\Set}[1]{\mathcal{#1}}

\newcommand{\ie}{\emph{i.e.}}

\newcommand{\eg}{\emph{e.g.}}

\newcommand{\st}{\emph{s.t. }}
\newcommand{\etc}{\emph{etc.}}

\newcommand{\cf}{\emph{cf. }}
\newcommand{\aka}{\emph{aka. }}

\newcommand{\red}[1]{{\color{red}{#1}}}
\newcommand{\green}[1]{{\color{OliveGreen}{#1}}}

\newcommand\blankpage{\null\thispagestyle{empty}\addtocounter{page}{-1}\newpage}

\begin{document}
	\pagenumbering{roman}
	\raggedbottom 
	\input{010_Title.tex}
	\input{020_Declaration}
	\input{030_Acknowledgements}
	\tocpage
	
	\input{040_Abstract}
	\tablepage
	
	\figurepage
	\newpage
	
	\input{050_Publications}

	\afterpage{\blankpage}
	
	\newpage
	\pagenumbering{arabic}
	\setcounter{page}{1}
	
	\include{1_Introduction}
	\include{2_Related}

	\include{work1/0_main}
	\include{work2/main}

	\include{work3/main}
	\include{work4/main}
	\include{conclusion}
	\addcontentsline{toc}{chapter}{Bibliography}
	\bibliographystyle{abbrv}
	\small \bibliography{Main}
	
	\newpage
	\appendix
	\input{work2/sec/X_supplementary}

	

\end{document}

%% file: 010_Title.tex
\renewcommand{\thesisdate}{15}
\renewcommand{\thesisyear}{2023}
\renewcommand{\thesismonth}{Nov}
\renewcommand{\thesistitle}{Causality Model for Semantic Understanding on Videos}
\renewcommand{\thesisauthor}{YICONG LI}
\renewcommand{\thesisauthoredu}{(M. S. Eng., Columbia University)}
\renewcommand{\universityname}{NATIONAL UNIVERSITY OF SINGAPORE}
\renewcommand{\thesisschool}{Integrative Sciences and Engineering Programme}
\renewcommand{\thesissupervisor}{Professor Tat-Seng Chua}
\thesistitlepage

%% file: 020_Declaration.tex
\declarationpage

%% file: 030_Acknowledgements.tex
\acknowledgementspage{
As moments swiftly pass, the days meld into years. Looking back on this incredible journey, I envision myself standing at the entrance of NExT, pondering when to knock. Arriving at the lab with humble knowledge, I have been immensely fortunate to be surrounded by empathetic and insightful individuals who have supported, guided, and inspired me throughout my pursuit of a Ph.D. Their collective wisdom has been pivotal in molding my academic and personal growth during this transformative period.

First and foremost, I extend my deepest gratitude to my supervisor, Prof. Chua, for his outstanding guidance. He has not only been an exceptional mentor in academia but has also taught me the true meaning of integrity, dedication, and compassion.

I also express my appreciation to senior researchers, Dr. Wang Xiang, Yang Xun, and Junbin, for their meticulous guidance on my paper and for exemplifying the quintessence of excellence as researchers. Their expertise, commitment, and passion for their field have been genuinely inspiring, and I am privileged to have had the opportunity to learn from them.

I wish to thank my basketball mates, who have provided a much-needed respite from the rigors of academic life. Your sportsmanship and friendship have been a constant source of joy and motivation throughout this journey.

As I reflect, I am overwhelmed with gratitude for my family's steadfast love and support. I offer my sincerest appreciation for the solid foundation they have built for me, nurturing my dreams with their selfless sacrifices, priceless guidance, and unconditional love. They have instilled in me the values and beliefs that have shaped who I am today.

Though my words may be unrefined and simple, I will forever treasure the time spent alongside my mentors, colleagues, and friends. To my dear Profs and friends, I wish you all the happiness and success life has to offer. May our paths intertwine again, and may we continue to support and elevate each other in our respective journeys.
}

%% file: 040_Abstract.tex
\abstractpage{

After a decade of prosperity, the development of video understanding has reached a critical juncture, where the sole reliance on massive data and complex architectures is no longer a one-size-fits-all solution to all situations. 
The presence of ubiquitous data imbalance hampers DNNs from effectively learning the underlying causal mechanisms, leading to significant performance drops when encountering distribution shifts, such as long-tail imbalances and perturbed imbalances. This realization has prompted researchers to seek alternative methodologies to capture causal patterns in video data.
To tackle these challenges and increase the robustness of DNNs, causal modeling emerged as a principle to discover the true causal patterns behind the observed correlations.
This thesis focuses on the domain of semantic video understanding and explores the potential of causal modeling to advance two fundamental tasks: Video Relation Detection (VidVRD) and Video Question Answering (VideoQA).

In summary, the major contributions of this thesis are as follows:
\begin{itemize}[leftmargin=*]
\item \noindent \textbf{We propose an interventional video relation detection method}, named IVRD, to address the long-tail imbalance of relation in VidVRD, where tail relations, despite being informative, are difficult to predict due to their scarcity in the dataset. Specifically, we form a set of relation prototypes in a hierarchical manner, which forces the relation reasoning module to focus on the visual content of dynamic interactions between entities rather than relying on spurious correlations between objects and relation labels. By incorporating causal reasoning, IVRD offers a promising direction for improving video understanding in the presence of long-tail imbalances, enabling models to better generalize to real-world scenarios where rare or infrequent relations may play a crucial role in the overall understanding of the scene.

\item \noindent \textbf{We introduce the invariant grounding for VideoQA}, dubbed as IGV, a model-agnostic learning framework that addresses the negative effect induced by the spurious correlations in the answer-environment. In essence, IGV discovers the causal reasoning pattern by grounding the question-critical (causal) scene. Specifically, it leverages the fact that the relations between causal scenes and answers are invariant regardless of changes in the environment, and the removal of causal scenes should cause failure in answering the question. By grounding these critical scenes, IGV compels the VideoQA models to focus on the essential visual content for accurate reasoning, while shielding them from the negative influence of the environment, thus significantly improving the reasoning ability of the backbone models.


\item \noindent \textbf{We introduce the equivariant grounding for VideoQA}, EIGV, to further advance the robustness and visual explainability in a model-agnostic manner.  Built on top of IGV, EIGV additionally incorporates equivariance, which encourages the answering process to be sensitive to the semantic changes in the causal scene and question. In comparison, invariant grounding enforces the answering to be insensitive to changes in the environmental scene.
As a result, these two regularizations work collaboratively to distinguish the causal scene from the environment while providing more transparency by presenting the visual-linguistic alignment. By combining the strengths of both invariant and equivariant grounding, EIGV creates a more robust and explainable framework for VideoQA.

\item \noindent \textbf{We discover spatio-Temporal rationales for VideoQA}, which solve the low accuracy on samples with long video and multiple objects (\ie complex VideoQA). 
Since current VideoQA practices (including pretrained models, \eg SeVila \cite{sevila}) are mostly trained with short video clips ($\sim$15s) with few entities ($\sim$2), they tend to suffer from poor transferability to complex video (over 80s and 5 objects). The reason behind is that long videos inevitably introduce vast redundancy and spurious correlations due to the presence of numerous question-irrelevant environmental objects. 
Confronting this challenge, we first highlight the importance of modeling question-critical temporal moments and spatial objects, and then introduce \textbf{S}patio-\textbf{T}emporal \textbf{R}ationalization (STR), which utilizes a differentiable selection module to adaptively gather question-critical moments and objects through cross-modal interaction. Coupled with a more reasonable candidate answer decoding strategy, STR effectively identifies question-irrelevant frames and objects as causal patterns, leading to improved predictions, especially in complex scenarios.

\end{itemize}
A limitation of this thesis pertains to the evaluation of the identified causal scenes. Throughout the research, we have relied on the overall Question Answering (QA) performance as an indirect indicator of the discovered causal scene's quality, which is based on the rationale that a more accurate localization of the causal scene can potentially yield richer question-relation visual cues, subsequently enhancing QA performance. However, it is essential to acknowledge that a direct quantitative measure rooted in the causal scene would offer more compelling insights. Regrettably, due to the absence of human-level grounding annotations, such a measurement remains absent in the current work. Therefore, future endeavors will focus on establishing an evaluation benchmark specifically tailored to the causal scene, involving human annotations on the visual elements that underpin the answering process. This initiative will contribute to a more comprehensive and rigorous assessment of causal scene discovery.

In summary, our contributions extend the frontiers of causal modeling in semantic video understanding, empowering AI systems to grasp causal patterns and improve performance on challenging video understanding tasks.
}

%% file: 050_Publications.tex
\publicationpage{

\begin{itemize}

    \item 
    \textbf{Yicong Li}, Xun Yang, Xindi Shang, Tat-Seng Chua. 
    \textit{Interventional Video Relation Detection}, ACM MM, 2021. \cite{ivrd}
 
    \item 
    \textbf{Yicong Li}, Xiang Wang, Junbin Xiao, Wei Ji, Tat-Seng Chua,
    \textit{Invariant Grounding for Video Question Answering},  CVPR, 2022. (Oral, Best Paper Finalist) \cite{IGV}

    \item 
    \textbf{Yicong Li}, Xiang Wang, Junbin Xiao, Tat-Seng Chua.
    \textit{Equivariant and Invariant Grounding for Interpretable Video Question Answering},  ACM MM,  2022. 
    \cite{EIGV}
	
    \item 
    \textbf{Yicong Li}, Xiang Wang, Junbin Xiao, Wei Ji, Tat-Seng Chua.
    \textit{Transformer-Empowered Invariant Grounding for Video Question Answering}, TPAMI, 2023. \cite{li2023transformer}
	
    \item 
    \textbf{Yicong Li}, Junbin Xiao, Chun Feng, Xiang Wang.
    \textit{Discovering Spatio-Temporal Rationales for Video Question Answering}, Tat-Seng Chua, ICCV, 2023. \cite{li2023discovering}
     
    \item  \textbf{Yicong Li}, Xun Yang, An Zhang, Chun Feng, Xiang Wang, Tat-Seng Chua. \textit{Redundancy-aware Transformer for Video Question Answering}, ACM MM, 2023. \cite{li2023redundancy}

     \item  \textbf{Yicong Li}, Na Zhao, Junbin Xiao, Chun Feng, Xiang Wang, Tat-Seng Chua. \textit{LASO: Language-guided Affordance Segmentation on 3D Object}, CVPR, 2024. \cite{li2024laso}
\end{itemize}

}

%% file: 1_Introduction.tex
\chapter{Introduction}\label{sec:intro}
\section{Motivation}
Videos serve as a direct and compelling medium for capturing and portraying our physical world. Their ability to record dynamic scenes presents a treasure trove of visual information that captivates AI researchers to construct human-level intelligent systems on videos. 
As the field advances, the focus of video understanding has deepened. While earlier efforts primarily centered on attribute-level recognition, such as identifying visual objects and actions within videos \cite{he2016deep,ren2015faster,tran2015learning,simonyan2014two}, recent studies have taken a significant step towards more challenging semantic-level understanding by eliciting natural language utterances from videos, enabling a deeper comprehension of the visual content \cite{zeng2017leveraging,jang2017,seo2021attend}.
Among them, two tasks stand out prominently, 1) Video Relation Detection (VidVRD) \cite{shang2017video}, which uncovers the intricate relationships between objects in a video, and 2) Video Question Answering (VideoQA) \cite{next-qa} which goes beyond object relation and focuses on overall comprehension of the video by answering questions.

\begin{figure}[t]
    \centering
    	\subcaptionbox{Long-tail imbalance in VidVRD. Due to object-relation shortcut, once the  ``person'' and  ``elephant'' are identified as subject and object, the model will blindly yield ``stand{\_}right'' as the relation without considering rare but informative ``feed''.\label{fig:1a}}{
		\includegraphics[width=0.48\textwidth]{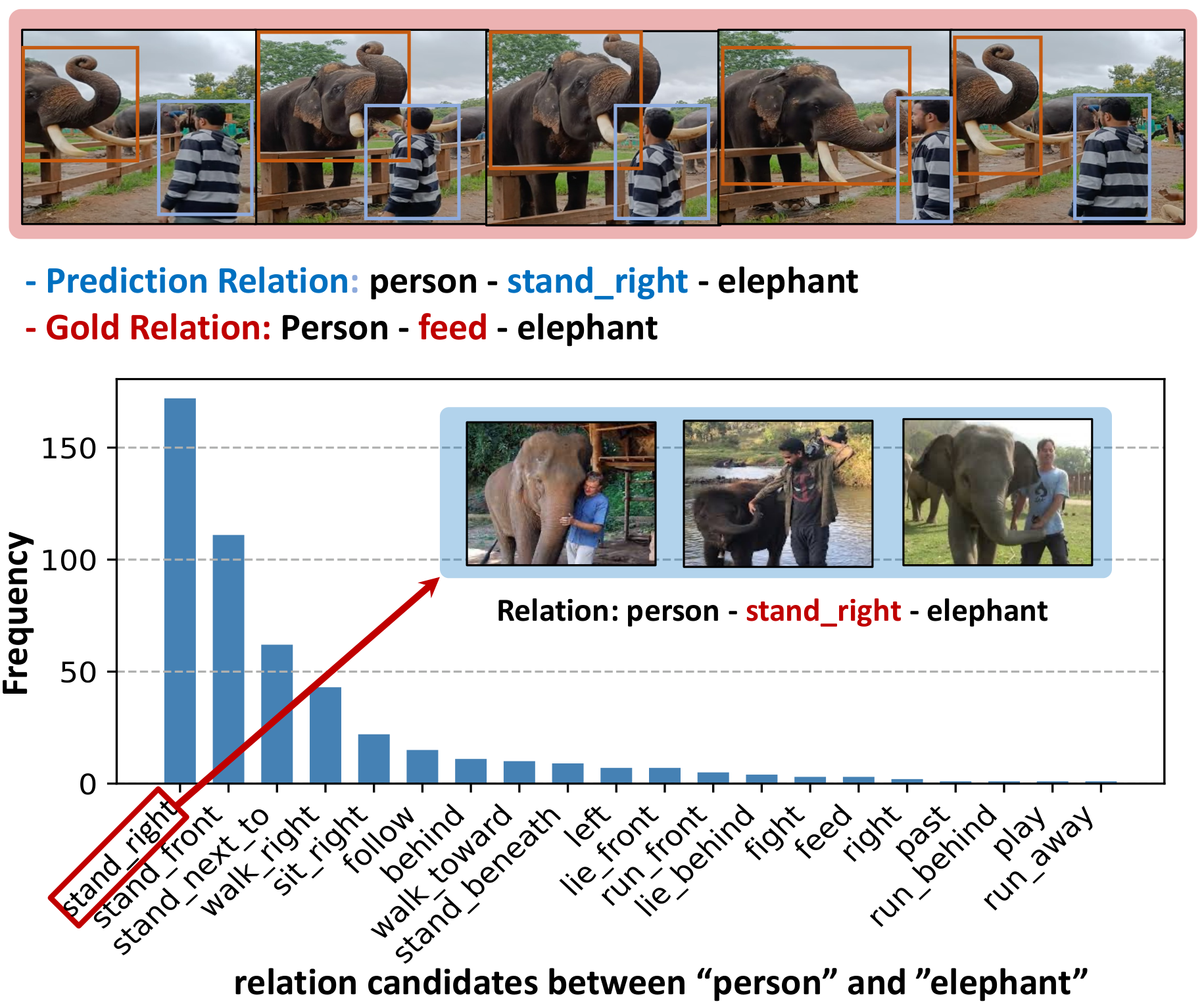}}
        \hspace{1pt}
		\subcaptionbox{Perturbed distributions in VideoQA. Due to the environment-answer shortcut, once the model detects the presence of ``track'' scene, it will directly predict ``race'' as an answer, without noticing the causal clue of ``jump'' in the last three frames. \label{fig:1b}}{
	    \includegraphics[width=0.48\textwidth]{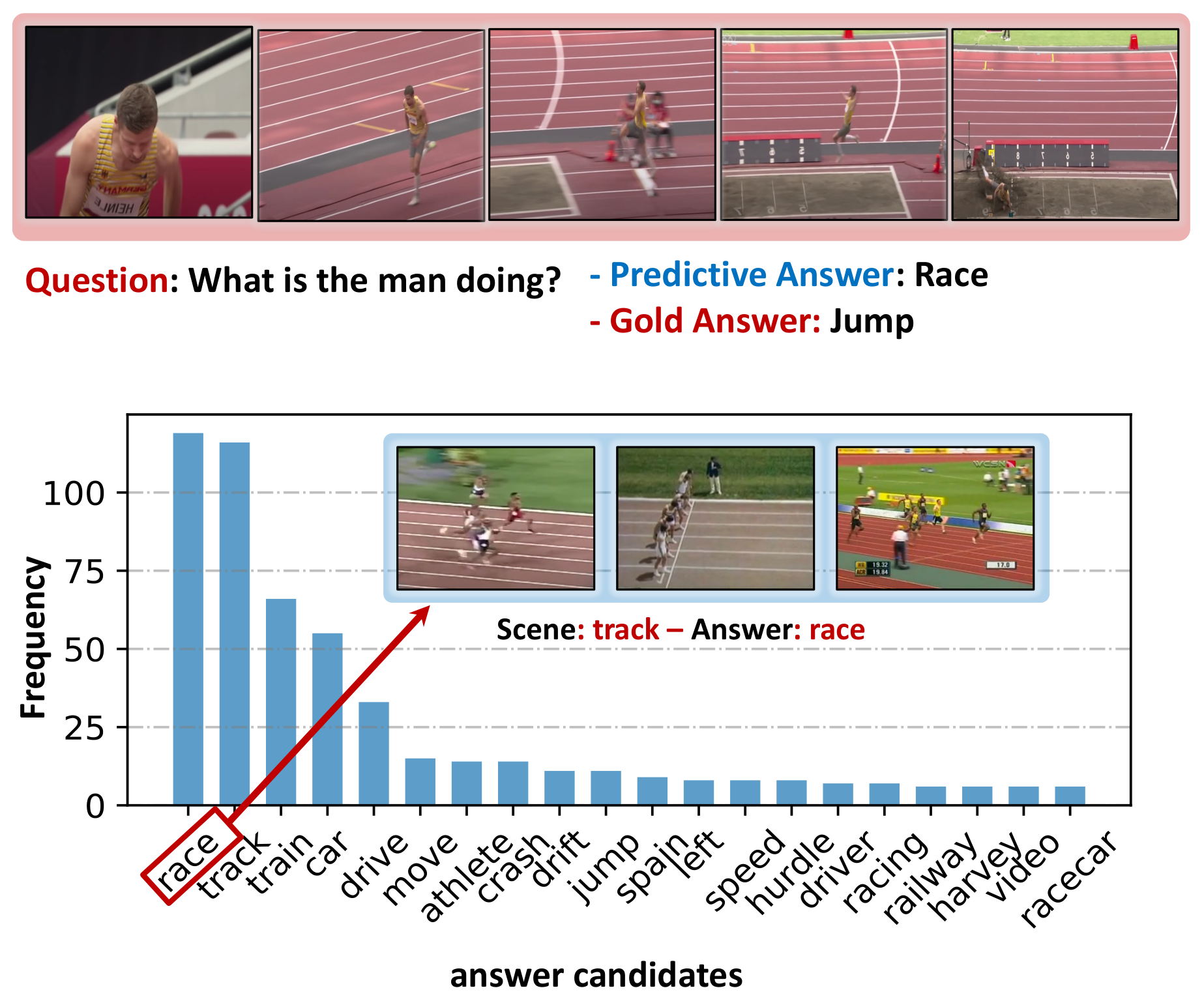}}
    \vspace{-8pt}
    \caption{Example of how rare cases in semantic video understanding tasks are distracted by (a) object-relation correlation, and (b) environment-answer correlation.}
    \vspace{-15pt}
\end{figure}

While video understanding at the semantic level has continually caught attention in the research field, \cite{shang2019relation,zeng2017leveraging}, we notice that most of the research still recklessly follows the ``learning from association" pattern \cite{pearl2009causal, pearl2000causality}, where the enlarged model capacity is fed with massive data to ``memorize" the sample distribution without understanding the underlying causality. Unfortunately, due to DNN's inherent tendency to learn spurious correlations as shortcuts, they inevitably end up with models that are fragile when facing data imbalance \cite{DBLP:conf/cvpr/0001AB19}, such as long-tail imbalance (\eg, in \cref{fig:1a} relation ``man-stand{\_}right-elephant" are much more frequently exposed than ``man-feed-elephant") or perturbed distributions (\eg, in \cref{fig:1b} ``track'' are mostly recorded with ``race'' than ``jump'' ). This seriously hinders their robustness and deployment to real-world applications. 


In this regard, we argue the importance of identifying the causal pattern from input video in semantic understanding, whose relation to ground truth remains stable even in the low-showed cases.

\section{Challenges}


DNNs have been observed to adopt unintended ``shortcut" strategies \cite{DBLP:journals/natmi/GeirhosJMZBBW20}, since the superficial patterns, are learned first via the current ERM training scheme.  As a result, DNNs tend to form strong associations with these superficial patterns as a shortcut, making the true causal patterns hard to discover.
In the case of video semantic understanding, the problem becomes more challenging due to the following reasons:

\begin{itemize}
    \item \noindent \textbf{Superficial Patterns at Different Levels.} Videos inherently exhibit a multi-level structure, \cite{hcrn, hqga}, where a video consists of a sequence of frames, and each frame contains multiple objects.  As a result, superficial patterns can manifest at different levels within this hierarchy.
    At the frame level, a superficial pattern can establish on the scene depicted in certain frames. For example, in \cref{fig:1b}, the spurious correlation could be attributed to the scene shown in the second frame, which contains only the "track" scene. 
    Similarly, at the object level, a superficial pattern can emerge when the presence of a single object. For instance, in the last three frames of \cref{fig:1b}, the ``track" object's presence might dominate the prediction, drawing attention away from the other essential clues (\eg, jumping pit and athlete) that best support the gold answer ``jump''.

    \item \noindent \textbf{Challenging Semantic Concept.} Due to the expressive nature of the video content, a semantic concept can manifest in various forms, making their causal patterns hard to capture. This is particularly challenging for the less frequently occurring ``tail" cases. For example, in VidVRD, the relation concept ``feed'' can be found in the form of `` man-feed-monkey'', ``children-feed-dog'', \etc. As these instances are rarely shown and their visual expressions are far from similar, capturing the causal pattern of ``feed'' is thus difficult.
    Likewise, in VideoQA, understanding the causal pattern of intricate semantic concepts such as ``because of" is essential to answer ``why" questions. However, comprehending such semantic concepts is difficult, as they are not explicitly stated in the video.
\end{itemize}

Addressing these challenges necessitates the development of robust causal models specifically tailored for video semantic understanding.

\section{Proposals}

The ability to naturally grasp causal patterns from videos is intrinsic to humans, yet emulating this level of causal understanding in AI presents significant challenges. In this thesis, our focus lies in causal modeling for semantic video understanding, with a particular emphasis on addressing two data imbalance scenarios: the long-tail imbalance in Video Relation Detection (VidVRD) and the perturbed imbalance in Video Question Answering (VideoQA).
These two tasks stand out prominently in semantic video understanding, as the VidVRD is the first step that goes beyond object recognition to their pair-wise semantic relation over time.
VideoQA, on the other hand, extends the temporal relationships to a more comprehensive semantic understanding by integrating nuances of video content and contextual cues.
In VidVRD, we target the long-tail imbalance of relation, where the causal pattern of a tail relation can be overlooked due to a strong spurious correlation between object class head relations.
In VideoQA, we focus on the perturbed imbalance of the environment scene, where a question-irrelevant frame or object can disrupt predictions due to their spurious correlation with the answer. In this thesis, we first approach this issue from a model-agnostic angle and develop two learning schemes to discover the causal frame that fully contains the answer clue. Built on this, we introduce two model-specific methods that advance causal modeling by improving its efficiency and providing fine-grained solutions at the object level. Our specific proposals are as follows.

\begin{figure*}
    \centering
    \includegraphics[width=1.0\textwidth]{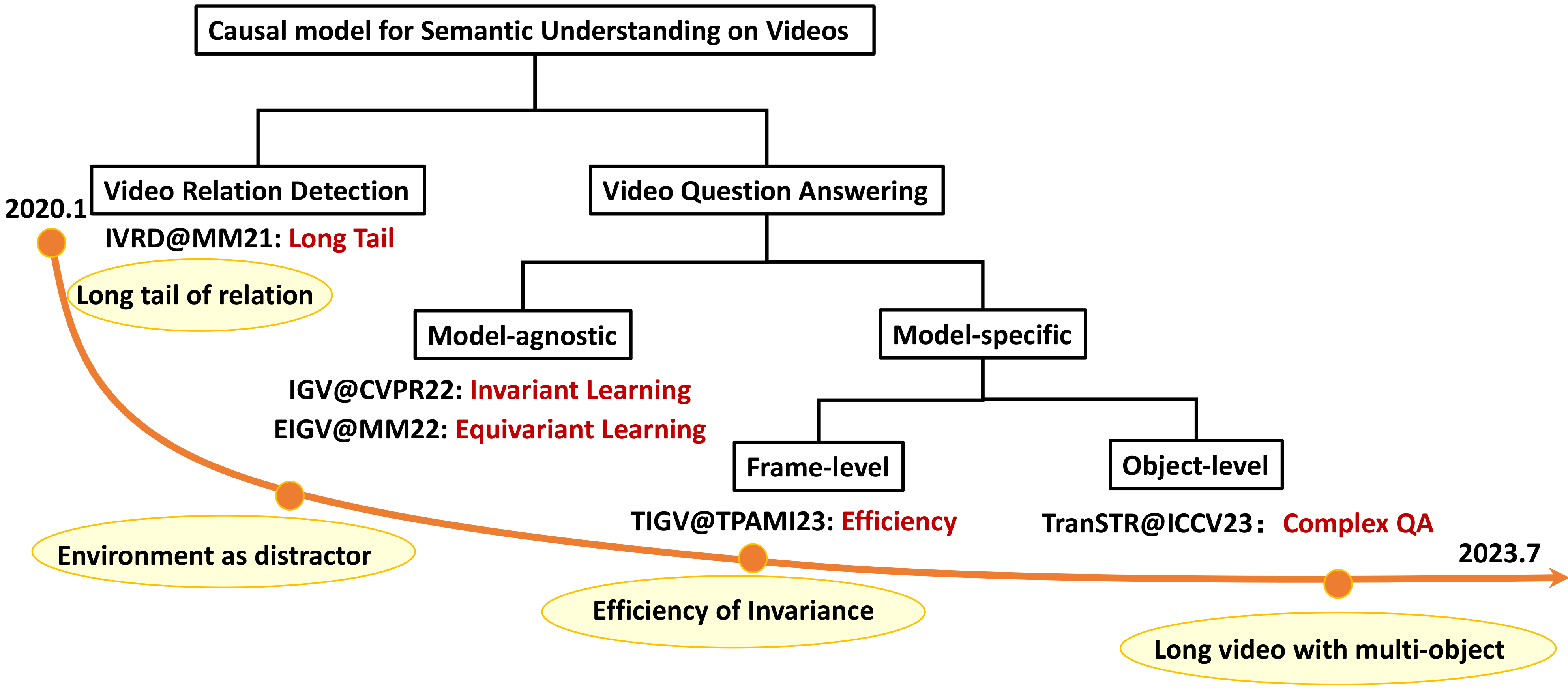}
    \caption{The research tree for towards causal model for Semantic Understanding on Videos. In VidVRD, we study the long tail issue in relation detection that discovers more rare but informative ``tail'' relations. In VideoQA, we first address frame-level shallow patterns induced by answer-environment spurious correlation and introduce two model-agnostic learning schemes: invariant grounding and equivariant grounding. Following this line, we extend a more efficient invariant grounding by instantiating a transformer-based design. Finally, we target object-level shallow patterns and manage a performance boost on complex VideoQA.}
    \label{fig:tree}
\end{figure*}

\paragraph{Interventional Video Relation Detection}
Our journey begins with Video Relation Detection (VidVRD), a crucial task that involves analyzing the interactions and dependencies among objects over a temporal sequence within videos. This task plays a pivotal role in enabling AI systems to comprehend the complex visual scenes and infer meaningful connections between different elements in videos.
However, relations in existing benchmark \cite{shang2021video, shang2019relation} follow a long-tail distribution, in which the majority of them are positional relation that conveys limited visual infomation, \eg ``right'', ``left'' and ``next-to'', \etc, whereas the more informative relations such as ``feed'', ``chase'' and ``play'' are rarely depicted. Under the current training scheme, models will inevitably capture a spurious correlation between the object and head relations. As a result, they might blindly predict positional relations like ``stand-right" once it detects ``person" and ``elephant" as subjects and objects, without considering the visual details of the dynamic interaction.
To overcome this issue, we introduce an \textbf{I}nterventional \textbf{V}ideo \textbf{R}elation \textbf{D}etection (IVRD) approach that force the model to fairly consider all relations, and make the final prediction based on the visual content of the dynamic interaction between the entities, rather than the spurious correlations between the objects and relation labels.

\paragraph{Invariant Grounding for Video Question Answering} 
Moving forward, we shift our focus to Video Question Answering (VideoQA), a task that poses a challenge in comprehending semantic concepts learned from videos. In VideoQA, the presence of certain environment scenes, although irrelevant to the question, can perturb the model's prediction through environment-answer correlations. As a pioneer to address this issue, we adopt a causal perspective toward VideoQA and propose a novel model-agnostic learning framework called Invariant Grounding for VideoQA (IGV). The IGV framework aims to discover the causal reasoning pattern by grounding the question-critical scene, whose causal relations with the answers remain invariant across different perturbations. By employing IGV, we compel VideoQA models to shield the answering process from the negative influence of spurious correlations. Consequently, the reasoning ability of the backbone models significantly improved, as they become more adept at identifying and focusing on the genuine causal connections between questions and answers in videos.

\paragraph{Equivariant and Invariant Grounding for Video Question Answering}
In addition to the advancement of robustness, grounding the causal scene in VideoQA also brings about the inherent benefit of visual explainability. Following this line, we enhance the invariant grounding scheme by integrating equivariant learning. Specifically, the equivariant grounding encourages the answering to be sensitive to the semantic changes in the causal scene and question. While the invariant grounding enforces the answering to be insensitive to the changes in the environment scene.
By incorporating both equivariant and invariant grounding into our model-agnostic learning framework EIGV (\textbf{E}quivariant and \textbf{I}nvariant \textbf{G}rounding for \textbf{V}ideoQA), we enable the model to effectively distinguish the causal scene from the environment information. This explicit distinction allows the model to explicitly present visual-linguistic alignment, offering improved interpretability and understanding of the reasoning process in VideoQA. As a result, EIGV not only enhances robustness but also provides meaningful visual explanations for the AI model's answers, contributing to more transparent and trustworthy AI systems in semantic video understanding.


\paragraph{Discovering Spatio-Temporal Rationales for Video Question Answering}
Despite our exploration of several frame-level causal modeling approaches, an object-level casual modeling method that not only identifies the causal frame, but also locates causal object is till-now lacking. In this work, we strive to solve complex VideoQA, where the performance of a well-trained model can be severely deteriorated on long video with multiple objects, due to the redundancy and spurious correlation induced by the large number of environmental objects.  
To tackle the challenge, we highlight the significance of identifying question-critical temporal moments and spatial objects within the video content. To achieve this, we propose a novel approach called \textbf{S}patio-\textbf{T}emporal \textbf{R}ationalization (STR), which utilizes a differentiable selection module to adaptively gather question-critical moments and objects through cross-modal interaction. Through careful verification, we demonstrate that STR is able to effectively identify the question-critical frames and objects as causal patterns, leading to improved predictions, especially in complex scenarios.

In summary, our research focuses on causal modeling for semantic video understanding, encompassing two main research directions: 1) resolving the long-tail imbalance in Video Relation Detection (VidVRD); and 2) mitigating the perturbed imbalance induced by the environmental scene in Video Question Answering (VideoQA). As we progress toward more advanced semantic understanding, we initially address the first direction by introducing a causality-inspired intervention on objects. This intervention compels the model to fairly incorporate each possible relation, enhancing its ability to detect and understand rare relations in videos. For the second research direction, we develop a series of methodologies, both model-agnostic and model-specific, to effectively mitigate the influence of the environment scene at both the frame and object levels. Our contributions pave the way for more robust and interpretable semantic video understanding, fostering progress in the field of AI technology for video analysis and comprehension.

\section{Outline}
\cref{fig:tree} provides an overview of our research tree, highlighting the key components of our work. To enhance the clarity and comprehension of the presented works, we structure the remainder of this thesis as follows:
\textbf{\cref{cha:background}} reviews the extensive body of literature related to causal modeling and semantic video understanding. This chapter serves as the foundation for understanding the background and context of our research.
\textbf{\cref{cha:ivrd}} presents our work on Video Relation Detection (VidVRD), where we address the issue of long-tail imbalance, aiming to improve the prediction of rare but informative relations.
Subsequently, three chapters delve into various solutions for handling perturbed imbalances in VideoQA, all centered around discovering causal patterns from input videos:
\textbf{\cref{cha:igv}} presents Invariant Grounding for VideoQA (IGV), a model-agnostic learning scheme designed to discover causal patterns at the frame level. The goal of IGV is to address the issue of spurious correlations induced by environment frames and improve the reasoning ability of VideoQA models. 
%
\textbf{\cref{cha:eigv}} explores the incorporation of equivariance on top of invariant learning, and introduces Equivariant and Invariant Grounding for VideoQA (EIGV). This combination yields improved causal patterns and visual interpretability in VideoQA.
\textbf{\cref{cha:transtr}} analyzes the necessity and challenges of complex VideoQA scenarios, emphasizing the importance of discovering object-level causal patterns for long videos with multiple objects.
\textbf{\cref{cha:conclusion}} concludes this thesis and discusses some of our future research directions.

%% file: 2_Related.tex
\chapter{Literature Review}\label{cha:background}

In the pursuit of causal modeling in the domain of semantic video nnderstanding, this thesis provides a comprehensive overview of previous works that align closely with our task or are technically relevant to our solutions. In this chapter, we offer a broad perspective on related work, aiming to situate our research within the existing literature. Meanwhile, as we progress through the subsequent chapters, we will delve into each specific work and provide more detailed reviews. 

\section{Video Relation Detection}
In the domain of semantic video understanding, the study of visual relations is a critical aspect that connects individual visual elements. Early efforts exploit spatial relations implicitly to enhance object segmentation \cite{galleguillos2008object}, as well as discover human-centric relations explicitly to assist human-object interactions \cite{yao2010modeling}. In recent times, there have been significant advancements in exploring visual relations as an independent task, leading to significant contributions in structural and interpretable visual representation.
The task of relation detection originally emerged in the image domain, and Lu et al. \cite{lu2016visual} were among the first to formulate visual relations as triplets $<$subject, predicate, object$>$. This representation captures the relationships between different visual elements in an image. Building on this concept,  \cite{krishna2017visual,zhang2017visual} introduced the concept of a scene graph, which consists of multiple relation triples. In this structural representation, nodes represent visual objects, and edges depict the object relations. The introduction of the scene graph provides a more comprehensive and organized way of understanding the complex interactions between various visual elements, paving the way to visual semantic understanding at a higher level. 
In the video domain, relation detection (VidVRD), presents more challenges compared to ImageVRD, as it involves more intricate relation types and requires spatial-temporal object localization.
As a pioneer in the field, \cite{shang2021video,DBLP:conf/mm/JiLWSXRC21} construct the first benchmark dataset and grand challenge for Video Relation Detection (VidVRD) and introduces multi-stage pipelines, which include object tracking, relation tracklet generation, and relation tracklet association. Building upon this pipeline, numerous works have achieved notable results by incorporating various techniques. Some approaches utilize contextual knowledge \cite{tsai2019video} and long-range temporal information \cite{liu2020beyond} to enhance relation representation. Others employ spatial-temporal graphs \cite{qian2019video} to capture complex dependencies.
Recently, an end-to-end framework has been introduced by \cite{DBLP:conf/cvpr/ZhengCJ22}, advancing the state of the art in this field. Meanwhile, new task settings such as semi-supervised approaches \cite{DBLP:conf/iclr/Chen0C23} VidVRD and open-vocabulary VidVRD \cite{DBLP:conf/iclr/Gao0Z0S23} have been introduced, further enriching the research landscape in this area.
Despite the prosperity and progress in Video Relation Detection (VidVRD), one of the challenges that persists is relation imbalance. Specifically, the causal pattern of tail relations is difficult to capture, given the richness of interactions between objects and head relations. In response to this issue, our proposed approach, Interventional Video Relation Detection (IVRD), has been designed to address this imbalance and bring more informative relations into the prediction process. By doing so, IVRD aims to enhance the deployment of VidVRD in real-world scenarios, making it more effective and applicable in practical applications.

\section{Visual Question Answering}

Video Question Answering (VideoQA) moves beyond relation and focuses on the semantics of the entire video. As an extension of Image Question Answering (ImageQA), VideoQA addresses the limitations of static images by incorporating temporal information from videos. This led to new challenges and opportunities for the research community.

The progression of VideoQA is closely tied to the evolution of datasets. Early datasets like TGIF-QA \cite{jang2017}, MSRVTT-QA, and MSVD-QA are created based on short GIFs to answer templated questions about single object actions, such as recognizing attributes like color, position, or number. These initial efforts led to various architectural innovations for handling simple questions and short videos. One of the pioneering branches of research in this area is graph-based models, which represent videos as visual graphs based on frame or object features. For example, \cite{hga} and \cite{park2021bridge} build such typologies based on the heterogeneity of input modality, while \cite{mspan} enables progressive relational reasoning between multi-scale graphs. \cite{hqga} extends topology to the object level by treating objects in different frames as nodes and using their visual similarity as the edge.
Due to the inherent noise and language bias in existing datasets \cite{atp}, some more complex VideoQA benchmarks emerged \cite{next-qa, causalvid}, and research start to focus on causal and temporal relations between visual entities. In response to this advancement, a line of research has focused on processing video as a multi-level hierarchy. \cite{hcrn} first develops a bottom-up pathway by assembling information from the frame-level and then merging it to the clip-level. Subsequent works \cite{hostr, hqga,xiao2023contrastive,li2023discovering} extend the hierarchy to the object-level, designing a modular network to connect objects on the same frame. Most recently, \cite{vgt} achieves improvement by enabling relation reasoning in the sense of object dynamics through a temporal graph transformer.
Aside from these architectural advancements, a purely data-driven approach, cross-modal pretraining, has gained increasing popularity. Related approaches take advantage of the abundant vision-text data available on the web to train transformer-style models \cite{vaswani2017attention} in a self-supervised manner \cite{zhu2020actbert,seo2021look,yang2021just,xu2021videoclip,sevila}. By leveraging large-scale pretraining, these models can learn to generalize better and potentially improve performance on various video-language tasks.

Despite the apparent effectiveness of existing VideoQA methods, they tend to rely heavily on spurious correlations between the environment and the answer. While such shortcuts may go undetected in I.I.D (independent and identically distributed) test sets, they can be highly detrimental in a perturbed environment where the context changes. Therefore, it becomes crucial for a robust VideoQA model to capture the causal answering pattern that respects visual-linguistic alignments, allowing the model to focus on the critical scene to answer the question accurately.
Moreover, existing methods (even the state-of-the-art pretrained SeVila \cite{sevila}) demonstrate poor performance on complex VideoQA tasks. These tasks involve longer videos (over 80 seconds) with multiple objects (more than 5) interacting in different ways at different times. The reason behind their declination is the massive redundancy present in complex videos, where only a few key objects on selected frames are responsible for providing the answer. This intricacy makes capturing the causal pattern difficult. As a result, there is a growing need within the research community to explore and develop techniques specifically tailored for complex video question answering.

In view of the challenges mentioned earlier, this thesis aims to address the issues of spurious correlations and capture causal patterns in VideoQA for semantic video understanding.

\section{Causality in Vision}
Deep Neural Networks (DNNs) have demonstrated impressive performance in various computer vision tasks. However, existing methods heavily rely on fitting data distributions and often capture spurious correlations between different modalities, hindering their ability to learn essential causal relations behind multi-modal knowledge with good generalization abilities. Inspired by the fact that most data in deployment follows an O.O.D pattern, researchers have turned to causal reasoning as a means to handle the distribution shift, where the main objective is to discover causal patterns from input modalities.

In this section, we introduce relevant concepts of causality and then review some causality-inspired directions in the current vision research community.

\paragraph{Structural causal models}
\begin{wrapfigure}{R}{0.3\textwidth}
  \centering
  \includegraphics[width=0.3\textwidth]{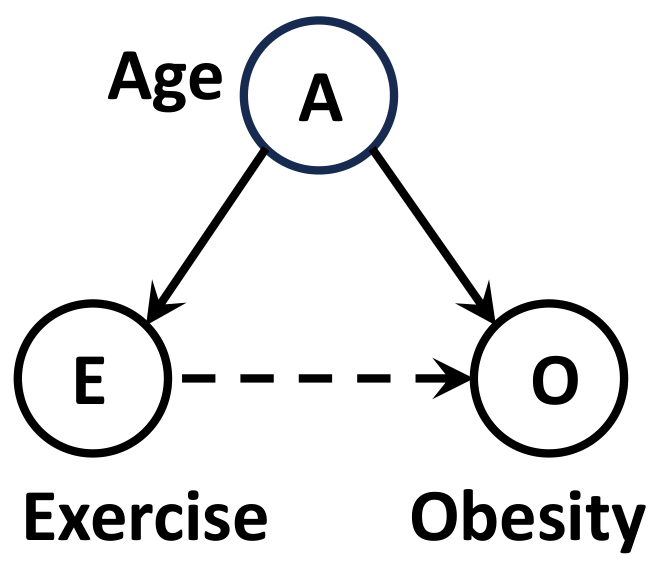}
  \caption{An example of SCM.}
  \label{fig:scm}
\end{wrapfigure}
Structural causal models (SCMs) are a class of graphical models used in causal inference and causal reasoning. They provide a formal framework for representing causal relationships between variables in a system and allow researchers to study the cause-effect relationships among variables.
In an SCM, variables are represented as nodes and causal relationships are depicted as directed edges between these nodes. Each variable is determined by a causal relation that expresses how it depends on its direct causes in the model. SCM captures the underlying mechanisms of how the variables interact with each other, providing a causal explanation of the system's behavior. 
Considering the SCM in \cref{fig:scm}, exercise $\leftarrow$ age $\rightarrow$ obesity, the elder people spend more time on physical exercise after retirement but, meanwhile, they are also easier to get obese due to the low metabolic rate, so the age (\aka the \textit{confounder} in causality) creates a \textit{spurious correlation} (indicate in dash arrow) that more physical exercise will increase the chance of getting obesity. As a result, the causal pattern between exercise and obesity became hard to discover.

\paragraph{Causal Intervention}
To tackle the above issue, we introduce causal intervention, a fundamental tool that involves actively perturbing a system to investigate the causal relationships between its variables. It goes beyond merely observing correlations between variables and allows researchers to explore the cause-and-effect mechanisms underlying complex systems. In causal intervention, specific variables are deliberately manipulated, and their effects on other variables are observed. Taking \cref{fig:scm} as an example, when studying the effect of ``excise'' on ``obesity'', we should intervene the variable ``age'', that is to control or stratify the ``age'' group and observe the effect of ``exercise'' on different ``age'' conditions and then conclude the outcomes of interventions. Causal intervention is a powerful tool in various fields, including medicine, social sciences, and machine learning, as it helps uncover causal dependencies, identify critical factors, and inform decision-making processes with a deeper understanding of the underlying causal structures.

\paragraph{Causal model in visual understanding}

The application of causal modeling in computer vision has been gaining momentum in various sub-domains. In scene graph generation, Tang et al. \cite{tang2020unbiased} propose an unbiased inference method to mitigate object class bias during training. In visual recognition, Wang et al. \cite{wang2021causal} introduce a causal attention module for self-annotating confounders in an unsupervised manner. Yue et al. \cite{DBLP:conf/cvpr/YueWS0Z21} present a generative causal model for zero-shot and open-set recognition, addressing the seen-unseen classification imbalance.
In the context of visual representation learning, \cite{DBLP:conf/nips/WangYHSZ21} develops an unsupervised disentangled representation learning approach based on self-supervised learning, which partitions the dataset into semantic-related subsets and learning an invariant representation across the subsets.

The application of causal modeling has recently expanded to the vision-language domain. In cross-modal representation learning,  \cite{wang2020visual} proposes a self-supervised feature representation learning method for advanced visual region encoding in Vision-Language tasks. Additionally, \cite{wang2023equivariant} explores the concept of equivariance for the similarity measure of vision-language models.

The most related works lie in the ImageQA task, where previous works have primarily focused on addressing language bias. Niu et al. \cite{niu2021counterfactual} introduce a counterfactual inference method to mitigate language bias, which improves O.O.D performance at the cost of I.I.D performance. To strike a balance between I.I.D and O.O.D performance, Niu et al. \cite{DBLP:conf/nips/NiuZ21} propose a novel debiasing method using introspective distillation. This method effectively improves the overall performance of the ImageQA model, making it more robust and accurate in both I.I.D and O.O.D scenarios
In contrast to the aforementioned causal ImageQA modeling, which primarily focuses on mitigating language bias through counterfactual inference, this thesis takes a different direction. Our primary objective is to address the harmful impact caused by the environmental scene in VideoQA.

%% file: work1/0_main.tex
\chapter{Interventional Video Relation Detection}
\label{cha:ivrd}

This chapter introduces our research on long tail imbalance in Video Relation Detection (VidVRD). Specifically, (VidVRD) aims to semantically describe the dynamic interactions across visual concepts localized in a video in the form of $<$\textit{subject}, \textit{predicate}, \textit{object}$>$. It can help to mitigate the \textit{semantic gap} between vision and language in video understanding, thus receiving increasing attention in multimedia communities. Existing efforts primarily leverage the multimodal/spatio-temporal feature fusion to augment the representation of object trajectories as well as their interactions and formulate the prediction of {predicates} as a multi-class classification task. Despite their effectiveness, existing models ignore the severe long-tailed bias in VidVRD datasets. As a result, the models' prediction will be easily biased towards the popular \textit{head} predicates (e.g., ``next-to" and ``in-front-of"), thus leading to poor generalizability.

To fill the research gap, this chapter proposes an Interventional Video Relation Detection (IVRD) approach that aims to improve not only the accuracy but also the robustness of the model prediction. Specifically, to better model the high-level visual predicate, our IVRD consists of two key components: 1) we first learn a set of predicate prototypes, where each prototype vector describes a set of relation references with the same predicate; and 2) we apply a causality-inspired intervention on the model input $<$\textit{subject}, \textit{object}$>$, which forces the model to fairly incorporate each possible predicate prototype into consideration. We expect the model to focus more on the visual content of the dynamic interaction between subject and object, rather than the spurious correlations between the model input and predicate labels. Extensive experiments on two popular benchmark datasets show the effectiveness of IVRD and also its advantages in reducing long-tailed bias. 

\input{work1/1_intro}
\input{work1/2_related}
\input{work1/3_method}

\input{work1/4_experiment}

\input{work1/5_conclusion}

\clearpage

%% file: work1/1_intro.tex
\section{Introduction}

\begin{figure}[t]
    \centering
    \subcaptionbox{\label{w1-fig:feat_analysis}}{
		\includegraphics[width=0.65\linewidth]{ 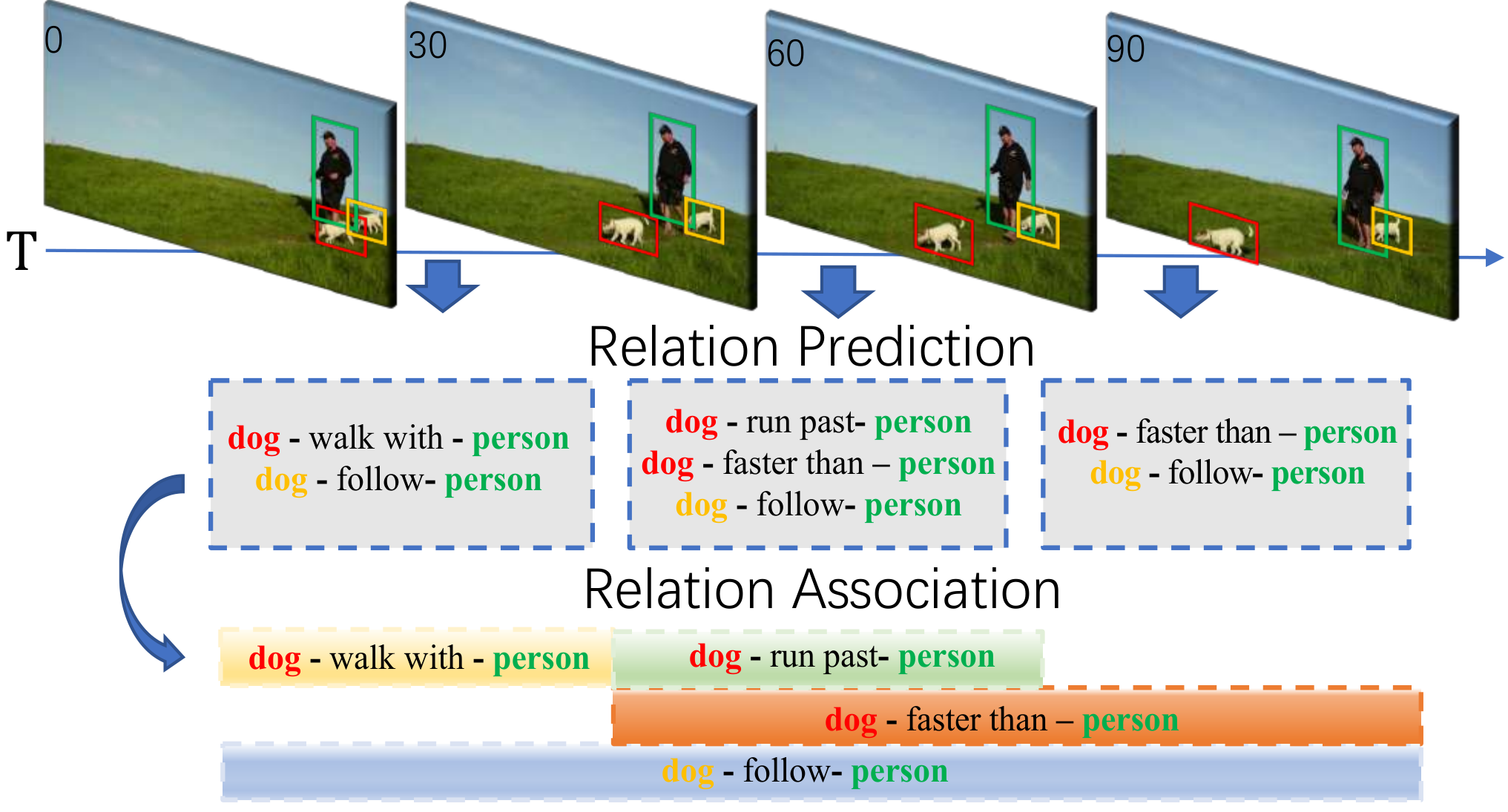}}
		\subcaptionbox{\label{w1-w1-fig:lambda}}{
	    \includegraphics[width=0.32\linewidth]{ 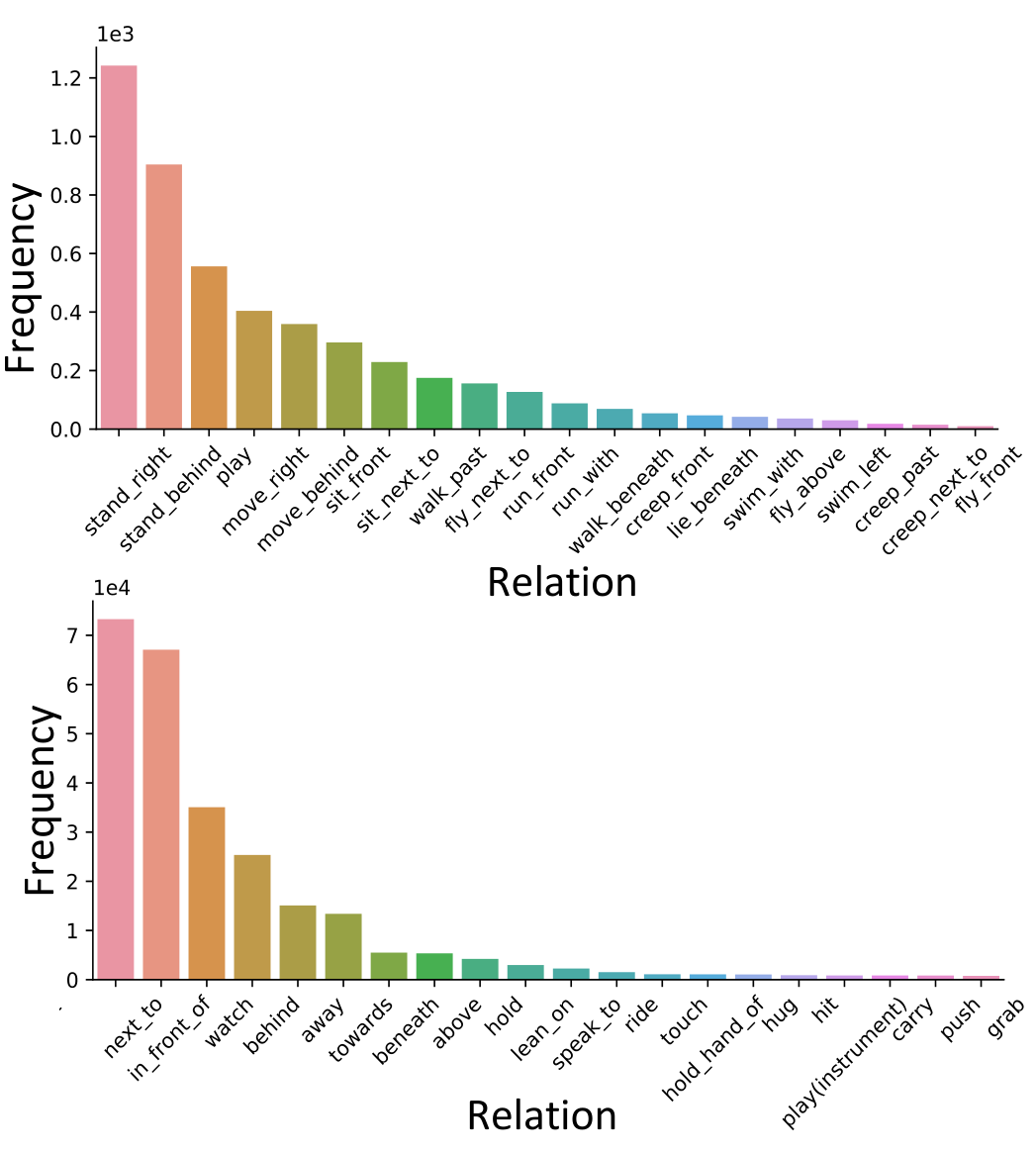}}
    \vspace{-8pt}
    \caption{(a) The pipeline of the VidVRD task. Relations are separately detected in each short segment and merged afterwards. 
    (b) denotes the long-tailed distributions of relations in ImageNet-VidVRD (b) and Vidor (c). Only 20 predicates are depicted to avoid cluttering.}
    \label{w1-fig:1}
    \vspace{-15pt}
\end{figure}

Deep learning techniques have boosted unprecedented progress on low-level visual tasks, e.g., the classification, detection, matching~\cite{yang2018person,yang2019interpretable}. However, it is still challenging for the machine to understand the visual content at a higher semantic level. Recent research efforts have focused on the detection of the relations in the visual (e.g., image and video) domain as a key step to narrowing the \textit{semantic gap} between low-level visual content and high-level semantic description. Visual Relation Detection not only involves detecting visual concepts but also identifying the high-level interactions across visual concepts (e.g., spatial/temporal relations, actions) in the form of relation triplet $<$\textit{subject}, \textit{predicate}, \textit{object}$>$ \cite{shang2019annotating}. It can significantly improve the ability of cross-media reasoning of multimedia systems, thus facilitating diverse downstream multimedia tasks, e.g.,  QA~\cite{next-qa}, and retrieval~\cite{yang2020tree,dong2021dual,MMGCN,wei2019neural}
.

Although extensive efforts have been devoted to relation detection, most works just focus on image-level detection. They lack the ability to reason complex relations that are inherently temporal and/or correlated in nature. For example, it is ambiguous to infer from a static image whether the ``dog" is ``faster than" or ``near" the person. Therefore, Video Visual Relation Detection (VidVRD) has emerged as a new challenging but important task, especially after the publication of the two benchmark datasets, ImageNet-VidVRD~\cite{shang2017video} and VidOR~\cite{shang2019annotating}. VidVRD usually requires to first extracting the object trajectories from the given video, and then predicting the semantic label (e.g., \textit{predicate}) of the dynamic interactions between the two video objects (e.g., \textit{subject} and \textit{object}). Compared with image-level detection, the task of VidVRD is much more challenging due to its complex temporal-spatial characteristics. Given a pair of localized video objects, there is usually more than one target relation between them, such as static spatial relations (e.g., ``next to", ``stand left") and dynamic temporal relations (e.g., ``move away", ``move toward"), thus leading to a much larger search space and higher complexity. 
Besides, the user annotations in current VidVRD datasets are usually unbalanced, as shown in   \cref{w1-fig:1} (b) and (c). For example, the number of spatial relation predicate ``next-to" is significantly larger than that of temporal relation predicate ``move-away" in  \cref{w1-fig:1} (b), since the spatial relation is much easier to be recognized and then annotated. Overall, VidVRD is a relatively high-level task and still in the early stage of research.   

Existing efforts \cite{cao2021vsrn,liu2020beyond,tsai2019video,shang2017video,qian2019video,su2020video} in VidVRD primarily leveraged the multimodal/spatio-temporal feature fusion~\cite{shang2017video,liu2020beyond,qian2019video,tsai2019video} 
to augment the representation of object trajectories as well as their dynamic interaction. 
They then formulated the prediction of the \textit{predicate} label in a relation triplet as a multi-class classification problem. Despite their effectiveness, existing methods did not consider the severe long-tailed bias in VidVRD datasets. 
In fact, prior works achieved their improvement mainly on the frequently-appearing (e.g., top-20) predicates, leaving the performance of the rest untouched. The long-tailed training data forces the model to learn a shortcut to maximize the overall likelihood of training data and therefore fail to deal with the less dominant (i.e., few-shot) predicates. If a model is trained to predict ``next-to" much more frequently than ``walk-with", then the former is more likely to prevail over the latter during testing. From a \textbf{causal} perspective, such a long-tailed dataset bias will inevitably lead to a bad \textit{spurious correlation} between the model input $<$\textit{subject}, \textit{object}$>$ and the predicate label. 

To fill the research gap in VidVRD, we develop an Interventional Video Relation Detection (IVRD) approach that aims to improve not only the accuracy but also the robustness of the model prediction on predicate labels. Our IVRD is mainly inspired by the causal intervention techniques~\cite{wang2020visual,yue2020interventional,yang2020deconfounded,zhang2020causal,yang2021SIGIR} in the field of vision and language, which can effectively prevent the model from utilizing the spurious correlation in biased training data. Specifically, in order to better model the high-level visual predicate, our IVRD consists of two key components: 1) we first learn a set of predicate prototypes, where each prototype vector describes a set of relation triplets with the same predicate; and 2) we apply an intervention on the model input $<$\textit{subject}, \textit{object}$>$, which forces the model to fairly incorporate each possible predicate prototype into consideration for making a more robust prediction. We expect our model to focus more on the visual content of the dynamic subject-object interaction, thus improving the prediction on temporal relations rather than the frequently-appearing spatial relations. We conducted extensive experiments and analysis on the two popular benchmark datasets, ImageNet-VidVRD~\cite{shang2017video} and VidOR~\cite{shang2019annotating}. Extensive evaluation results show that our IVRD can effectively reduce the long-tailed bias and improve the performance on the tail categories, thus indicating the effectiveness of the intervention strategy in our IVRD. We also observed substantial improvements over state-of-the-art methods.

In summary, this work makes the following main contributions:
\begin{itemize}
\item \textbf{Addressing Dataset Bias in Video Relation Detection}: Our research marks the first initiative to tackle the issue of long-tailed dataset bias specifically in the context of video relation detection. This contribution is significant as it paves the way for more balanced and fair model training processes, enhancing the accuracy and reliability of video relation detection systems. This advancement could lead to more equitable and efficient computer vision applications, setting a new standard in the field.

\item \textbf{Superiority on Both ``head" and ``tail" parts of VidVRD}: Our extensive experiments on two public VidVRD datasets are a testament to the effectiveness and superiority of the IVRD method. These experiments demonstrate not just the method's theoretical validity but its practical applicability and benefits in real-world scenarios. The results showcase marked improvements in accuracy and efficiency, validating the IVRD method as a significant leap forward in the field of video relation detection.
\end{itemize}

%% file: work1/2_related.tex
\section{Related Work} \label{w1-sec:ivrd-related-work}

\noindent\textbf{Visual Relation Detection}. 
Visual relation detection has been intensively studied since \cite{7298990} first brought up this notion in static images (ImgVRD). Understanding this
diversity of relationships in images is central to accurate visual retrieval~\cite{chen2016deep,chen2017cross,DBLP:journals/tip/ZhuLCLZ20,8839750,Gao2020Exploring} and to a richer semantic understanding of our visual world.
Ever since then, a huge proportion of works were dedicated to better feature extracting module\cite{dai2017detecting,xu2017scene,li2017scene,gu2019scene,shang2021video}.  \cite{zellers2018neural} is the first work that brings the unbalanced problem onto the table, and a following work \cite{chen2019knowledgeembedded} proposes an unbiased metric for comprehensive evaluation. \cite{liang2019vrrvg} also attacked the tail part of the dataset by pruning those dominant and easy-to-predict relations in the training set.

Compared to ImgVRD, VidVRD is more difficult because the temporal nature of video has enriched objects with more diverse combinations, such as relations involving actions (\eg ``chase") or position changes (\eg ``run-toward"). 
This poses a greater challenge for the model to incorporate both the statistical dependency and the visual information provided in videos. It also has a great potential for enhancing the downstream video tasks~\cite{hao2019neighbourhood,hao2020compact,hao2020person,Gao2021Pairwise,yang2020weakly}.
VidVRD did not receive sufficient attention until the release of ImageNet-VidVRD dataset \cite{shang2017video} and VidOR dataset \cite{shang2019annotating}, the former also proposed a three-stage detection framework that adopted by most following works. 
Qian et al. \cite{qian2019video} built a spatio-temporal graph convolutional networks within adjacent video segments to refine the object and relation features.
Sun et al. \cite{sun2019video} utilized language context feature along with spatial-temporal feature for predicate prediction. Cao et al. \cite{cao2021vsrn} proposed a 3DRN to fuse the spatio-temporal visual characteristics, object label features, and spatial interactive features for learning the relation instances with multi-modal cues. In order to capture relations involving long motions, Liu et al. \cite{liu2020beyond} proposed a sliding-window scheme to predict short-term and long-term relations simultaneously.
Similar to existing works, our IVRD adopts a similar three-stage framework as proposed in \cite{shang2017video}, however, we are the first to introduce the notion of causal intervention in VidVRD, which tackles the long-tail issue in existing benchmarks.

\smallskip
\noindent\textbf{Video Object Detection}. 
Video object detection is a fundamental step to tackling the VidVRD, some methods utilized optical flow to either propagate features from key frames to non-key frames~\cite{zhu2017deep} or enhance the features of each frame~\cite{zhu2017flowguided}.  Other works also employed RNN~\cite{hochreiter1997long} to process the video sequence, \cite{liu2019looking} reported the offline LSTM solutions that exploit all frames in the video, whereas the online solution\cite{8802920} only considered adjacent frames. 
Tracking-based methods utilize the temporal information in an efficient way. \cite{mao2019catdet, feichtenhofer2018detect} detected objects on fixed interval frames and tracked objects across frames.  \cite{luo2018detect} further made their improvement by detecting interval frames adaptively. In this chapter, we re-generated the object trajectories for ImageNet-VidVRD based on \cite{gkioxari2015finding}, which lies in the last category. It computed the cross-frame affinity score via visual feature similarity and IOU score and then associated bounding-box regions accordingly.

\begin{table}[h]
  \centering
  \caption{Key Related Papers}
    \begin{tabular}{lp{10cm}} 
    \toprule
    Paper & Remark \\
    \midrule
    \cite{shang2017video} & This paper proposed a three-stage detection pipeline, and we follow such pipeline in this work.
    \\
    \midrule
    \cite{yang2020deconfounded} & This concurrent work adopts an interventional method to tackle visual bias in video grounding tasks. However, the task we focus on and the form of the confounding dictionary are far from similar. \\
    \bottomrule
    \end{tabular}%
\end{table}%

%% file: work1/3_method.tex
\section{Preliminaries}

Given an untrimmed video $V$ with arbitrary temporal lengths, our goal is to detect visual relations from the video in the form of triplet $\langle\textit{subject},\textit{predicate},\textit{object}\rangle\in\mathcal{O}\times\mathcal{P}\times\mathcal{O}$ . The \textit{subject} (\textit{object}) corresponds to the semantic label of spatio-temporally localized subject (object) trajectory $\mathcal{T}_s$ ($\mathcal{T}_o$) in a temporal window. Each trajectory is a sequence of object bounding-boxes detected by the pretrained object detectors, e.g., Faster-RCNN~\cite{renNIPS15fasterrcnn}. The \textit{predicate} denotes the label of dynamic interaction between the \textit{subject} and \textit{object}. The $\mathcal{O}$ denotes the set of video object semantic categories and $\mathcal{P}$ denotes the set of \textit{predicate} semantic categories. 
As defined by Shang et al. \cite{shang2017video}, the standard video relation detection usually consists of three components: 1) detecting all the possible object trajectory proposals $\{\mathcal{T}_i\}_{i=1}^{T}$ in a short video segment; 2) predicting the short-term video relations; and 3) associating short-term relation prediction into video-level relation instances. 

\section{Method}\label{w1-sec:our-model}
In this section, we describe our proposed method in three steps: 1) object trajectory encoding and prediction in section \cref{w1-OTE}; 2) interventional relationship prediction with predicate prototype learning in section \cref{w1-IVRD_P}; and 3) learning objective in section \cref{w1-learning}.

\subsection{Object Trajectory Encoding and Prediction}\label{w1-OTE}

\smallskip
\noindent \textbf{Trajectory Proposal Generation.} The first step to training a good VidVRD model is to detect object trajectory proposals from the given video $V$. Following the popular strategies~\cite{shang2017video,sun2019video} in VidVRD, we first employ the pretrained image-level object detector, such as Faster-RCNN~\cite{renNIPS15fasterrcnn}, to densely detect visual objects in each video frame; we then can apply robust object tracking techniques~\cite{danelljan2014accurate} or customized bounding-box linking strategies~\cite{sun2019video,gkioxari2015finding} to link the detected object bounding-boxes across the video frames for generating valid trajectory proposals. The non-maximum suppression (NMS)~\cite{han2016seq} algorithm is applied to reduce the number of overlapping proposals. 

Formally, given a video (segment), we generate a set of object trajectory proposals $\{\mathcal{T}_i\}_{i=1}^{N}$, and each proposal is a set of (sampled) spatio-tempoal correlated bounding-boxes $\mathcal{T}=\{B_j\}_{j=1}^{M}$, where $N$ is the number of objects in a video segment, and $M$ is the number of frames in a video segment. Here, each bounding-box $B$ is represented by a high-dimensional visual Region-of-Interest (RoI) feature $\mathbf{v}\in\mathbb{R}^{d^*}$, extracted by pre-trained region proposal networks, and the box coordinates $(x,\gamma,w,h)$ where $(x, \gamma)$ are the central coordinates of the bounding-box, and $w$ and $h$ denote its width and height, respectively.

\smallskip
\noindent \textbf{Trajectory Proposal Encoding.} \label{w1-traj_rep} 
To effectively model the spatio-temporal characteristic of object trajectories, we apply a simple and effective method to learn the trajectory representation in this work. Given the set of bounding-box RoI features $\{\mathbf{v}_j\}_{j=1}^M$ of a trajectory proposal, we use the bidirectional Long short-term memory (LSTM)~\cite{hochreiter1997long} network to model the temporal dependencies between adjacent object bounding boxes. Formally, the LSTM unit, at the $j$-the time step, takes the bounding box RoI feature $\mathbf{v}_j$, the previous hidden state $\mathbf{h}_{j-1}$, and cell state $\mathbf{c}_{j-1}$  as inputs, and outputs the current hidden state  $\mathbf{h}_{j}$ and cell state $\mathbf{c}_{j}$:
\begin{equation}\label{w1-lstm}
\left(\mathbf{h}_j, \mathbf{c}_j\right)=\textrm{LSTM}\left(\mathbf{v}_j, \mathbf{h}_{j-1}, \mathbf{c}_{j-1}\right). 
\end{equation} 
The output of the bidirectional LSTM network is $\mathcal{H}=\{\mathbf{h}^*_1,\cdots,\mathbf{h}^*_M\}$ where $\mathbf{h}^*_j\in\mathbb{R}^{d}$ is the channel-wise vector concatenation of forward and backward hidden states at the $j$-the time step. The final video object trajectory representation $\mathbf{x}\in\mathbb{R}^{d}$ is obtained by applying average-pooling over $\mathcal{H}$. 

Note that we only exploit the visual RoI feature to learn the object trajectory representation. We do not use any form of language features, e.g., \textit{classme}~\cite{shang2017video,sun2019video}, since it will exacerbate the \textit{spurious correlation} between semantic context and predicate labels. A similar issue is also pointed out and addressed in a scene-graph generation work~\cite{tang2020unbiased} based on a causal inference method.

\smallskip
\noindent \textbf{Video Object Prediction.} \label{w1-traj_pre} We predict the semantic label of the object trajectory proposal by a prediction function $f^o(\cdot)$:
\begin{equation}\label{w1-obj_class}
y^o=\textrm{Softmax}\left(f^o(\phi(\mathbf{x}))\right),
\end{equation}
where $f^o(\cdot)$ is implemented by a fully-connected layer with learnable weighting parameter $\mathbf{W}_o\in\mathbb{R}^{d\times (|\mathcal{O}|+1)}$, $\phi(\cdot)$ denotes the non-linear activation function \textit{Relu}, and $y^o$ is a normalized probabilistic distribution over the $|\mathcal{O}|$ classes and a \textit{background} class. The class with the highest score is treated as the object label of the trajectory.

\subsection{Interventional Relation Prediction}\label{w1-IVRD_P}

\smallskip
\subsubsection{Baseline Solution}\label{w1-baseline} 
The key of the VidVRD task is to predict the \textit{predicate} label $y^p$ of the dynamic interaction between the localized \textit{subject} trajectory $\mathcal{T}_s$ and \textit{object} trajectory $\mathcal{T}_o$. The first step is to represent the dynamic interaction by an interaction function $g(\mathcal{T}_s, \mathcal{T}_o)$ which should consider not only the visual feature interaction but also the spatial interaction across two trajectories. We implement $g(\cdot)$ in a simple and effective way:
\begin{equation}\label{w1-fusion}
g(\mathcal{T}_s, \mathcal{T}_o) = \left[\mathbf{x}_s, \mathbf{x}_o, \mathbf{r}_{loc}\right],
\end{equation}
where $\left[\cdot\right]$ denotes the channel-wise vector concatenation, $\mathbf{x}_s\in\mathbb{R}^{d}$ and $\mathbf{x}_o\in\mathbb{R}^{d}$ denote the {visual features} of subject and object trajectories, respectively, and $\mathbf{r}_{loc}$ is the {spatial location feature} of the paired trajectories. We first compute the spatial location feature of the paired bounding boxes $(B^s, B^o)$ as a vector $\left[s_x, s_\gamma, s_w, s_h, s_a\right]$. It is calculated as follows: $s_x=\frac{x^s-x^o}{x^o}$, $s_\gamma=\frac{\gamma^s-\gamma^o}{\gamma^o}$, $s_w=\log\frac{w^s}{w^o}$, $s_h=\log\frac{h^s}{h^o}$, and $s_a=\log\frac{w_s\cdot h_s}{w_o\cdot h_o}$, where  $(x^s,\gamma^s,w^s,h^s)$, and $(x^o,\gamma^o,w^o,h^o)$ are the subject and object bounding-boxes respectively. The trajectory-level spatial location feature $\mathbf{r}_{loc}\in\mathbb{R}^{d}$ is computed by applying a bidirectional LSTM over the sequence of box-level spatial location feature, followed by an average-pooling. Then, we cast the output of the interaction into the predicate space by a predication function $f^p(\cdot)$. Formally, we can describe our baseline solution of predicate prediction as 
\begin{equation}\label{w1-baseP}
P(y^p|\mathcal{T}_s, \mathcal{T}_o) = \sigma\left(f^p\left(g(\mathcal{T}_s, \mathcal{T}_o)\right)\right),
\end{equation}
where $f^p(\cdot)$ is implemented by a fully-connected layer with weighting parameter $\mathbf{W}_p\in\mathbb{R}^{3d\times|\mathcal{P}|}$, and $\sigma(\cdot)$ is the sigmoid function $\sigma(x)=\frac{1}{1+\textrm{exp}(-x)}$ that transforms the output of $f^p(\cdot)$ into $(0,1)$.

\smallskip
\noindent{\textbf{Weakness.}} Most existing methods~\cite{sun2019video,qian2019video,liu2020beyond} use   \cref{w1-baseP} for predicting the predicate in a relation triplet with a different implementation of the interaction function $g(\cdot)$. Despite the effectiveness and simplicity of   \cref{w1-baseP}, it is easily misled by the long-tailed dataset bias that leads to a spurious correlation between the paired input $(\mathcal{T}_s, \mathcal{T}_o)$ and the predicate label. The first reason is that current pre-trained object detection networks may entangle the semantic context of the detected object with its visual content. Then, due to the imbalanced and incomplete relation annotations in current VidVRD datasets~\cite{shang2017video,shang2019annotating}, the model would spuriously predict the predicate of $(\mathcal{T}_s, \mathcal{T}_o)$ as a frequently-appearing label, e.g., $\langle\textit{person},\textbf{next-to},\textit{dog}\rangle$, while ignores the temporal-aware transitive label, e.g., $\langle\textit{person},\textbf{walk-with},\textit{dog}\rangle$. Although the spatial predicate \textit{next-to} sometimes is also a right answer, it is not the main focus of our VidVRD task. In short, existing methods mostly achieve their performance improvement on the frequently-appearing predicate categories while leaving the performance of the rest untouched. They thus fail to deal with less-dominant but sometimes more meaningful predicates, leading to poor generalization.

\begin{figure}[t]
  \centering
  \includegraphics[width=0.8\linewidth]{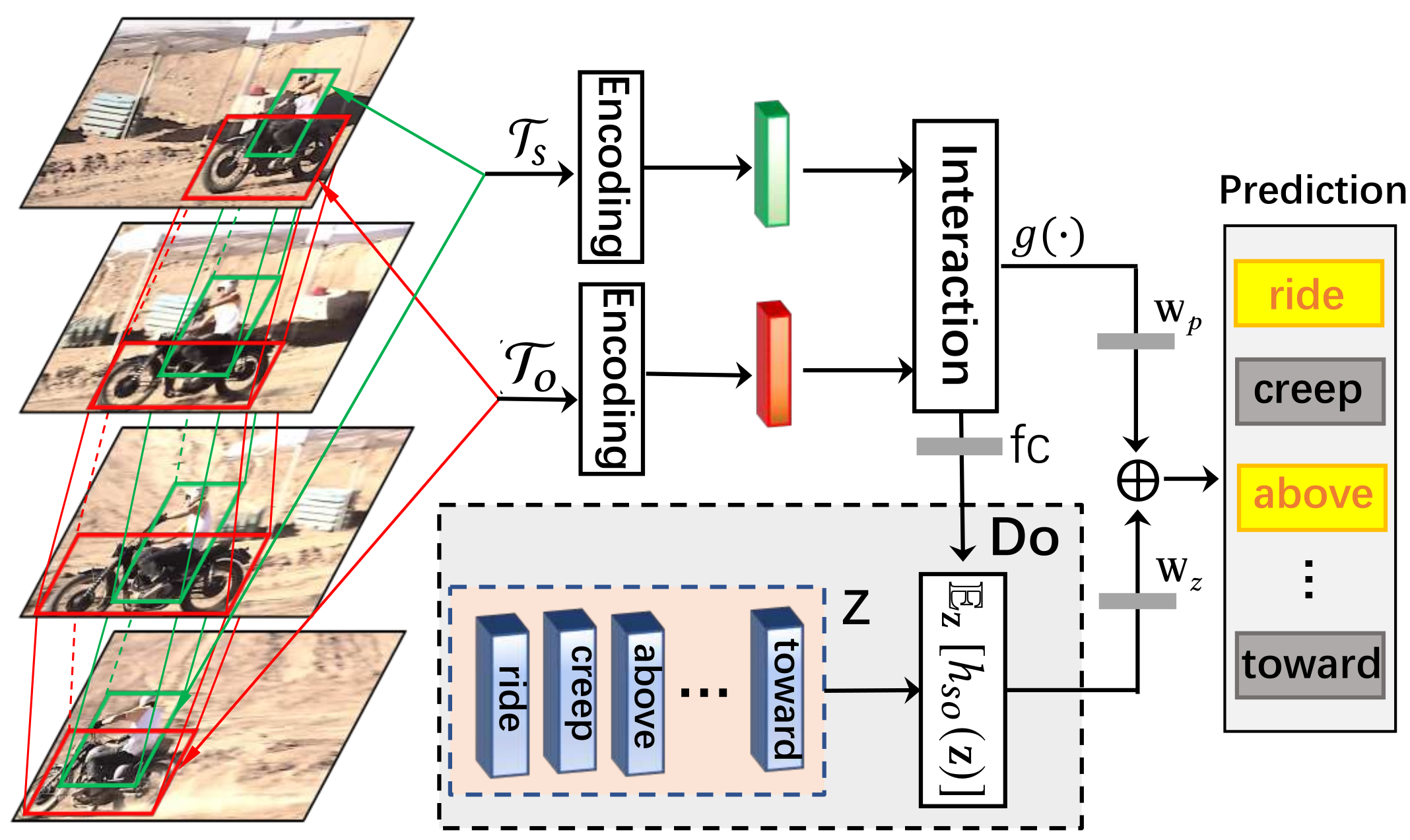}
  \caption{A brief overview of the \textit{predicate} prediction in our IVRD based on an intervention mechanism. $\oplus$ denotes element-wise add operation and ``fc" denotes a fully-connected layer. (\textit{some operations are omitted for simplicity.)}}
  \label{w1-fig:main}
 \vspace{-0.15in}
\end{figure}

\subsubsection{Our Solution (IVRD)}\label{w1-IVRD_learning} 
Our goal is to prevent the model from relying on the spurious correlation caused by the long-tailed dataset bias. We expect the model to treat all predicates fairly during training. Inspired by the recent advances of causal reasoning in vision and language~\cite{wang2020visual,yue2020interventional,yang2020deconfounded,zhang2020causal,yang2021SIGIR}, we develop an \textbf{I}nterventional \textbf{V}ideo \textbf{R}elation \textbf{D}etection (IVRD) method. 
As shown in \cref{w1-fig:main}, the basic idea of IVRD is to intervene in the paired model input $(\mathcal{T}_s, \mathcal{T}_o)$ using the \textit{do}-calculus~\cite{pearl2016causal,pearl2018book} that forces the model to fairly incorporate each possible predicate into consideration:
\begin{equation}\label{w1-do_caculus}
P(y^p|do\left(\mathcal{T}_s, \mathcal{T}_o\right))=\sum_{z\in\mathcal{Z}}P\left(y^p|\left(\mathcal{T}_s, \mathcal{T}_o\right), z\right)P(z),
\end{equation}
where $\mathcal{Z}=\{z_1, \cdots, z_{K_z}\}$ ($K_z=|\mathcal{P}|$) denotes a set of visual predicate prototypes~\cite{plesse2020modelling}. $P(z)$ denotes the prior probability of the predicate prototype $z$. Note that we force the prior $P(z)$ to be a constant $\frac{1}{K_z}$ to alleviate the long-tailed dataset bias. We can see from   \cref{w1-do_caculus} that the decision on the predicate label $y^p$ depends on not only the model input $\left(\mathcal{T}_s, \mathcal{T}_o\right)$ but also all the predicate prototypes $\{z_1, \cdots, z_{K_z}\}$. It can be seen as a way of removing spurious \textit{observational bias} by forcing the model to virtually see more. 

\subsubsection{Implementation of IVRD}\label{w1-implementation} To implement \cref{w1-do_caculus}, we organize all the predicate prototypes into a prototype dictionary $\mathbf{Z}=\left[\mathbf{z}_1,\cdots,\mathbf{z}_{K_z}\right]\in \mathbb{R}^{K_z\times d^p}$ where $\mathbf{z}\in \mathbb{R}^{d^p}$ represents the predicate prototype $z$.
Then, we first rewrite \cref{w1-do_caculus} as
\begin{equation}\label{w1-Eq.6}
P(y^p|do\left(\mathcal{T}_s, \mathcal{T}_o\right)):=\mathbb{E}_{\mathbf{z}}\left[\sigma \left(f\left(g(\mathcal{T}_s, \mathcal{T}_o), \mathbf{z}\right) \right)   \right]\approx \sigma\left( \mathbb{E}_{\mathbf{z}}\left[   f\left(g(\mathcal{T}_s, \mathcal{T}_o), \mathbf{z}\right)    \right]\right),
\end{equation}
where the approximation is based on the Normalized Weighted Geometric Mean (NWGM)~\cite{xu2015show} to avoid the expensive sampling of $\mathbb{E}_{\mathbf{z}}$. We further implement $f(\cdot)$ as a linear model: 
\begin{equation}\label{w1-Eq.7}
f\left(g(\mathcal{T}_s, \mathcal{T}_o), \mathbf{z}\right) = \mathbf{W}_p^T g(\mathcal{T}_s, \mathcal{T}_o) + \mathbf{W}_z^T h_{so}(\mathbf{z}),
\end{equation}
where $\mathbf{W}_z\in \mathbb{R}^{d^p \times |\mathcal{P}|}$ is a learnable matrix. $h_{so}(\mathbf{z})$ is a transformation function of $\mathbf{z}$, parameterized by the representation of $(\mathcal{T}_s, \mathcal{T}_o)$: $\left[\mathbf{x}_s, \mathbf{x}_o\right]$. Then, we move $\mathbb{E}_{\mathbf{z}}$ into $f(\cdot)$ and rewrite \cref{w1-Eq.6} as:
\begin{equation}\label{w1-Eq.8}
P(y^p|do\left(\mathcal{T}_s, \mathcal{T}_o\right)) \approx \sigma\left(\mathbf{W}_p^T g(\mathcal{T}_s, \mathcal{T}_o) + \mathbb{E}_{\mathbf{z}}  \left[h_{so}(\mathbf{z}) \right] \right).
\end{equation}
The key to compute \cref{w1-Eq.8} is $\mathbb{E}_{\mathbf{z}}  \left[h_{so}(\mathbf{z}) \right]$. We implement $h_{so}(\mathbf{z})$ as the scaled Dot-Product attention~\cite{vaswani2017attention} to adaptively assign weights on different predicate prototypes in dictionary $\mathbf{Z}$ with the input $\left[\mathbf{x}_s, \mathbf{x}_o\right]$ as \textit{query}. Then we have: 
\begin{equation}
\mathbb{E}_{\mathbf{z}}  \left[h_{so}(\mathbf{z}) \right] = \sum_{z}\left[ \textrm{Softmax}\left(\mathbf{K}^T \mathbf{q}/\sqrt{d^p}\right)\odot \mathbf{Z} \right]P(\mathbf{z}),
\end{equation}
where $\mathbf{q}=\mathbf{W}_1\left[\mathbf{x}_s, \mathbf{x}_o\right]$, $\mathbf{K}=\mathbf{W}_2\mathbf{Z}^T$ with learnable parameters $\mathbf{W}_1\in\mathbb{R}^{d^p\times 2d}$ and $\mathbf{W}_2\in\mathbb{R}^{d^p\times d^p}$.  If removing $\mathbb{E}_{\mathbf{z}}  \left[h_{so}(\mathbf{z}) \right]$ from   \cref{w1-Eq.8}), we can see that   \cref{w1-Eq.8}) will be the same as the baseline   \cref{w1-baseP}. 

\smallskip
\noindent{\textbf{Prototype Dictionary.}} 
The quality of the prototype dictionary $\mathbf{Z}$ is also important to our IVRD. We construct $\mathbf{Z}$ in a simple and effective manner.
Let $\mathbf{z}$ denote the prototype vector of a predicate $p\in\mathcal{P}$, e.g., \textit{ride}. We first obtain all the relation triplets sharing the predicate $p$ from the training set, $\{\left\langle s_i, p, o_i\right\rangle\}_{i=1}^{N_p}$. For each triplet $\left\langle s_i, p, o_i\right\rangle$, we represent it as $[\mathbf{v}^p_{s_i}, \mathbf{v}^p_{o_i}]$, where $\left[\cdot\right]$ denotes the channel-wise vector concatenation, and $\mathbf{v}^p_{s_i}\in\mathbb{R}^{d^p}$ is the average RoI features of the subject $s_i$'s bounding-boxes in all the annotated training instances of the triplet $\left\langle s_i, p, o_i\right\rangle$. We use the same way to compute the averaged object feature vector $\mathbf{v}^p_{o_i}$. During training, for the prototype of predicate $p$, we randomly sample $K_p$ ($K_p<=N_p$) triplets from $\{\left\langle s_i, p, o_i\right\rangle\}_{i=1}^{N_p}$ and calculate the prototype vector $\mathbf{z}$ as the averaged representation of the selected triplets: $\mathbf{z}=\frac{1}{K_p}\sum \mathbf{W}_3[\mathbf{v}^p_{s_i}, \mathbf{v}^p_{o_i}]$ where $\mathbf{W}_3\in \mathbb{R}^{d^p\times 2d^p}$ denotes a trainable transformation matrix.

So far, we have finished the introduction of our proposed IVRD method. Given the representation of subject trajectory and object trajectory and the predicate prototype dictionary, we can predict the predicate label using \cref{w1-Eq.8}. Together with the video object prediction in \cref{w1-obj_class}, we can finally finish the prediction of a video relation $\left\langle \textit{subject}, \textit{predicate}, \textit{object}\right\rangle$.

\subsection{Learning Objective}\label{w1-learning}
To train our method, we use the cross entropy loss function to calculate the loss of classifying video object trajectories in   \cref{w1-obj_class}, $\mathcal{L}_{obj}(\mathcal{T}, \hat{y}^o)$.
 Due to the fact that a subject-object pair might have multiple predicate labels, we use the binary cross entropy loss function for predicate prediction in   \cref{w1-Eq.8}, $\mathcal{L}_{pred}(\mathcal{T}_s, \mathcal{T}_o, \mathbf{Z}, \hat{y}^p)$. Both $\mathcal{L}_{obj}$ and $\mathcal{L}_{pred}$ are computed as the average loss over a training batch. $\hat{y}^o$ and $\hat{y}^p$ denote the ground-truths of video object and predicate, respectively. The final loss is given as:
\begin{equation}\label{w1-loss}
\mathcal{L} = \mathcal{L}_{obj} + \lambda\mathcal{L}_{pred}, 
\end{equation}
where $\lambda$ is a trade-off hyperparameter.

%% file: work1/4_experiment.tex
\section{Experiments}\label{w1-sec:experiments-analyses}
\subsection{Datasets, Settings, and Details} 

\smallskip
\noindent{\textbf{Datasets.}}  To validate the effectiveness of our method, we perform comprehensive experiments on two datasets ImageNet-VidVRD~\cite{shang2017video} and VidOR~\cite{shang2019annotating}. (1) ImageNet-VidVRD is the first VidVRD dataset, which consists of 1,000 videos
and is split into 800 training videos and 200 test videos. It covers 35 categories of subject/objects and 132 categories of predicates in total. The videos are densely annotated with relation triplets in the form of $<$\textit{subject}, \textit{predicate}, \textit{object}$>$ as well as the corresponding subject/object trajectories. There are 2,961 relation triplets in training set and 1,011 in test set. Particularly, the test set contains 258 zero-shot relation triplets that are unseen in training set. (2) VidOR consists of 7,000 videos for training, 835 videos for validation, and 2,165 videos for testing. We just use the validation set for evaluation, since the test set is used for grand challenge and not released for public use. Its annotations cover 80 categories of subject/objects and 50 categories of predicates. It has 6,258 relation triplets in the training set and 295 zero-shot relation triplets in the validation set.

\smallskip
\noindent{\textbf{Task Settings.}}
Following \cite{shang2017video}, we evaluate our method on two tasks: \textbf{relation detection} and \textbf{relation tagging}. (1) The objective of relation detection is to generate a set of $<$\textit{subject}, \textit{predicate}, \textit{object}$>$ relation triplets with spatio-temporally localized trajectories in the video. A correct prediction should have the same relation triplet as the ground-truth and both the subject and object trajectories should have a high voluminal Intersection over Union (vIoU) score with those in ground-truths. In our experiments, the overlapping threshold of vIoU is set to 0.5. (2) The tagging task is less strict. In the evaluation stage, it just requires the predicted relation triplets to fall into the corresponding video's ground-truth set without the requirement of accurate object localization. The effectiveness of tagging task can support various video applications, such as video retrieval.

\begin{table}[tbp]
	\centering
		\caption{Comparison with SOTA methods on ImageNet-VidVRD. Z-mAP denotes the mAP in zero-shot setting.}
		\vspace{-0.1in}
				\renewcommand{\arraystretch}{1.2}
			\setlength{\tabcolsep}{0.5mm}{
		\scalebox{0.9}{
			\begin{threeparttable}
				\begin{tabular}{c|ccc|c||ccc}
					\hline
					\multirow{2}{*}{Method} & \multicolumn{4}{c||}{Relation Detection} & \multicolumn{3}{c}{Relation Tagging} \\
					\cline{2-8}       
					& mAP   & R@50  & R@100 & Z-mAP & P@1   & P@5   & P@10 \\
					\hline
					\hline
					Shang's \cite{shang2017video} & 8.58  & 5.54  & 6.37  & 0.40  & 43.00  & 28.90  & 20.80  \\
					GSTEG \cite{tsai2019video} & 9.52  & 7.05  & 7.67  & 0.15  & 51.50  & 39.50  & 28.23  \\
					MHRA \cite{di2019multiple}  & 13.27  & 6.82  & 7.39  & 0.51  & 41.00  & 28.70  & 20.95  \\
					VRD-GCN \cite{su2020video} & 16.26  & 8.07  & 9.33  & 0.74  & 57.50  & 41.00  & 28.50  \\
					PPN-ST \cite{liu2020beyond} & 18.38  & 11.21  & 13.69  &  -    & 60.00  & 43.10  & 32.24 \\
					MHA \cite{su2020video}   & 19.03  & 9.53  & 10.83  & {1.18}  & 57.50  & 41.10  & 29.45 \\
					\hline\hline
					Baseline & 21.24  & 11.96  & 14.17  & 0.60  & 66.50  & 47.60  & 34.88  \\
					IVRD & \textbf{22.97} & \textbf{12.40} & \textbf{14.46} & \textbf{1.47}  & \textbf{68.83}  & \textbf{49.87} & \textbf{35.57} \\
					\hline
				\end{tabular}%
					\vspace{0.05in}
				\begin{tablenotes}
					\footnotesize              
					\item[1] We report the average performance of multiple runs with different random seeds. 
					\item[2] ``-" means that the result is not reported by the paper.
				\end{tablenotes} 
			\end{threeparttable} 
	}}%
	\label{w1-tab:vidvrd-result}%
		\vspace{-0.2in}
\end{table}%

\smallskip
\noindent{\textbf{Evaluation protocols and Metrics.}}
Following the standard evaluation protocol in \cite{shang2017video}, for relation detection task, we adopt mean Average Precision (mAP) and Recall@K (R@K) to evaluate the model performance. The mAP score calculates the average precision on each video and then takes the average over all videos. The Recall@K score measures the fraction of the positive relation detection in the Top-K results, and K is set to 50 and 100. As for relation tagging, 
following \cite{shang2017video}, we use Precision@K (P@K) to measure the tagging accuracy. K is set to 1, 5, and 10. 

\smallskip
\noindent{\textbf{Implementation Details.}}
(1) For frame-level pre-processing on ImageNet-VidVRD~\cite{shang2017video}, we adopt the fine-tuned Faster-RCNN \cite{renNIPS15fasterrcnn} on this dataset with Resnet101 \cite{he2015deep} backbone to detect object bounding boxes and extract RoI features, and videos are cut into consecutive 30-frames segments with 15-frames overlapping. We use our re-generated trajectories (termed \textbf{Traj-O}) based on the linking strategy in \cite{gkioxari2015finding} for evaluation on ImageNet-VidVRD by default. The standard trajectories (termed \textbf{Traj-S}) released by \cite{shang2017video} is also used for comparison with {Traj-O} in  \cref{w1-tab:ab1}. 
(2) For frame-level prepossessing on VidOR~\cite{shang2019annotating}, the pre-trained Faster-RCNN on the MS-COCO dataset is adopted with Resnet101 backbone to detect object bounding boxes and extract RoI features. All the trajectories are adopted from \cite{sun2019video}. (3) During training, the batch size is set to 64 for ImageNet-VidVRD and 256 for VidOR. The number of trajectory proposals in each batch is set to 8. We use all the 30 frames in each segment for trajectory encoding on ImageNet-VidVRD by default and randomly select 5 frames in each segment on VidOR for efficiency. All the experiments are conducted on Nvidia-DGX with the Adam optimizer and the learning rate of 0.001. We select the best model with the highest mAP score on a validation set for final evaluation. The hyperparameter $\lambda$ in Eq. (\cref{w1-loss}) is set to 50. 


\subsection{Overall Performance Comparison}
In this section, we compare our proposed IVRD method with the reported results of state-of-the-art (SOTA) methods on the two public benchmark datasets ImageNet-VidVRD~\cite{shang2017video} and VidOR~\cite{shang2019annotating}. The experimental results are depicted in  \cref{w1-tab:vidvrd-result} and \cref{w1-tab:vidor-result}. The \textbf{Baseline} method in  \cref{w1-tab:vidvrd-result} and \cref{w1-tab:vidor-result} denotes our baseline solution in Eq. (\cref{w1-IVRD_P}) without using the intervention strategy. It is implemented in the same framework and settings as \textbf{IVRD} for fair comparison. 

The SOTA methods on the two datasets include Shang's \cite{shang2017video}, GSTEG~\cite{tsai2019video}, PPN-ST~\cite{liu2020beyond}, VRD-GCN~\cite{qian2019video}, MHRA~\cite{di2019multiple}, MHA~\cite{su2020video}, MAGUS.Gamma~\cite{sun2019video}, RELAbuilder~\cite{zheng2019relation}, and the regenerated results of CAI~\cite{zhuang2017towards} and GSTEG in MAGUS.Gamma. 
\begin{itemize}
	\item Shang's \cite{shang2017video} is the first baseline method for the VidVRD task and released the trajectories and relation association method for public evaluation on ImageNet-VidVRD. 
	\item GSTEG, VRD-GCN, and PPN-ST mainly focus on modeling the spatio-temporal dependencies/contexts by different learning techniques, such as Markov conditional random fields and graph convolutional network (GCN). Particularly, PPN-ST extracts refined video features with spatial-temporal GCN.
	\item MHRA~\cite{di2019multiple} and MHA~\cite{su2020video} are mainly designed for improving the short-term relation association, where MHRA generates the hypothesis for each subject-object pair, whereas MHA generates the hypothesis on each possible relation instead.
	\item MAGUS.Gamma and RELAbuilder are the top 2 competitors in ACM MM 2019 Video Relation Understanding Challenge~\cite{shang2019relation}. In particular, MAGUS.Gamma released its generated object trajectories on VidOR for public evaluation (used by our IVRD).
\end{itemize}
We have the following observations from  \cref{w1-tab:vidvrd-result} and \cref{w1-tab:vidor-result}:
\begin{itemize}
	\item Our IVRD achieves the best performance on ImageNet-VidVRD in all metrics. 
	IVRD obtains the highest mAP score and the best tagging performance on VidOR. 
	The worse Recall scores in  \cref{w1-tab:vidor-result} are mainly because PPN-ST~\cite{liu2020beyond} uses a much stronger object detector for generating trajectories while we just use the released trajectories from MAGUS.Gamma~\cite{sun2019video}. The overall comparison results clearly indicate the effectiveness of our IVRD.
	\item IVRD consistently outperforms the baseline in all metrics on two datasets, which clearly demonstrates the effectiveness of our proposed intervention strategy in Eq. (\cref{w1-do_caculus}).
	\item IVRD achieves more significant relation tagging performance than all other methods on two datasets. This is because our intervention strategy focuses on the predicate label prediction, thus leading to a large improvement on the relation tagging task where accurate object localization is not required
	\item In  \cref{w1-tab:vidvrd-result}, our IVRD achieves the best \textbf{zero-shot} detection result (Z-mAP, \%), which reflects the advantages of IVRD in detecting unseen video relations and indicates its good generalizability.
\end{itemize}

\input{ work1/tab/tab2}
\input{ work1/tab/tab3}

\subsection{Study of IVRD} 
We conduct extensive experiments to investigate the effectiveness of IVRD by answering the following research questions: 
1) how does the construction strategies of the predicate prototypes affect the performance of IVRD? (\textbf{R1}), 
2) how does IVRD perform with different features of Sub-Object (SO) Interaction, different trajectory generation strategies when compared to the baseline? (\textbf{R2}), 
3) how does the hyperparameters, like the number of sampled frames in segments, affect the performance of IVRD? (\textbf{R3}), 
4) how does IVRD perform on the \textit{tail} and \textit{head} predicate groups? (\textbf{R4}).

\smallskip
\noindent{{\textbf{Predicate Prototype Dictionary Construction.}}
As shown in Eq. (\cref{w1-Eq.8}) in the section \cref{w1-implementation}, our IVRD requires a predicate prototype dictionary $\mathbf{Z}$ for intervening the common input $(\mathcal{T}_s, \mathcal{T}_o)$. In this section, we design two variants of IVRD, termed \textbf{IVRD-1} and \textbf{IVRD-2}, to investigate the effect of the construction strategies: 1) IVRD-1: for constructing the dictionary, we directly compute each predicate prototype vector as the channel-wise concatenation of the averaged subject RoI features and the averaged object RoI features in all the annotated relation instances with the corresponding predicate. 2) IVRD-2: we compute the triplet-level prototype vector in the same way as IVRD in the section \cref{w1-implementation}, and then compute each predicate prototype vector as the averaged triplet-level representation. The predicate prototype dictionaries of both IVRD-1 and IVRD-2 do not require randomly sample the triplets and are fixed (without the learnable transformation matrix.)

Table \cref{w1-tab:ab2} shows the performance comparison among IVRD, IVRD-1, IVRD-2, and the baseline. We have the following observations:
\begin{itemize}
	\item Both IVRD and its variants clearly improve the baseline in all metrics, especially in mAP and P@1, which indicates the prototype dictionary-based intervention strategy is effective.
	\item IVRD performs much better than IVRD-1 and IVRD-2. It indicates that our learnable prototype dictionary with a dynamic triplet sampling can work effectively in IVRD. In our design, the triplet sampling and the learnable pooling operation enables the loss to be back-propagated to the triplet level, so the model can perceive the discrepancy between subject-object pairs with the same predicate, e.g., $<$dog, chase, car$>$ and $<$boy, chase, ball$>$.
\end{itemize}

\begin{table}[htbp]
	\centering
	\caption{Comparison with the baseline on two datasets using different subject-object interaction features and trajectories. "-V" denotes that only visual RoI feature is used.}  
			\vspace{-0.1in}
		\renewcommand{\arraystretch}{1.2}
		\setlength{\tabcolsep}{1.5mm}{		
	\scalebox{0.9}{
		\begin{tabular}{c|ccc|ccc|ccc}
			\hline
			 \multirow{3}{*}{Method} & \multicolumn{6}{c|}{ImageNet-VidVRD} & \multicolumn{3}{c}{\multirow{2}{*}{VidOR}} \\
			\cline{2-7}                & \multicolumn{3}{c|}{Traj-O} & \multicolumn{3}{c|}{Traj-S} &\\
				\cline{2-10}             & mAP   & R@50  & R@100 & mAP   & R@50 & R@100  & mAP   & R@50 & R@100  \\
			\hline
			\hline
			Baseline-V & 19.45 & 10.01 &11.83 & 17.59  & 9.45  & 11.23 & 5.12  & 6.57  & 8.80 \\
			 IVRD-V & \textbf{21.20} & \textbf{10.51} & \textbf{12.89} & \textbf{19.24} & \textbf{9.89} & \textbf{12.08} & \textbf{6.66} & \textbf{6.92} & \textbf{9.11} \\
			\cline{1-10}          
			 Baseline & 21.24 & 11.96 & 14.17 & 18.48  & 10.44 & 12.16 & 6.07  & 7.07  & 9.08\\
			 IVRD  & \textbf{22.97} & \textbf{12.40} & \textbf{14.46} & \textbf{19.88} & \textbf{10.69} &\textbf{12.60} & \textbf{7.42} & \textbf{7.36} & \textbf{9.41} \\
			\hline
	\end{tabular}}}
	\label{w1-tab:ab1}
\end{table}

\smallskip
\noindent{\textbf{SO-Interaction Representation and Trajectories.}}
As mentioned in Eq. (\cref{w1-fusion}), to effectively represent the dynamic subject-object (SO) interaction, we concatenate the subject's visual embedding, object's visual embedding, and their relative spatial location feature. We investigate the effect of this design in  \cref{w1-tab:ab1} on the two datasets by comparing the performance of IVRD with/without the spatial location feature. We also conduct the comparison on ImageNet-VidVRD in two different trajectory settings: 1) Traj-O: our re-generated object trajectories (Used by default) and 2) Traj-S: the standard trajectories released by Shang's~\cite{shang2017video}. We have the following observations from  \cref{w1-tab:ab1}:

\begin{itemize}[leftmargin=*]
	\item Both IVRD and the baseline improve over their counterparts, IVRD-V and Baseline-V, in all metrics and trajectory settings. It indicates the effectiveness of the relative spatial location feature in Eq. (\cref{w1-fusion}), since the prediction of most predicates in ImageNet-VidVRD requires a good understanding the relative spatial locations of subject and object, such as ``ride", ``follow", and ``walk-with".
	\item Our re-generated object trajectories (Traj-O) achieve better relation detection results than the released trajectories (Traj-S) by \cite{shang2017video}, which demonstrates the advantage of our bounding-box linking technique \cite{gkioxari2015finding} over the widely used method in \cite{shang2017video}.
	\item IVRD outperforms the baseline consistently in all features and trajectory settings, which further demonstrates the robustness of our intervention strategy.
\end{itemize}

\smallskip
\noindent{\textbf{Hyperparameters.}}
We test the performance of IVRD on ImageNet-VidVRD with: 1) different numbers of sampled triplets for learning the prototype dictionary, $K_p \in \{5, 10, 15, \cdots, 50\}$, as depicted in \cref{w1-fig:sample} (a), and 2)  different numbers of sampled frames in a segment for trajectory encoding, $M \in\{3, 5, 7, 10, 15, 20, 30\}$, as depicted in \cref{w1-fig:sample} (b).  In \cref{w1-fig:sample} (b), ``Ave" denotes a variant of IVRD that uses the averaged bounding-box RoI features to encode trajectories. ``Ave" is shown as a baseline result. We have the following observations:
\begin{itemize}[leftmargin=*]
	\item We observe in \cref{w1-fig:sample} (a) that the mAP score of IVRD goes up stably when we increase $K_p$ from 5 to 20 and tends to fluctuate when $K_p \in \{20, \cdots, 50\}$. It is reasonable, since the number of relation triplets of each predicate is unbalanced. Using too larger $K_p$ will lead to an over-smoothed prototype representation of the \textit{head} predicates. In this work, we set $K_p$ to 20 as a trade-off, because the average number of relation triplets of each predicate in the training set is 22.8.
	\item We see in \cref{w1-fig:sample} (b) that a larger number of sampled frames in a segment can lead to a better performance of IVRD, since the number of sufficient frames may facilitate the modeling of temporal dependence across frames. But a large number of sampled frames in a segment also incurs a significantly high computational complexity for a larger video dataset. Therefore, we use all 30 frames for the small dataset, while we just sample 5 frames in a segment, as a trade-off, for the large dataset, VidOR. The performance on VidOR could be higher, if we increase the number of sampled frames.
\end{itemize}

\begin{figure}[t]
  \centering
  \includegraphics[width=0.85\linewidth]{ 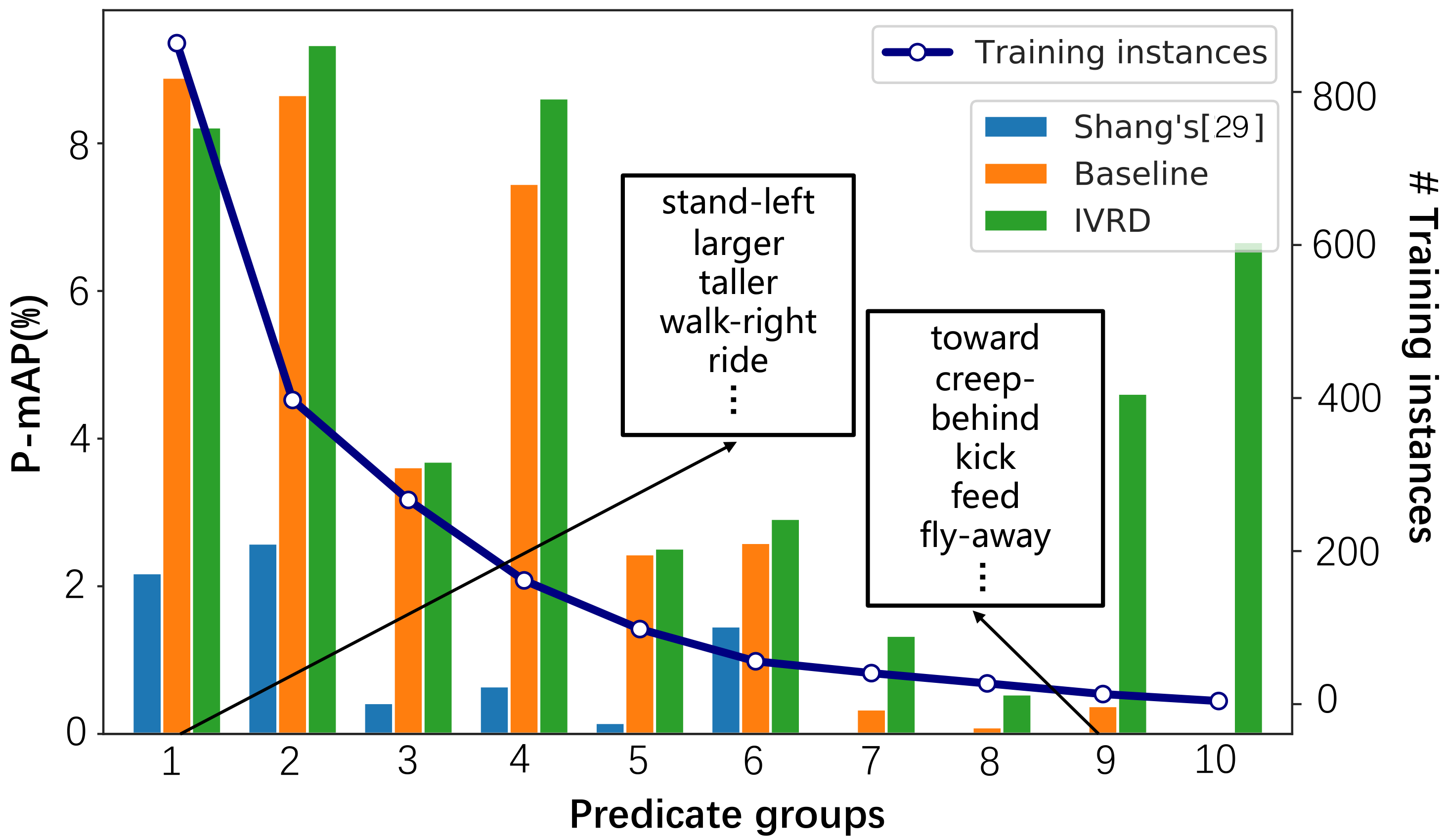}
  \caption{Performance comparison (P-mAP, \%) on 10 predicate groups in ImageNet-VidVRD dataset. P-mAP is a macro-averaging mAP metric over the predicate categories in the same group. The solid line in blue depicts the average training instances in each group.}
  \label{w1-fig:improve}
\end{figure}

\smallskip
\noindent{\textbf{Long-tailed Predicate Distribution.}}
We investigate how our IVRD performs on the long-tailed predicate categories in ImageNet-VidVRD. We first sort the 132 predicate categories in a descending order based on their corresponding training instances and then we divide the sorted 132 predicate categories into 10 predicate groups evenly for easy illustration. As depicted in \cref{w1-fig:improve}, the top groups mostly correspond to the \textit{head} predicate categories. We have the following observations from \cref{w1-fig:improve}:
\begin{itemize}[leftmargin=*]
	\item IVRD performs slightly worse than the baseline on the first group and achieves a bit better performance than the baseline on the top-6 predicate groups except the first group. However, our IVRD can significantly outperform the baseline on the four \textit{tail} groups. The observation verifies the rationale of our aforementioned analysis on the long-tailed dataset bias. By relying on the dataset bias, the baseline performs very well on the popular predicate categories, but it cannot effectively generalize to the few-shot and zero-shot predicate categories. While, IVRD has a stronger ability of predicting the \textit{tail} predicates, which should be attributed to the effectiveness of the intervention strategy in Eq. (\cref{w1-do_caculus}).
\end{itemize}

\begin{figure}[t]
    \centering
    \subcaptionbox{}{
		\includegraphics[width=0.45\linewidth]{ 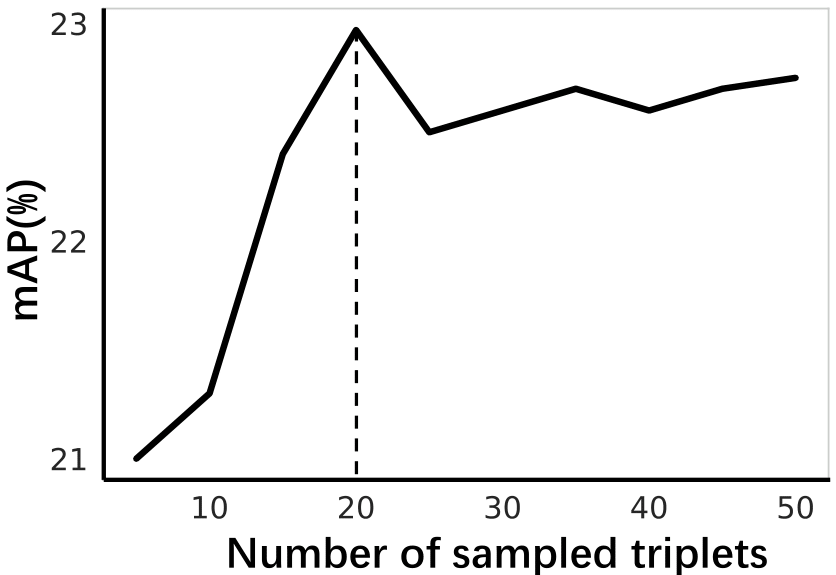}}
    \hspace{3pt}
		\subcaptionbox{\label{w1-fig:lambda}}{
	    \includegraphics[width=0.45\linewidth]{ 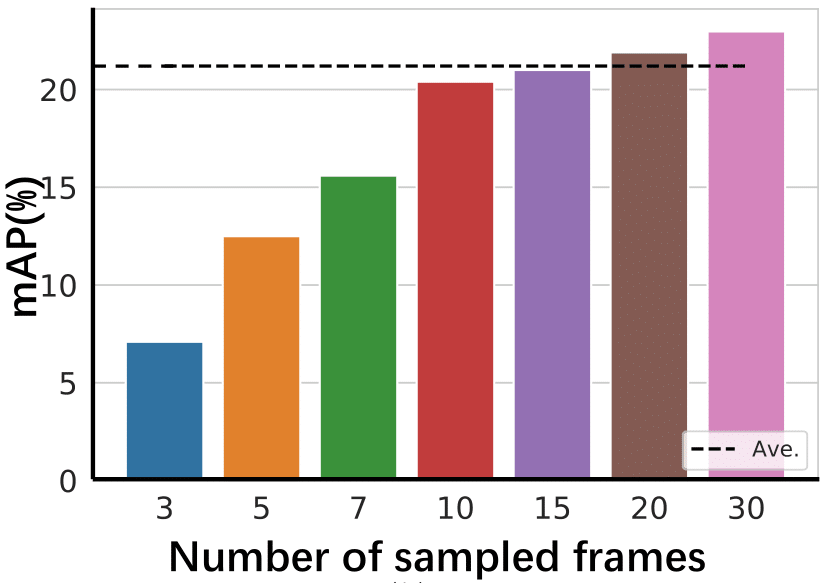}}
    \caption{Evaluation of IVRD with (a) different numbers of sampled triplets for learning the dictionary, and (b) different numbers of sampled frames for trajectory encoding.}
    \label{w1-fig:sample}
\end{figure}

\subsection{Quantitative Results and Analysis}

\smallskip
\noindent{\textbf{Visualization of the Attended Prototypes.}}
The key of our IVRD is to compute $\mathbb{E}_{\mathbf{z}}  \left[h_{so}(\mathbf{z}) \right]$. It is implemented by a scaled Dot-Product attention over the prototype dictionary, given the features of the subject-object trajectories as the \textit{query}. \cref{w1-fig:att-fig} shows two interesting real cases where the top-5 attended predicate prototypes are shown. We have the following observations:
\begin{itemize}[leftmargin=*]
	\item In \cref{w1-fig:att-fig} (a), given the paired trajectories of $<$``person", ``car"$>$ as input, we surprisingly find that both the top-2 attended prototypes (``toward" and ``move-toward") are the target predicates to be predicted and they semantically overlap with each other. Although the 5-th prototype ``chase" is not the target predicate, it is semantically correlated with ``move-toward". When the input is $<$person, bicycle$>$, all the top-5 attended prototypes (``ride", ``above", ``sit-above", ``taller", and ``stand-above") are semantically correlated/similar. It demonstrates the effectiveness of the computation of $\mathbb{E}_{\mathbf{z}}  \left[h_{so}(\mathbf{z}) \right]$. Less popular predicates have more chances to join the model training based on our intervention strategy, and the model is also able to capture the latent relationship between different predicates rather than treating each predicate independently.
	\item In the case depicted in \cref{w1-fig:att-fig} (b), where the input is  $<$``adult", ``adult"$>$. Three target predicates (``in-front-of", ``hold-hand-of", and ``watch") fall in the top-5 attended prototypes. In particular, the 4-th attended prototype is ``hug" which has a strong temporal causal relationship with ``hold-hand-of". In the training set of VidOR, the action ``hold-hand-of" is frequently (30\%) followed by the action ``hug" in the same video. It further verifies the rationale of our IVRD.
\end{itemize}

\begin{figure*}[th]
  \centering
  \includegraphics[width=5.2 in]{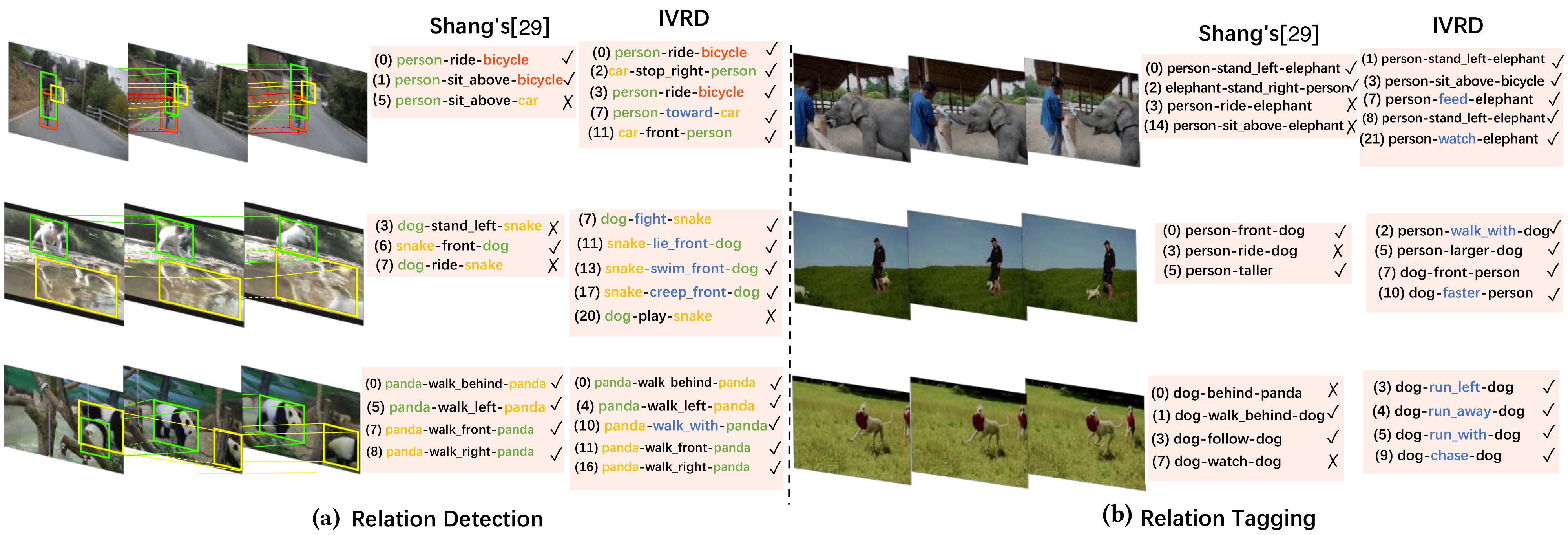}
  \caption{Qualitative examples of IVRD and Shang's \cite{shang2017video}. The correct relation instances in the top 30 results are shown, and their ranks are marked in front of them with parentheses. (a) Results of relation detection and (b) results of relation tagging.}
  \label{w1-fig:case-study}
 	\vspace{-0.13in}
\end{figure*}

\smallskip
\noindent{\textbf{Visualization of Detection and Tagging Results.}}
We illustrate the qualitative results of the relation detection and tagging tasks in \cref{w1-fig:case-study}. Three real cases are shown for each task. In \cref{w1-fig:case-study} (a), we observe that our IVRD
can not only well predict the spatial predicates, e.g., ``stop-right", but also the transitive predicates, e.g. ``toward", ``creep-front", ``fight", and ``walk-with", where the model is required to accurately understand the spatio-temporal interaction across objects. We see a wrong prediction, $<$``dog", ``play", ``snake"$>$, in \cref{w1-fig:case-study} (a). But such prediction is acceptable and even may be correct, just because the annotators thought the ``dog" was fighting with the ``snake". The action ``fight" is visually similar to the action ``play".
In \cref{w1-fig:case-study} (b), more accurate tagging results are ranked in top-20, where the temporal predicate (e.g. ``feed") and few-shot predicate (e.g. ``faster") are successfully predicted by IVRD.

\begin{figure}[t]
  \centering
  \includegraphics[width=5.2in]{ 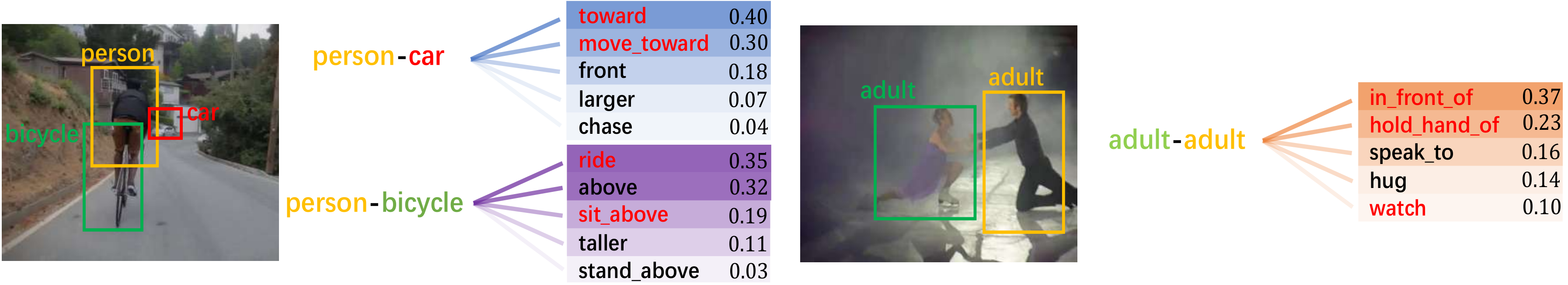}
    \caption{Visualization of the top-5 attended predicate prototypes, given the features of the subject-object trajectories, on ImageNet-VidVRD (left) and VidOR (right), where the numbers denote the normalized attention weights. Besides, the prototypes in red correspond to the ground-truth predicate labels.}
  \label{w1-fig:att-fig}
\end{figure}

%% file: work1/tab/tab2.tex
\begin{table}[tbp]
	\centering
			\renewcommand{\arraystretch}{1.1}
	\setlength{\tabcolsep}{2.7mm}{
		\caption{Comparison with SOTA methods on VidOR.}
		\scalebox{0.95}{
			\begin{tabular}{c|ccc||cc}
				\hline
				\multirow{2}{*}{Method} & \multicolumn{3}{c||}{Relation Detection} & \multicolumn{2}{c}{Relation Tagging} \\
				\cline{2-6}       
				&mAP   & R@50  & R@100 & P@1   & P@5 \\
				\hline
				\hline
				RELAbuilder~\cite{zheng2019relation}  & 1.47  & 1.58  & 1.85  & 33.05  & 35.27  \\
				OTD+CAI~\cite{sun2019video,zhuang2017towards}   & 5.65  & 6.19  & 8.16  & 48.31  & 38.49  \\
				OTD+GSTEG~\cite{sun2019video,tsai2019video} & 5.58  & 6.40  & 8.43  & 51.20  & 37.26  \\
				MAGUS.Gamma~\cite{sun2019video}  & 6.56  & 6.89  & 8.83  & 51.20  & 40.73  \\
				PPN-ST~\cite{liu2020beyond} & 6.85  & \textbf{8.21}  & \textbf{9.90}  & 48.92  & 36.78  \\
				MHA~\cite{su2020video}   & 6.59  & 6.35  & 8.05  & 50.72  & 41.56  \\
				\hline
				\hline
				Baseline & 6.08  & 7.07  & 9.08  & 48.60  & 40.00  \\
				IVRD  & \textbf{7.42} & {7.36} & {9.41} & \textbf{53.40} & \textbf{42.70} \\
				\hline
	\end{tabular}}
	\label{w1-tab:vidor-result}}%
\end{table}%

%% file: work1/tab/tab3.tex
\begin{table}[tbp]
	\centering
	\caption{Effects of different construction strategies of the prototype dictionary on ImageNet-VidVRD.}
				\renewcommand{\arraystretch}{1.2}	
		\setlength{\tabcolsep}{2.5mm}{
	\scalebox{0.95}{
		\begin{tabular}{c|ccc|c||ccc}
			\hline
		\multirow{2}{*}{Method} 	& \multicolumn{4}{c||}{Relation Detection} & \multicolumn{3}{c}{Relation Tagging} \\
		\cline{2-8}
			 & mAP   & R@50  & R@100  & Z-mAP & P@1   & P@5  & P@10 \\
			\hline\hline
			Baseline  & 21.24  & 11.96 & 14.17  & 0.60  & 66.50  & 47.60 & 34.88 \\
			IVRD-1  & 21.67  & 11.98  & 14.15  & 0.75  & 65.00  & 47.8 & 34.56  \\
			IVRD-2  & 22.18  & 12.36  & 14.40 & 0.47  & 66.25  & 49.70 & 35.50 \\
			IVRD    & \textbf{22.97}  & \textbf{12.40}  & \textbf{14.46}  & \textbf{1.47}  & \textbf{68.83}  & \textbf{49.87}  & \textbf{35.57} \\
			\hline
	\end{tabular}}%
	\label{w1-tab:ab2}}%
\end{table}%

%% file: work1/5_conclusion.tex
\section{Conclusion}
This chapter presented a causality-inspired video relation detection approach, termed IVRD, that improved the model's robustness by discovering the causal relation pattern. In particular, the intervention is performed via a predicate prototype dictionary that enables the model to incorporate every predicate prototype fairly for making a more robust decision. Extensive experiments on two popular VidVRD datasets show that our IVRD is effective on few-shot and even zero-shot transitive predicate categories.

%% file: work2/main.tex
\chapter{Invariant Grounding for VideoQA}
\label{cha:igv}

\input{ work2/sec/0_abstract}

\input{ work2/sec/1_introduction_new}

\input{ work2/sec/2_related}

\input{ work2/sec/2_preliminary_new}

\input{ work2/sec/3_method_new}

\input{ work2/sec/4_results}

\input{ work2/sec/5_conclusions}


%% file: work2/sec/0_abstract.tex
This chapter introduces our first work on the perturbed imbalance for Video Question Answering (VideoQA). Specifically, VideoQA is the task of answering questions about a video. At its core is in understanding the alignments between visual scenes in video and linguistic semantics in question to yield the answer.
In leading VideoQA models, the typical learning objective, empirical risk minimization (ERM), latches on superficial correlations between video-question pairs and answers as the alignments.
However, ERM can be problematic, because it tends to over-exploit the spurious correlations between question-irrelevant scenes and answers, instead of inspecting the causal effect of question-critical scenes.
As a result, the VideoQA models suffer from unreliable reasoning.

In this work, we first take a causal look at VideoQA and argue that invariant grounding is the key to ruling out spurious correlations.
Towards this end, we propose a new learning framework, Invariant Grounding for VideoQA (IGV), to ground the question-critical scene, whose causal relations with answers are invariant across different interventions on the environment.
With IGV, the VideoQA models are forced to shield the answering process from the negative influence of spurious correlations, which significantly improves the reasoning ability.
Experiments on three benchmark datasets validate the superiority of IGV in terms of accuracy, visual explainability, and generalization ability over the leading baselines.

%% file: work2/sec/1_introduction_new.tex
\section{Introduction}
   \label{w2-sec:intro}

Video Question Answering (VideoQA) \cite{fan2019heterogeneous} is growing in popularity and importance to interactive AI, such as vision-language navigation for in-home robots and personal assistants \cite{DBLP:conf/cvpr/WangHcGSWWZ19,DBLP:conf/cvpr/AndersonWTB0S0G18}.
It is the task of multi-modal reasoning, which answers the natural language question about the content of a given video.
Clearly, inferring a reliable answer requires a deep understanding of visual scenes, linguistic semantics, and more importantly, the visual-linguistic alignments.

Towards this end, a number of VideoQA models have emerged \cite{fan2019heterogeneous,li2019beyond,DBLP:conf/mm/XuZX0Z0Z17,gao2018motionappearance,dang2021hierarchical}.
Scrutinizing these models, we summarize their common paradigm as a combination of two modules:
(1) video-question encoder, which encapsulates the visual scenes of video and the linguistic semantics of question as representations;
and (2) answer decoder, which exploits these representations to model the visual-linguistic alignment and yield an answer.
Consequently, the criterion of empirical risk minimization (ERM) is widely adopted as the learning objective to optimize these modules --- that is, minimizing the loss between the predictive answer and the ground-truth answer.


\begin{figure}[t]
	\centering
	\subcaptionbox{Local mutual information (LMI) between the ``track'' scene and answers.   \label{w2-fig:lmi-show}}{
		\includegraphics[width=0.7\linewidth]{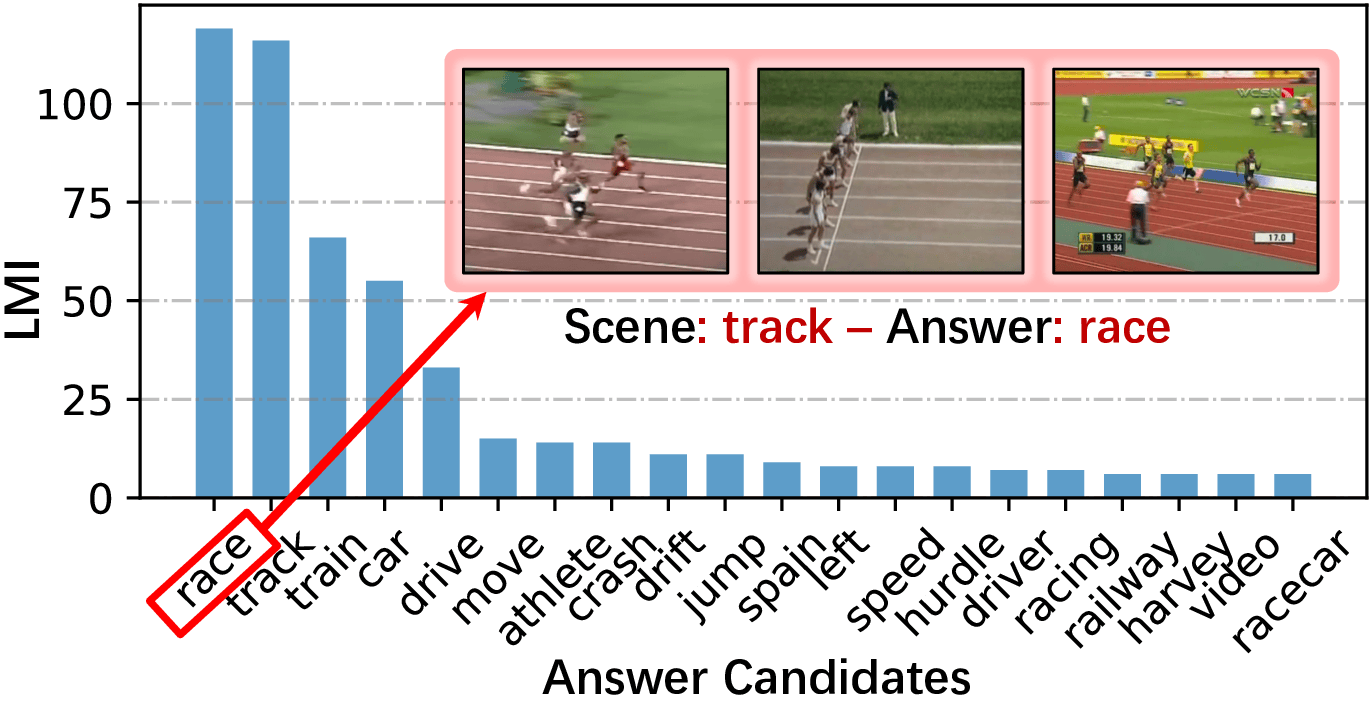}}
	\subcaptionbox{Running example of how the environment ``track'' deviates the answering.   \label{w2-fig:runing_example}}{
	   \vspace{3pt}
		\includegraphics[width=0.7\linewidth]{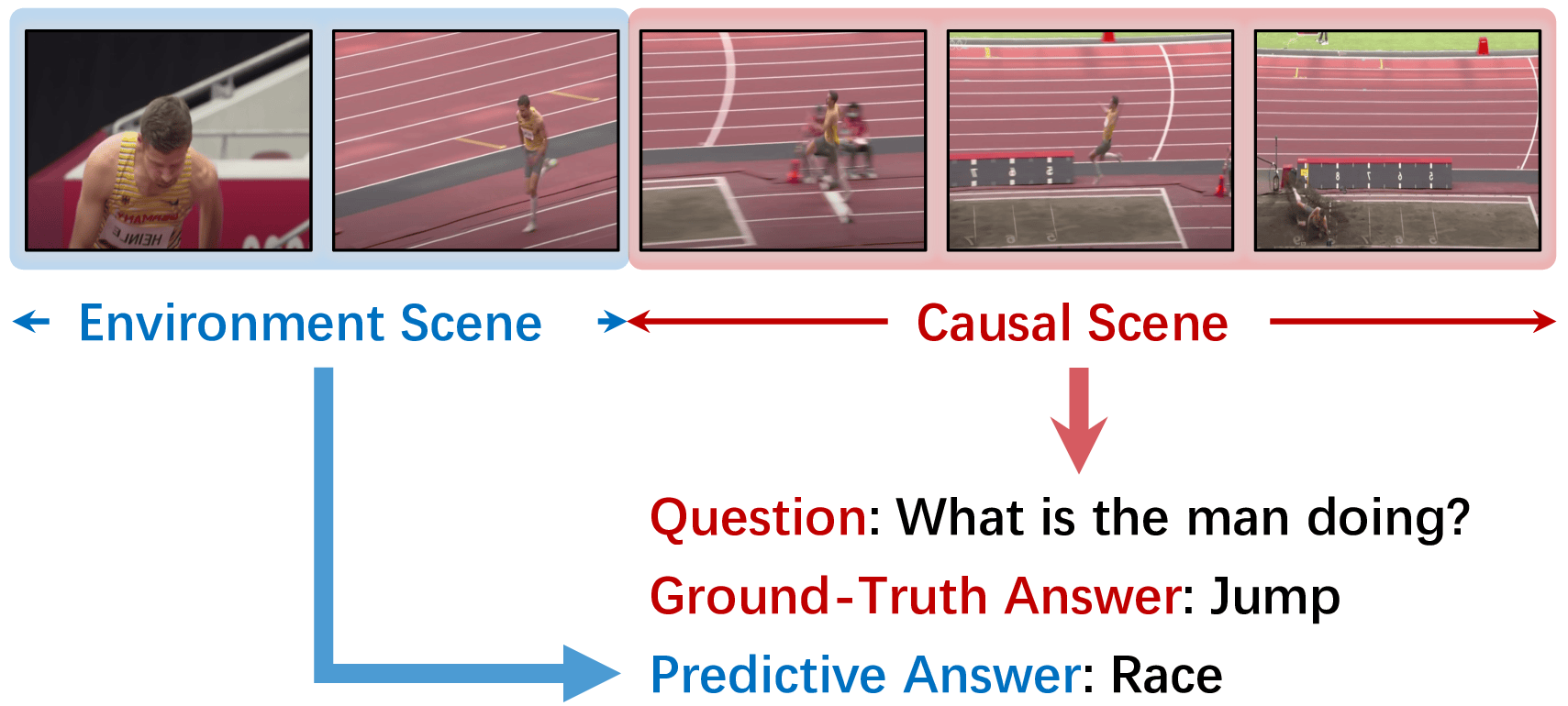}}
	\caption{Running example. (a) Superficial correlations between visual scenes and answers; (b) Suffering from the spurious correlations, VideoQA model fails to answer the question.}
	   \label{w2-fig:intro-example}
\end{figure}



However, the ERM criterion is prone to over-exploiting the superficial correlations between video-question pairs and answers.
Specifically, we use the metric of local mutual information (LMI) \cite{schuster2019debiasing} to quantify the correlations between the ``track'' scene and answers.
As    \cref{w2-fig:lmi-show} shows, most videos with ``track'' scene are associated with the ``race'' answer.
Instead of inspecting the visual-linguistic alignments (\ie which scene is critical to answer the question), ERM blindly captures all statistical relations.
As    \cref{w2-fig:runing_example} shows, it makes VideoQA model naively link the ``track''-relevant videos with the strongly-correlated ``race'' answer, instead of the gold ``jump'' answer.
Taking a causal look \cite{pearl2000causality, pearl2016causal} at VideoQA (see Section    \cref{w2-sec:causal-view}), we partition the visual scenes into two parts: (1) causal scene, which holds the question-critical information, and (2) its complement, the environment, which is irrelevant to the answer.
We scrutinize that the environment is spuriously correlated with the answer, thus ERM hardly differentiates the effects of causal and environment scenes on the answer.
Worse still, the unsatisfactory reasoning obstacles the VideoQA model to own the intriguing properties:
\begin{itemize}[leftmargin=*]
\setlength\itemsep{-2pt}
    \item \textbf{Visual-explainability} to exhibit ``Which visual scene are the right reasons for the right answering?'' \cite{DBLP:conf/ijcai/RossHD17,CSS}.
    Taking    \cref{w2-fig:runing_example} as an example to answer ``What is the man doing?'', the model should attend the ``jump'' event present in the last three clips, rather than referring to the ``track'' environment in the first two clips.
    One straightforward solution is ``learning to attend'' \cite{xiao2020visual,xu2015show,vaswani2017attention} to ground some scenes via the attentive mechanism.
    Nonetheless, guided by ERM, such attentive grounding still suffers from the spurious correlations, thus making the highly-correlated environment grounded.

    \item \textbf{Introspective learning} to double-check ``How would the predictive answer change if the causal scenes were absent?''.
    On top of attentive grounding, the model needs to introspect whether the learned knowledge (\ie attended scene) reliably and faithfully reflects the logic behind the answering. Briefly put, it should fail to answer the question if the causal scenes were removed.
    
    \item \textbf{Generalization ability} to enquire ``How would the predictive answer response to the change of spurious correlations?''.
    As spurious correlations poorly generalize to open-world scenarios, the model should instead latch on the causal visual-linguistic relations that are stable across different environments.
\end{itemize}

Inspired by recent invariant learning \cite{arjovsky2020invariant, wang2021causal,REx}, we conjecture that invariant grounding is the key to distinguishing causal scenes from the complemented environment and overcoming these limitations.
By ``invariant'', we mean that the relations between question-critical scenes and answers are invariant regardless of changes in environment.
Towards this end, we propose a new learning framework, \underline{I}nvariant \underline{G}rounding for \underline{V}ideoQA (\textbf{IGV}).
Concretely, it integrates two additional modules into the VideoQA backbone model: a grounding indicator, a scene intervener.
Specifically, the grounding indicator learns to attend the causal scenes for a given question and leaves the rest as the environment.
Then, we collect visual clips from other training videos to compose a memory bank of environment stratification.
For the causal part of interest, the scene intervener conducts the causal interventions \cite{pearl2000causality, pearl2016causal} on its environment --- that is, replace it with the stratification sampled from the memory bank and compose the ``intervened videos''.
After pairing the casual, environment, and intervened scenes with the question, we feed them into the backbone model to obtain the corresponding predictions:
(1) causal prediction, which approaches the gold answer, so as to achieve visual explainability;
(2) environment prediction, which contains no critical clues to the ground-truth answer, thus enforces the backbone model to perform introspective reasoning;
and (3) intervened prediction, which is consistent with the causal prediction across different intervened environments.
Jointly learning these predictions enables the backbone model to alleviate the negative influence of multi-modal data bias.
It is worthwhile emphasizing that IGV is a model-agnostic strategy, which trains the VideoQA backbones in a plug-and-play fashion.

The contribution of this chapter are summarized as follows:
\begin{itemize}[leftmargin=*]
\setlength\itemsep{-2pt}
\item \textbf{Elevating Visual-Explainability and Generalization in VideoQA Models}: Our work underscores the critical role of effectively grounding causal scenes from various environments in enhancing the visual-explainability, generalization, and introspective learning capabilities of Video Question Answering (VideoQA) models. This approach marks a substantial leap in making VideoQA models more intuitive and interpretable, leading to advancements in how these models understand and interact with dynamic visual environments. This contribution is pivotal in paving the way for more sophisticated, user-friendly, and contextually aware VideoQA applications.

\item \textbf{Superiority through Benchmark Datasets}: Through comprehensive experiments on three benchmark datasets (MSRVTT-QA, MSVD-QA, and NExT-QA), we have validated the superiority of the IGV training scheme in enhancing the performance of VideoQA backbones. Notably, IGV demonstrates a significant improvement over state-of-the-art models. This empirical evidence not only proves the efficacy of IGV but also highlights its potential as a transformative tool in the VideoQA field, setting new benchmarks for model performance and reliability.
\end{itemize}

%% file: work2/sec/2_related.tex
\section{Related works}
\label{w2-sec:related}
\noindent\textbf{Video Question Answering (VideoQA).}
Aiming to answer the question in a video scenario, VideoQA is defined as an escalation of imageQA, because the temporal nature of the input has enriched its reasoning process as well as the answer space. Previous efforts towards VideoQA establish their contribution on either a better multi-modal interaction or stronger video representation. Specifically, early studies tend to impose sophisticated cross-modal fusion via attention \cite{zeng2016leveraging,li2019beyond,jiang2020reasoning} or dynamic memory \cite{gao2018motionappearance, DBLP:conf/mm/XuZX0Z0Z17, fan2019heterogeneous}, while more recent approaches perform relation reasoning through visual or textual graph \cite{jiang2020reasoning,huang2020locationaware,park2021bridge}. In addition, current efforts that model video as a hierarchical structure also intrigue wide interest. Among them, HCRN \cite{le2021hierarchical} stack conditional relation blocks in different feature granularity, whereas HOSTR \cite{dang2021hierarchical} employs a spatio-temporal graph for multilevel reasoning. Despite their effectiveness, their visual-explainability still dwells on ERM-guided attention weights, which only reflect the intensity of feature-prediction correlation.

\vspace{5pt}
\noindent\textbf{Invariant Learning.} 
Multi-modal datasets tend to display inherent bias in some forms \cite{DBLP:journals/corr/abs-1811-05013, DBLP:conf/naacl/ThomasonGB19, DBLP:conf/emnlp/RohrbachHBDS18}. In contrast to overarching reality, the collection process \cite{DBLP:conf/cvpr/TorralbaE11, DBLP:conf/naacl/ChaoHS18} degrades its generalization ability by introducing undesirable correlations between the inputs and the ground truth annotations. 

To overcome such correlation, invariant learning is developed to discover causal relations from the causal factors to the response variable, which remains constant across distributions. 
As the most prevailing formulation, IRM \cite{arjovsky2020invariant} promotes this philosophy from feature level to representation level by finding a data representation $\Upsilon$, from which the optimal predictor $\varphi$ can yield the prediction $\Upsilon\circ \varphi$ that is stable across all environments.
In terms of environment acquisition, previous studies either manually partition the training set by prior knowledge \cite{DBLP:conf/cvpr/AndersonWTB0S0G18}, or generate data partition iteratively via adversarial environment inference \cite{DBLP:conf/icml/CreagerJZ21,wang2021causal}. Our method, instead of partitioning the training, assumes no prophets about environments but performs causal intervention to perturb the original distribution. To the best of our knowledge, IGV is the first work that introduces invariant learning as a model-agnostic framework to 
VideoQA.
\vspace{-3pt}

\begin{table}[h]
  \centering
  \caption{Key Related Papers}
    \begin{tabular}{lp{10cm}} 
    \toprule
    Paper & Remark \\
    \midrule
    \cite{fan2019heterogeneous, gao2018motionappearance} & These are classic VideoQA methods, we adopt these work as the backbone of IGV, to show the model-agnostic nature of our design
    \\
    \midrule
    \cite{DBLP:conf/cvpr/AndersonWTB0S0G18,DBLP:conf/icml/CreagerJZ21,wang2021causal} & Adopts a similar Invariant learning philosophy, but they acquire the environment via human prior or adversarial environment inference. In contrast, we assume no prophets about environments but perform causal intervention to perturb the original distribution. \\
    \bottomrule
    \end{tabular}%
\end{table}%

%% file: work2/sec/2_preliminary_new.tex
\section{Preliminaries} \label{w2-sec:preliminary}
In this section, we summarize the common paradigm of VideoQA models.
Throughout the chapter, we denote the random variables and their deterministic values by upper-cased (\eg $V$) and lower-cased (\eg $v$) letters, respectively.

\vspace{5pt}
\noindent\textbf{Modeling}.
Given the video-question pair $(V,Q)$, the primer task of VideoQA is to generate an answer $\hat{A}$ as:
\begin{gather}\label{w2-eq:conventional-modeling}
    \hat{A} = f_{\hat{A}}(V,Q),
\end{gather}
where $f_{\hat{A}}$ is the VideoQA model, which is typically composed of two modules: video-question encoder, and answer decoder.
Specifically, the encoder includes two components:
(1) a video encoder, which encodes visual scenes of the target video as a visual representation, such as motion-appearance memory design \cite{gao2018motionappearance, fan2019heterogeneous}, structural graph representation \cite{jiang2020reasoning, park2021bridge, huang2020locationaware, Wang_2018_ECCV}, hierarchical architecture \cite{le2021hierarchical, dang2021hierarchical};
and (2) a question encoder, which encapsulates linguistic semantics of the question into a linguistic representation, such as global/local representation of textual content \cite{Jiang_Chen_Lin_Zhao_Gao_2020, 2021}, graph representation of grammatical dependencies \cite{park2021bridge}.
On top of these representations, the decoder learns the visual-linguistic alignments to generate the answer.
In particular, the alignments are modeled via cross-modal interaction like graph alignment\cite{park2021bridge}, cross-attention \cite{jiang2020reasoning, li2019beyond,zeng2016leveraging, Jiang_Chen_Lin_Zhao_Gao_2020} and co-memory \cite{gao2018motionappearance}, \etc.

\vspace{5pt}
\noindent\textbf{Learning}.
To optimize these modules, most of the leading VideoQA models \cite{fan2019heterogeneous,gao2018motionappearance,le2021hierarchical,jiang2020reasoning,Jiang_Chen_Lin_Zhao_Gao_2020} cast the multi-modal reasoning problem as a supervised learning task and adopt the learning objective of empirical risk minimization (ERM) as:
\begin{gather}\label{w2-equ:erm-loss}
    \vspace{-30pt}
    \min_{h}\Lapl_{\text{ERM}}(\hat{A}, A),
    \vspace{-30pt}
\end{gather}
where $\Lapl_{\text{ERM}}$ is the risk function to measure the loss between the predictive answer $\hat{A}$ and ground-truth answer $A$, which is usually set as cross-entropy loss \cite{gao2018motionappearance,le2021hierarchical} or hinge loss \cite{fan2019heterogeneous,jiang2020reasoning}.
In essence, ERM encourages these VideoQA modules to capture the statistical correlations between the video-question pairs and answers.

\section{Causal Look at VideoQA}
\label{w2-sec:causal-view}
From the perspective of causal theory \cite{pearl2000causality,pearl2016causal}, we revisit the VideoQA scenario to show superficial correlations between video-question pairs and answers.
We then analyze ERM's suffering from the spurious correlations.

\subsection{Causal Graph of VideoQA}

\begin{figure}[t]
    \centering
    \includegraphics[width=0.7\linewidth]{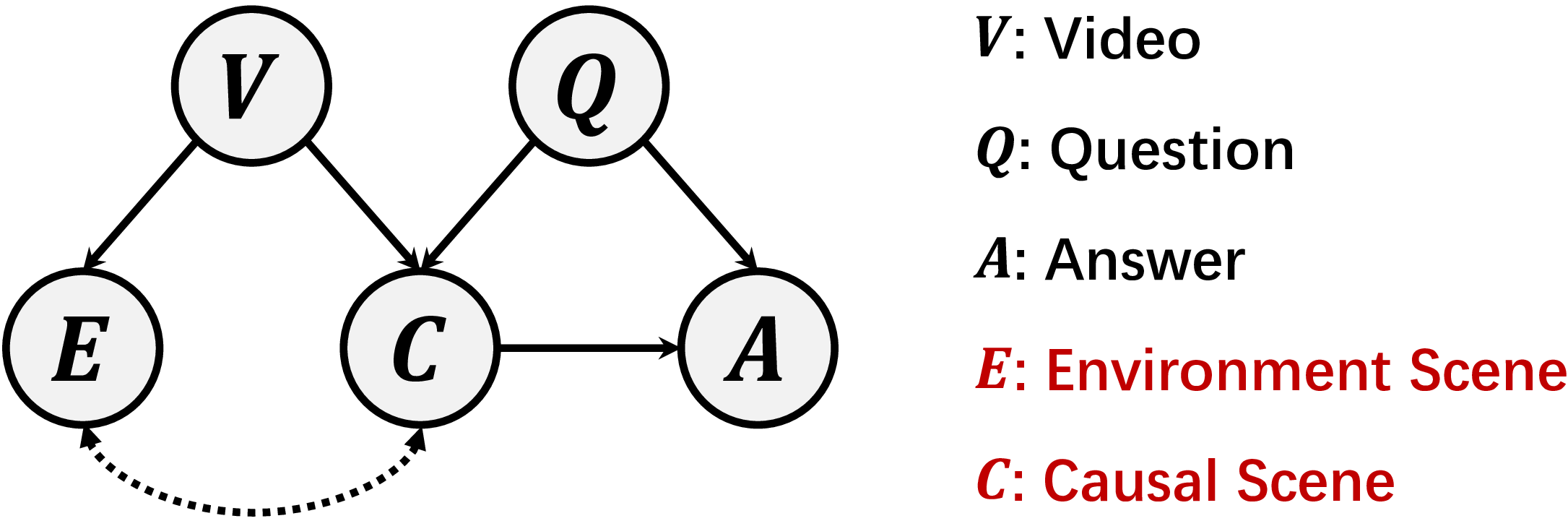}
    \caption{Causal graph of VideoQA}
    \label{w2-fig:scm}
\end{figure}

In general, multiple visual scenes are present in a video.
But only part of the scenes are critical to answering the question of interest, while the rest hardly offers information relevant to the question.
Moreover, the linguistic variations in different questions should activate different scenes of a video.
These facts inspire us to split the video into the causal and environment parts in terms of the question.
Here we use a causal graph \cite{pearl2000causality,pearl2016causal} to exhibit the relationships among five variables:
input video $V$, input question $Q$, causal scene $C$, environment scene $E$, ground-truth answer $A$.
\cref{w2-fig:scm} illustrates the causal graph, where each link is a cause-and-effect relationship between two variables:
\begin{itemize}[leftmargin=*]
    \setlength\itemsep{-2pt}
    \item $C\gets V\to E$. The input video $V$ consists of $C$ and $E$. For example, the video in \cref{w2-fig:runing_example} is the combination of the first two clips (\ie $C$) and the last three clips (\ie $E$).
    \item $V\to C\gets Q$. The causal scene $C$ is conditional upon the video-question pair $(V,Q)$, which distills $Q$-relevant information from $V$. For a given $V$, the variations in $Q$ result in different $C$.
    
    \item $Q\to A\gets C$. The answer $A$ is determined by the question $Q$ and causal scene $C$, reflecting the visual-linguistic alignments. Considering the example in \cref{w2-fig:runing_example} again, $C$ is the oracle scene that perfectly explains why ``jump'' is labeled as the ground truth to answer the question.
    \item $E\dashleftarrow\dashrightarrow C$. The dashed arrow summarizes the additional probabilistic dependencies \cite{pearl2000causality} between $C$ and $E$.
    Such dependencies are usually caused by the selection bias or inductive bias during the process of data collection or annotation \cite{DBLP:conf/cvpr/TorralbaE11,DBLP:conf/naacl/ChaoHS18}. For example, one mostly collects the videos with the ``jump'' events on the ``track''.
    Here we list three typical scenarios: (1) $C$ is independent of $E$ (\ie $E \bot C$); (2) $C$ is the direct cause of $E$ (\ie $C\rightarrow E$), or vise versa (\ie $C\leftarrow E$); (3) $C$ and $E$ have a common cause $T$ (\ie $C\leftarrow T\rightarrow E$). See \cref{w2-app:critical-context-example} for details.
\end{itemize}

\vspace{-15pt}
\subsection{Spurious Correlations}
Taking a closer look at the causal graph, we find that the complemented environment scene $E$ and the ground-truth answer $A$ can be spuriously correlated.
Specifically, as the confounder \cite{pearl2000causality,pearl2016causal} between $E$ and $A$, $Q$ and $V$ open the backdoor paths: $E \gets V \to C \to A$ and $E \gets V \to C \gets Q \to A$, which make $E$ and $A$ spuriously correlated even though there is no direct causal path from $E$ to $A$.
Worse still, $E\dashleftarrow\dashrightarrow C$ can amplify this issue.
Assuming $C \to E$, $C$ becomes an additional confounder to yield another backdoor path $E \gets C \to A$.
Such spurious correlations can be summarized as the probabilistic dependence: $A\not\!\bot E$.

As ERM naively captures the statistical correlations between video-question pairs and answers, it fails to distinguish the causal scene $C$ and environment scene $E$, thus failing to mitigate the negative influence of spurious correlations.
As a result, it limits the reasoning ability of VideoQA models, especially in the following aspects:
(1) visual-explainability to reason about ``Which visual scenes are the supporting evidence to answer the question?'';
(2) introspective learning to answer ``How would the answer change if the causal scenes were absent?'';
and (3) generalization ability to enquire ``How would the answer response to the change of spurious correlations?''.

%% file: work2/sec/3_method_new.tex
\section{Methodology}

We get inspiration from invariant learning \cite{arjovsky2020invariant, wang2021causal,REx} and argue that invariant grounding of causal scenes is the key to reducing the spurious correlations and overcoming the foregoing limitations.
We then present a new learning framework, \underline{I}nvariant \underline{G}rounding for \underline{V}ideoQA (\textbf{IGV}).

\subsection{Invariant Grounding for VideoQA}

Upon closer inspection on the causal graph, we notice that the ground-truth answer $A$ is independent of the visual environment $E$, only when conditioned on the question $Q$ and the causal scene $C$, more formally:
\begin{gather}\label{w2-equ:invariance}
    A\bot E \mid C,Q.
\end{gather}
This probabilistic independence indicates the invariance --- that is, the relations between the $(C,Q)$ pair and $A$ are invariant regardless of changes in $E$.
The causal relationship $Q\rightarrow A \leftarrow C$ is invariant across different $E$.
Taking \cref{w2-fig:runing_example} as an example, if the question and the causal scene (\ie the last three clips) remain unchanged, the answer should arrive at ``jump'', no matter how the environment varies\footnote{Note that the environment substitutes will not involve the question-relevant scenes, in order to avoid creating additional paths from $E$ to $A$.} (\eg substitute the ``track'' clips by the ``cloud''- or ``sea''-relevant ones).
This highlights that the $(C,Q)$ pair is the key to shielding $A$ from the influence of $E$.

\vspace{5pt}
\noindent\textbf{Modeling.}
However, only the $(V,Q)$ pair and $A$ are available in the training set, while neither $C$ nor the grounding function towards $C$ is known.
This motivates us to incorporate visual grounding into the VideoQA modeling, where the grounded scene $\hat{C}$ aims to estimate the oracle $C$ and guide the prediction of answer $\hat{A}$.
More formally, instead of the conventional modeling (\cf \cref{w2-eq:conventional-modeling}), we systematize the modeling process as:
\begin{gather}
    \hat{C}=f_{\hat{C}}(V,Q),\quad \hat{A}=f_{\hat{A}}(\hat{C}, Q),
\end{gather}
where $f_{\hat{C}}$ is the grounding model, and $f_{\hat{A}}$ is the VideoQA model that relies on the $(\hat{C},Q)$ pair instead.
See \cref{w2-app:implementation} for our implementations of $f_{\hat{C}}$ and $f_{\hat{A}}$.

\vspace{5pt}
\noindent\textbf{Learning.}
Nonetheless, simply integrating visual grounding with the VideoQA model falls into the ``learning to attend'' paradigm, which still suffers from the spurious correlations and erroneously attends to the environment scenes as $\hat{C}$.
To this end, we exploit the invariance property of $C$ (\cf \cref{w2-equ:invariance}) and reformulate the learning objective of invariant grounding as:
\begin{gather}
    \min_{f_{\hat{A}},f_{\hat{C}}}\Lapl_{\text{IGV}}(\hat{A}, A),\quad\text{s.t.}~~A\bot \hat{E}\mid\hat{C},Q,
\end{gather}
where $\Lapl_{\text{IGV}}$ is the loss function to our IGV; $\hat{E}=V\setminus\hat{C}$ is the complemented environment of $\hat{C}$.
In the next section, we will elaborate how to implement $\Lapl_{\text{IGV}}$ and achieve invariant grounding.

\subsection{IGV Framework}
\label{w2-sec:implementations}

\begin{figure}[t]
    \centering
    \includegraphics[width=0.8\linewidth]{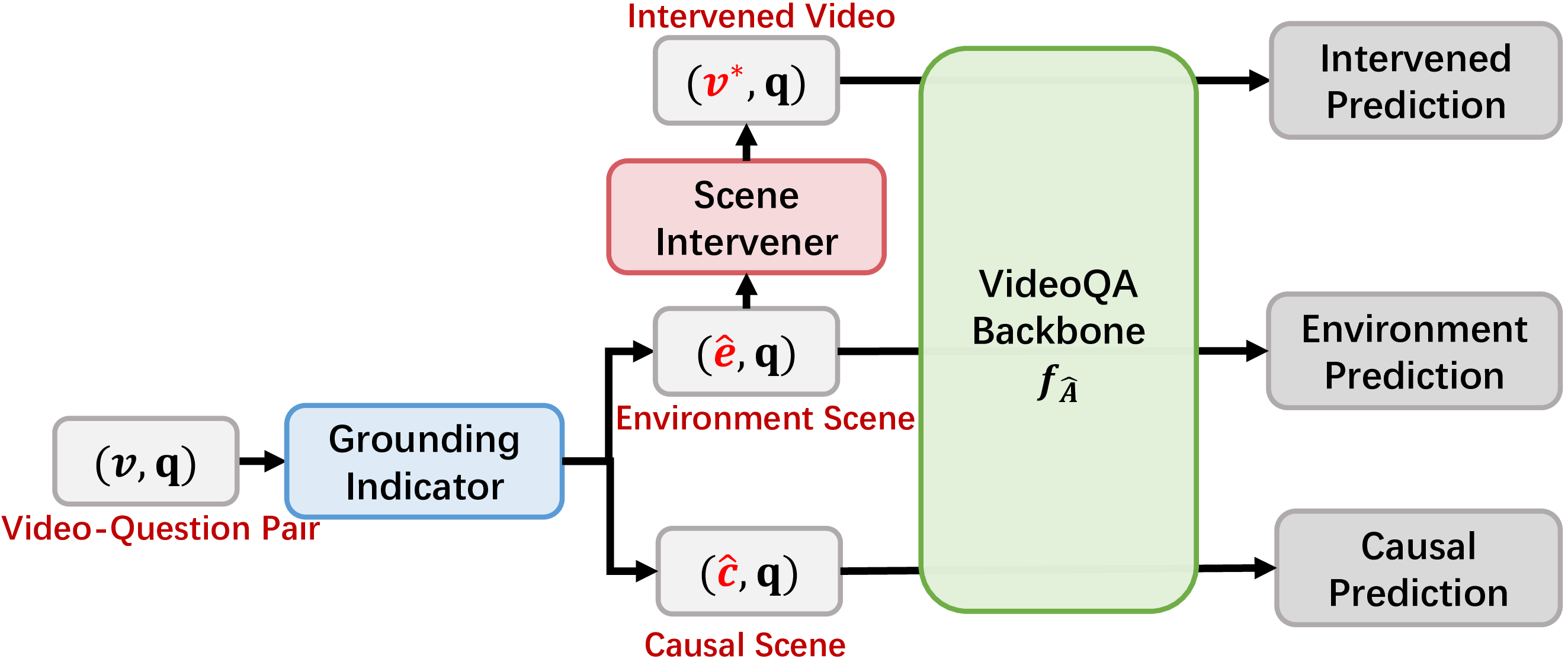}
    \caption{Overview of our IGV framework.}
    \label{w2-fig:overview}
\end{figure}

\cref{w2-fig:overview} displays our IGV framework, which involves two additional modules, the grounding indicator and scene intervener, beyond the VideoQA backbone model $f_{\hat{A}}$.

\subsubsection{Grounding Indicator}
For a video-question pair instance $(v,q)$, at the core of the grounding indicator is to split the video instance $v$ into two parts, $\hat{c}$ and $\hat{e}$, according to the question $q$.
Towards this end, it first employs two independent LSTMs \cite{10.1162/neco.1997.9.8.1735} to encode the visual and linguistic characteristics of $v$ and $q$, respectively:
\begin{gather}\label{w2-equ:vl-lstm}
    \Mat{v}_{g},\Mat{v}_{l} = \text{LSTM}_{1}(v),\quad \Mat{q}_{g},\Mat{q}_{l} = \text{LSTM}_{2}(q),
\end{gather}
where the features of $v$ are $K$ fixed visual clips, while $q$ is associated with $L$ language tokens;
LSTM$_1$ outputs $\Mat{v}_{l}\in\Space{R}^{K\times d}$ as the local representations of clips, and yields the last hidden state $\Mat{v}_{g}\in\Space{R}^{d}$ as the global representations of the holistic video.
Analogously, LSTM$_2$ generates $\Mat{q}_{l}\in\Space{R}^{L\times d}$ as the local representations of tokens, and makes the last hidden state $\Mat{q}_{g}\in\Space{R}^{d}$ represent the question holistically, here $d$ is the hidden dimension.

Upon these representations, the attention scores are constructed to indicate the importance of each visual clip.
Here we devise $\Mat{p}_{\hat{c}}\in\Space{R}^{K}$ to exhibit the probability of each clip belonging to the causal scene $\hat{c}$, while $\Mat{p}_{\hat{e}}\in\Space{R}^{K}$ is in contrast to $\Mat{p}_{\hat{c}}$ to show how likely each clip composes the environment $\hat{e}$.
The formulations are as follows:
\begin{align}
    \Mat{p}_{\hat{c}} &= \text{Softmax}(\text{MLP}_{1}(\Mat{v}_{l})\cdot\Trans{\text{MLP}_{2}(\Mat{q}_{g})}),\\
    \Mat{p}_{\hat{e}} &= \text{Softmax}(\text{MLP}_{3}(\Mat{v}_{l})\cdot\Trans{\text{MLP}_{4}(\Mat{q}_{g})}),
\end{align}
where four multilayer perceptrons (MLPs) are employed to distill useful information:
$\text{MLP}_{1}(\Mat{v}_{l}),\,\text{MLP}_{3}(\Mat{v}_{l})\in\Space{R}^{K\times d'}$,  $\text{MLP}_{2}(\Mat{q}_{g}),\,\text{MLP}_{4}(\Mat{q}_{g})\in\Space{R}^{d'}$
; $d'$ is the feature dimension.
However, as the soft masks make $\hat{c}$ and $\hat{e}$ overlapped, the attentive mechanism cannot shield the answering from the influence of the environment.
Hence, the grounding indicator produces discrete selections instead to make $\hat{c}$ and $\hat{e}$ disjoint.
Nonetheless, simple sampling or selection is not differentiable.
To achieve differentiable discrete selection, we apply Gumbel-Softmax \cite{DBLP:conf/iclr/JangGP17}:
\begin{gather}
    \Mat{I} = \text{Gumbel-Softmax}([\Mat{p}_{\hat{c}},\Mat{p}_{\hat{e}}]),
\end{gather}
where Gumbel-Softmax is built upon the concatenation of $\Mat{p}_{\hat{c}}$ and $\Mat{p}_{\hat{e}}$ (\ie $[\Mat{p}_{\hat{c}},\Mat{p}_{\hat{e}}]\in\Space{R}^{K\times 2}$), and outputs the indicator vector $\Mat{I}\in\Space{R}^{K\times2}$ whose first and second column indexes $\hat{c}$ and $\hat{e}$ over k clips, respectively. 
As such, we can devise $\hat{c}$ and $\hat{e}$ as follows:
\begin{gather}
    \hat{c} = \{I_{k0}\cdot v_{k}|I_{k0}=1\},\quad \hat{e} = \{I_{k1}\cdot v_{k}|I_{k1}=1\},
\end{gather}
where $I_{0k}$ and $I_{1k}$ suggests that the $k$-th clip belongs to the causal and environment scenes, respectively. Notablely, since  the indicator vector $\Mat{I}$ is output of the Gumbel-Softmax, thus, for each raw of $\Mat{I}$, only one of two column are set to 1, and the other is 0.


\subsubsection{Scene Intervener}
\begin{figure}[t]
    \centering
    \includegraphics[width=0.9\linewidth]{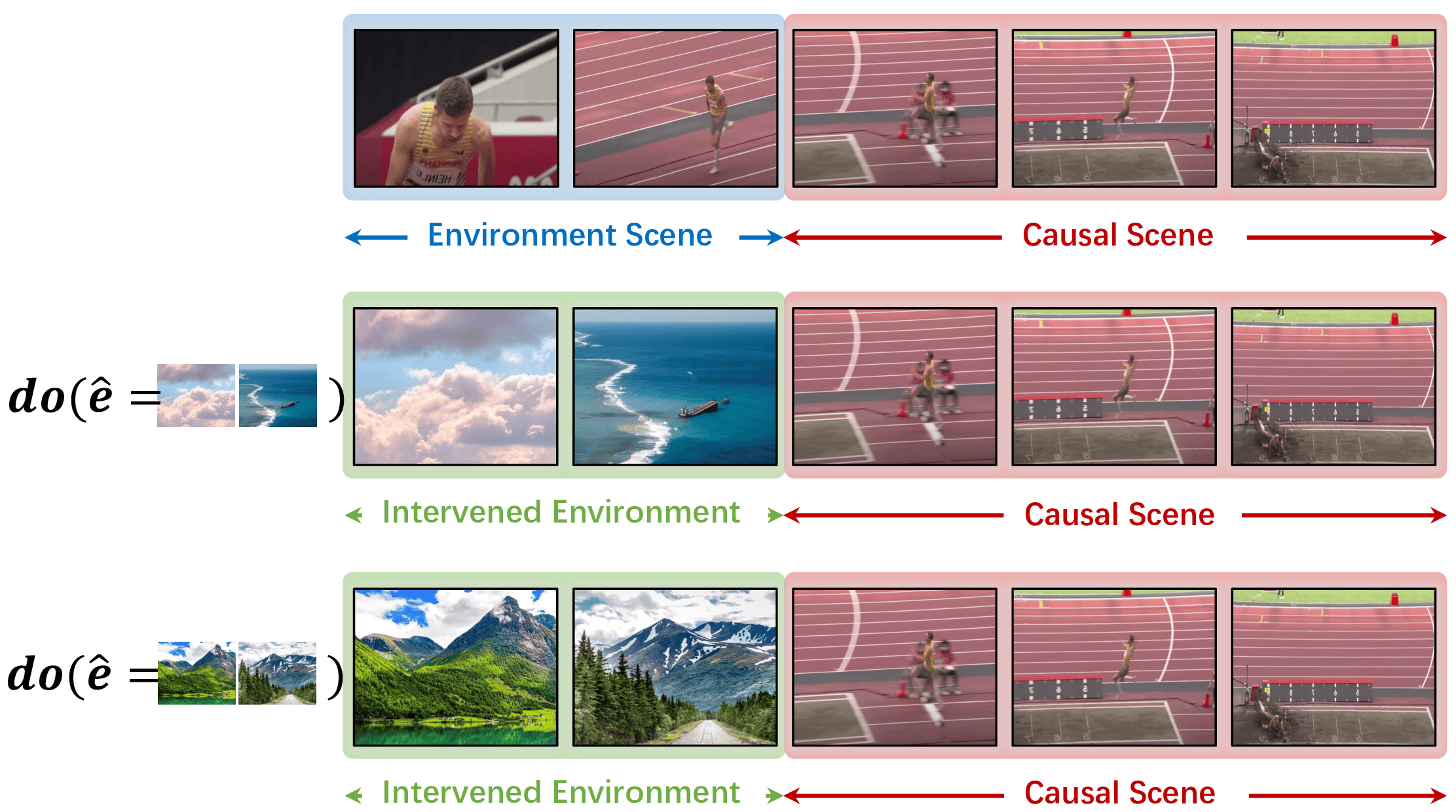}
    \caption{Illustration of interventional distribution.}
    \label{w2-fig:intervene_exp}
\end{figure}

It is challenging to learn the grounding indicator, owing to the lack of supervisory signals of clip-level importance.
To remedy this issue, we propose the scene intervener, which preserves the estimated causal scene $\hat{c}$ but intervenes the estimated environment $\hat{e}$ to create the ``intervened videos'', as \cref{w2-fig:intervene_exp} shows.

Specifically, for the observed video-question pairs during training, the scene intervener first collects visual clips from other training videos as a memory bank of environment stratification, $\hat{\Set{E}}=\{\hat{e}\}$.
Then, for the video of interest $v=\hat{c}\cup\hat{e}$, the intervener conducts causal interventions \cite{pearl2000causality,pearl2016causal} on its $\hat{e}$ --- that is, random sample a environment stratification $\hat{e}^{*}\in\hat{\Set{E}}$ to replace $\hat{e}$ and combine it with $\hat{c}$ at hand as a new video $v^{*}=\hat{c}\cup\hat{e}^{*}$.

It is worthwhile mentioning that, distinct from the current invariant learning studies \cite{arjovsky2020invariant, wang2021causal,REx} that only partition the training set into different environments, our scene intervener exploits the interventional distributions \cite{DBLP:conf/aaai/TianKP06} instead.
The interventional distribution (\ie, the videos with the same interventions) can be viewed as one environment.

\subsubsection{VideoQA Backbone Model}
Inspired by \cite{jiang2020reasoning}, we design a simple yet effective architecture
as our backbone predictor, where the video encoder is shared with the grounding indicator.
It embodies convolutional graph networks (GCN) to propagate clip-level visual messages, then integrates cross-modal fused local and global representations via BLOCK fusion \cite{BenYounes_2019_AAAI}.

\subsubsection{Joint Training}
For a video-question pair instance $(v,q)$, we have established the causal scene $\hat{c}$, environment scene $\hat{e}$, and intervened video $v^{*}$ via the grounding indicator and scene intervener.
Pairing them with $q$ synthesizes three new instances: $(\hat{c},q)$, $(\hat{e},q)$, $(v^{*},q)$.
We next feed these instances into the backbone VideoQA model $f_{\hat{A}}$ to obtain three predictions:
\begin{itemize}[leftmargin=*]
    \setlength\itemsep{-2pt}
    \item \textbf{Causal prediction}. As the causal scene $\hat{c}$ is expected to be sufficient and necessary to answer the question $q$, we leverage its predictive answer $f_{\hat{A}}(\hat{c},q)$ to approach the ground-truth answer $a$ solely:
    \begin{gather}
        \Lapl_{\hat{c}} = \text{XE}(f_{\hat{A}}(\hat{c},q), a),
    \end{gather}
    where XE denotes the cross-entropy loss.
    
    \item \textbf{Environment prediction}. As no critical clues should exist in the environment scene $\hat{e}$ to answer the question $q$, we encourage its predictive answer $f_{\hat{A}}(\hat{e},q)$ to evenly predict all answers.
    This uniform loss is formulated as:
    \begin{gather}
        \Lapl_{\hat{e}} = \text{KL}(f_{\hat{A}}(\hat{e},q), u),
    \end{gather}
    where KL denotes KL-divergence, and $u$ is the uniform distribution over all answer candidates.
    
    \item \textbf{Intervened prediction}. According to the invariant constraint (\cf \cref{w2-equ:invariance}), the causal relationship between the causal scene and the answer is stable across different environments.
    To parameterize this constraint, we enforce all $v$'s intervened versions to hold the consistent predictions:
    \begin{gather}
        \Lapl_{v^{*}} = \Space{E}_{\hat{e}^{*}\in\Set{E}}(\text{KL}(f_{\hat{A}}(v^{*},q), f_{\hat{A}}(\hat{c},q))).
    \end{gather}
    \vspace{-25pt}
\end{itemize}
Aggregating the foregoing risks, we attain the learning objective of IGV:
\begin{gather}
    \Lapl_{\text{IGV}} = \Space{E}_{(v,q,a)\in\Set{O}^{+}}\Lapl_{\hat{c}} + \lambda_{1}\Lapl_{\hat{e}} + \lambda_{2}\Lapl_{v^{*}},
\end{gather}
where $\Set{O}^{+}$ is the training set of the video-question pair $(v,q)$ and the ground-truth answer $a$; $\lambda_{1}$ and $\lambda_{2}$ are the hyper-parameters to control the strengths of invariant learning.
Jointly learning these predictions enables the VideoQA backbone model to uncover the question-critical scene, so as to mitigate the negative influence of spurious correlations between the question-irrelevant environment scene and answer.
In the inference phase, we use the causal prediction $f_{\hat{A}}(\hat{c},q)$ to answer the question.

%% file: work2/sec/4_results.tex
\section{Experiments}
We conduct extensive experiments to answer the following research questions:
\begin{itemize}[leftmargin=*]
\setlength\itemsep{-.20em}
    \item \textbf{RQ1:} How effect is IGV in training VideoQA backbones as compared with the State-of-the-Art (SoTA) models?
    \item \textbf{RQ2:} How do the loss component and feature setting affect the performance?
    \item \textbf{RQ3:} What are the learning patterns and insights of IGV training?
\end{itemize}
\noindent\textbf{Settings:}
We compare IGV with seven baselines from families of Memory, GNN and Hierarchy on three VideoQA datasets: 
\textbf{NExT-QA} \cite{next-qa} which features causal and temporal action interactions among multiple objects. It contains about 47.7K manually annotated questions for multi-choice QA collected from 5.4K videos with an average length of 44s.
\textbf{MSVD-QA} \cite{DBLP:conf/mm/XuZX0Z0Z17} and \textbf{MSRVTT-QA} \cite{DBLP:conf/mm/XuZX0Z0Z17} are two prevailing datasets that focus on the description of video elements. They respectively contain 50K and 243K QA pairs with open answer space over 1.6K and 6K. For all three datasets, we follow their official data splits for experiments and report accuracy as evaluation metric. 

\vspace{5pt}
\noindent\textbf{Implementation Details:}
For the visual feature, we follow previous works \cite{le2021hierarchical, jiang2020reasoning, next-qa} and extract video feature as a combination of motion and appearance representations by using the pre-trained 3D ResNeXt-101 and ResNet-101, respectively. Specifically, each video is uniformly sampled into $K$=16 clips, where each clip is represented by a combined feature vector $v_{k}^{d_v}$, where $d_v$ equals 4096. Similar to \cite{hqga}, we obtain the contextualized word representation from the finetuned BERT model, and the feature dim $d_q$ is 768.  For our model, the dimension of the hidden states are set to $d$ = 512, and the number of graph layers in
IGV backbone predictor is 2. During training, IGV is optimized by Adam optimizer with the initial learning rate of 1e-4, which will be halved if no validation improvements in 5 epochs. We set the batch size to 256 and a maximum of 60 epochs. 

\vspace{5pt}

\subsection{Main Results (RQ1)}\label{w2-section:rq1}
\subsubsection{Comparisons with SoTA Methods}
\input{work2/tab/sota-next}
\input{work2/tab/sota-ms}

As shown in \cref{w2-tab:sota-next} and \cref{w2-tab:sota-ms}, our method outperforms SoTAs with questions of all sub-types surpassing their competitors. Specifically, we have two major observations: 

    First, on NExT-QA, IGV gains remarkable improvement on \textit{temporal} type (+1.65\%), the underlying explanation are: 1) \textit{temporal} question generally corresponds to video content with a longer time span, which requires more introspective grounding of the causal scene. Fortunately, IGV's design philosophy comfort such demand by wiping out the trivial scenes, which takes up a huge proportion in \textit{temporal} type, thus making the predicting faithful.  2) \textit{temporal} questions tend to include a temporal indicative phase (\eg ``\textit{at the end of the video}") that serves as a strong signal for grounding indicator to locate the target window.
    
    Second, along with \textit{descriptive} questions on NExT-QA, the result on MSRVTT-QA and MSVD-QA (both emphases on question of \textit{descriptive} type) demonstrate the superiority in \textit{descriptive} question across all three datasets (+1.85\% on NExT-QA, +1.4\% on MSRVTT-QA and MSVD-QA). Such improvement is underpinned by logic that answering \textit{descriptive} questions requires scrutiny on the scene of interest, instead of a holistic view of the entire sequence. Accordingly, targeted prediction inducted by IGV concentrates reasoning on keyframes, thus achieving better performance. 
    As a consequence, such improvement strongly validates that IGV generalizes better over various environments.
    \vspace{-10pt}

 
\subsubsection{Backbone Agnostic}\label{w2-section:model-agnostic}
By nature, our IGV principle is orthogonal to backbone design, thus helping to boost any off-the-shelf SoTAs without compromising the underlying architecture.
We therefore experimentally testify the generality and effectiveness of our learning strategy by marrying the IVG principle with methods from two different categories: Co-Mem \cite{gao2018motionappearance} from memory-based architecture and HGA \cite{jiang2020reasoning} from Graph-based method. 
\cref{w2-tab:model-agnostic} shows the results on three backbone predictors (including ours). Our findings are:

    \textbf{1. Better improvement for severe bias.}
    We notice that the improvement on MSVD-QA (+3.1\%$\sim$4.7\%) is considerably larger than that on MSRVTT-QA (+1.4\%$\sim$2\%). Such expected discrepancy is caused by the fact that, although identical in question type, MSRVTT-QA is almost 5 times larger than MSVD-QA (\#QA pairs 243K \textit{vs} 50K). As a result, the baseline model trained on MSRVTT-QA is gifted with better generalization ability, whereas the model on MSVD-QA still suffers from severe shortcut correlation. For the same reason, the IGV framework achieves much better improvement in the severe-shortcut situation (\eg MSVD-QA). Such discrepancy validates our motivation of eliminating statistic dependency.
    
    \textbf{2. Constant improvement for each method.} 
    Through row-wise inspection, we notice that for each benchmark, IGV can bring considerable improvement across different backbone models (+3.1\%$\sim$4.7\% for MSVD-QA, +1.4\%$\sim$2\% for MSRVTT-QA). Such stable enhancement strongly verifies our modal-agnostic statement.
\input{work2/tab/model-agnostic}


\begin{figure*}[t]
  \centering
  \includegraphics[width=0.99\textwidth]{ 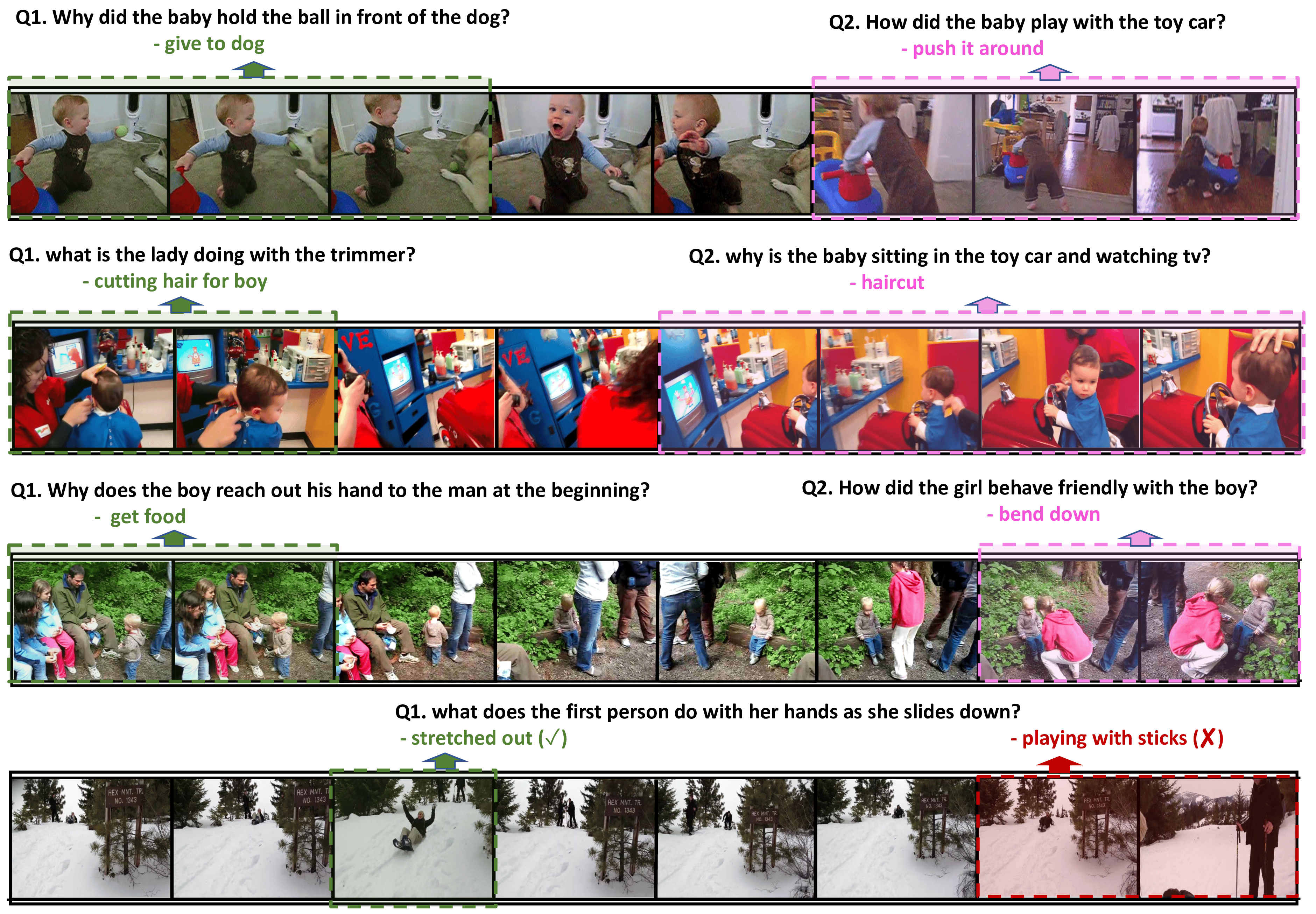}
  	\vspace{-0.07in}
  \caption{Visualization of grounding result on the correct prediction cases from NExT-QA. Each video comes with two questions that demand causal scene of different time span.  The green and pink windows indicate the causal scenes for the corresponding questions. The last row demonstrate a failure case.}
  \label{w2-fig:case-study}
 	\vspace{-13pt}
\end{figure*}

\subsection{In-Depth Study (RQ2)}\label{w2-section:rq2}
\subsubsection{Contributions of Different Loss Components}

An in-depth comprehension of IGV framework requires careful scrutiny on its components. Alone this line, we exhaust the combination of IGV loss components and design three variants: $\Lapl_{\hat{c}},$ $\Lapl_{\hat{c}}+\Lapl_{\hat{e}}$ and $\Lapl_{\hat{c}}+\Lapl_{v^{*}}$.
\begin{itemize}[leftmargin=*]
\setlength\itemsep{-.20em}
    \item Using $\Lapl_{\hat{c}}$ solely, which can be viewed as a special case of ERM-guided attention, hardly outperforms the baseline, because grounding indicators can not identify the causal scene without clip-level supervision. Such an expected result reflects our motivation in interventional design. 
    \item $\Lapl_{\hat{c}}+\Lapl_{\hat{e}}$ and $\Lapl_{\hat{c}}+\Lapl_{v^{*}}$ matched equally in accuracy that consistently surpass baseline and $\Lapl_{\hat{c}}$ in all cases. Such progress shows the effectiveness of intervention strategy and introspective regularization imposed on environment.
    \item In all cases, $\Lapl_{\hat{c}}+\Lapl_{\hat{e}}+\Lapl_{v^{*}}$ further boosts the performance significantly, which shows $\Lapl_{\hat{e}}$ and $\Lapl_{v^{*}}$ contribute in different aspects and their benefits are mutually reinforcing. 
    \vspace{-25pt}
\end{itemize}
\input{work2/tab/ablation-loss}

\subsubsection{Study of Feature}
By convention, we study the effect of the input condition by ablation on the visual feature. Particularly, we denote \textbf{APP} for tests that adopt only appearance feature as input and  \textbf{MOT} for tests that utilize motion feature alone. Figure \cref{w2-fig:feat_analysis} delivers results on two benchmarks, where we observe:

First, IGV can improve the performance significantly for all input conditions, which generalizes the effectiveness of our framework. Similar to \cref{w2-tab:model-agnostic}, the improvement on MSVD-QA is larger than that on MSRVTT-QA, which solidifies our finding in Section \cref{w2-section:model-agnostic}.

Second, compared to motion feature, IGV brings distinctively larger improvements using appearance feature. Considering the causal nature of IGV, we conclude that static correlation tends to bias more in appearance feature. 
\vspace{-5pt}

\subsubsection{Study of Hyper-parameter}

\begin{figure}[t]
    \centering
    	\subcaptionbox{\label{w2-fig:feat_analysis}}{
		\includegraphics[width=0.45\linewidth]{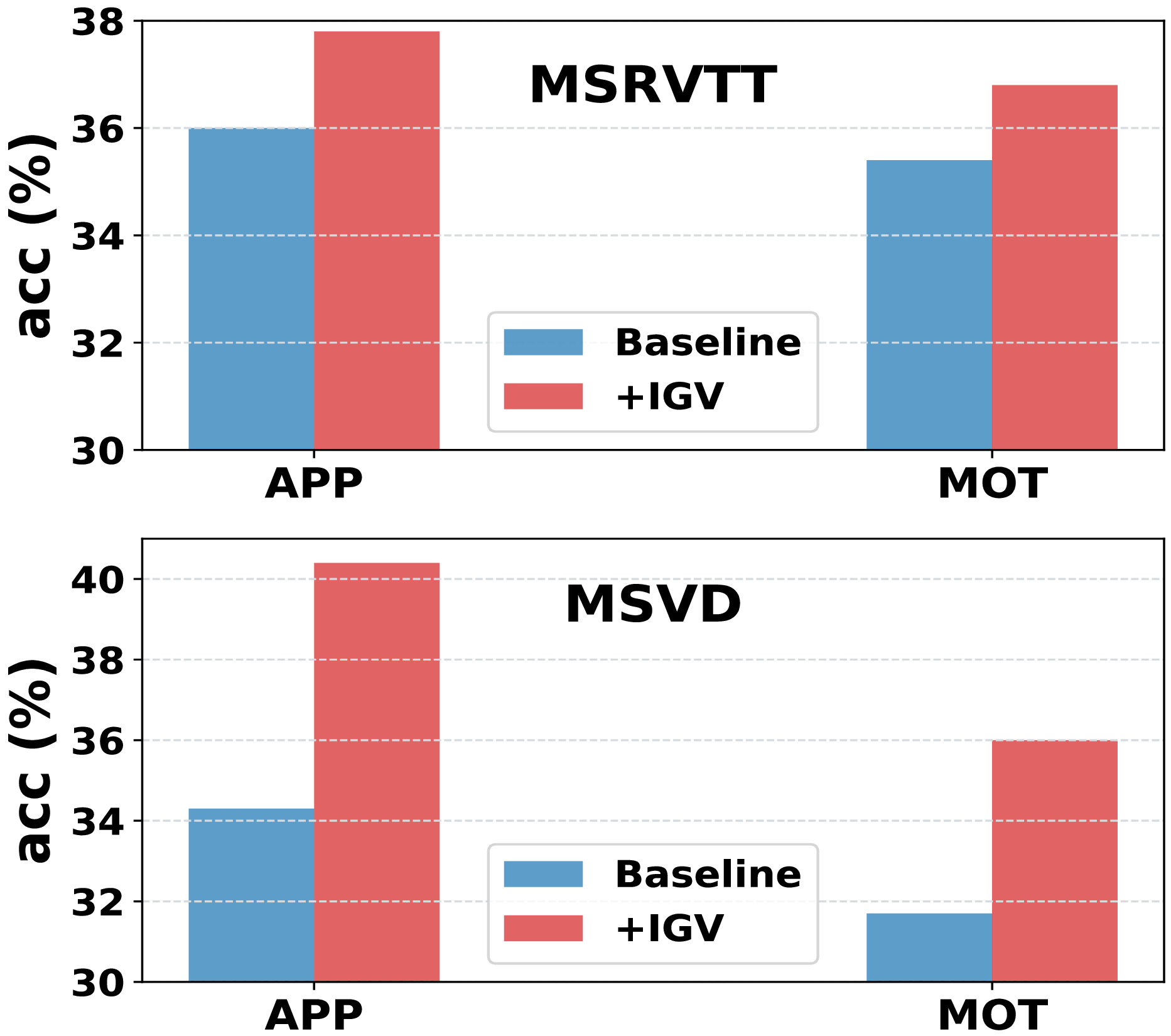}}
		\subcaptionbox{\label{w2-fig:lambda}}{
	    \includegraphics[width=0.45\linewidth]{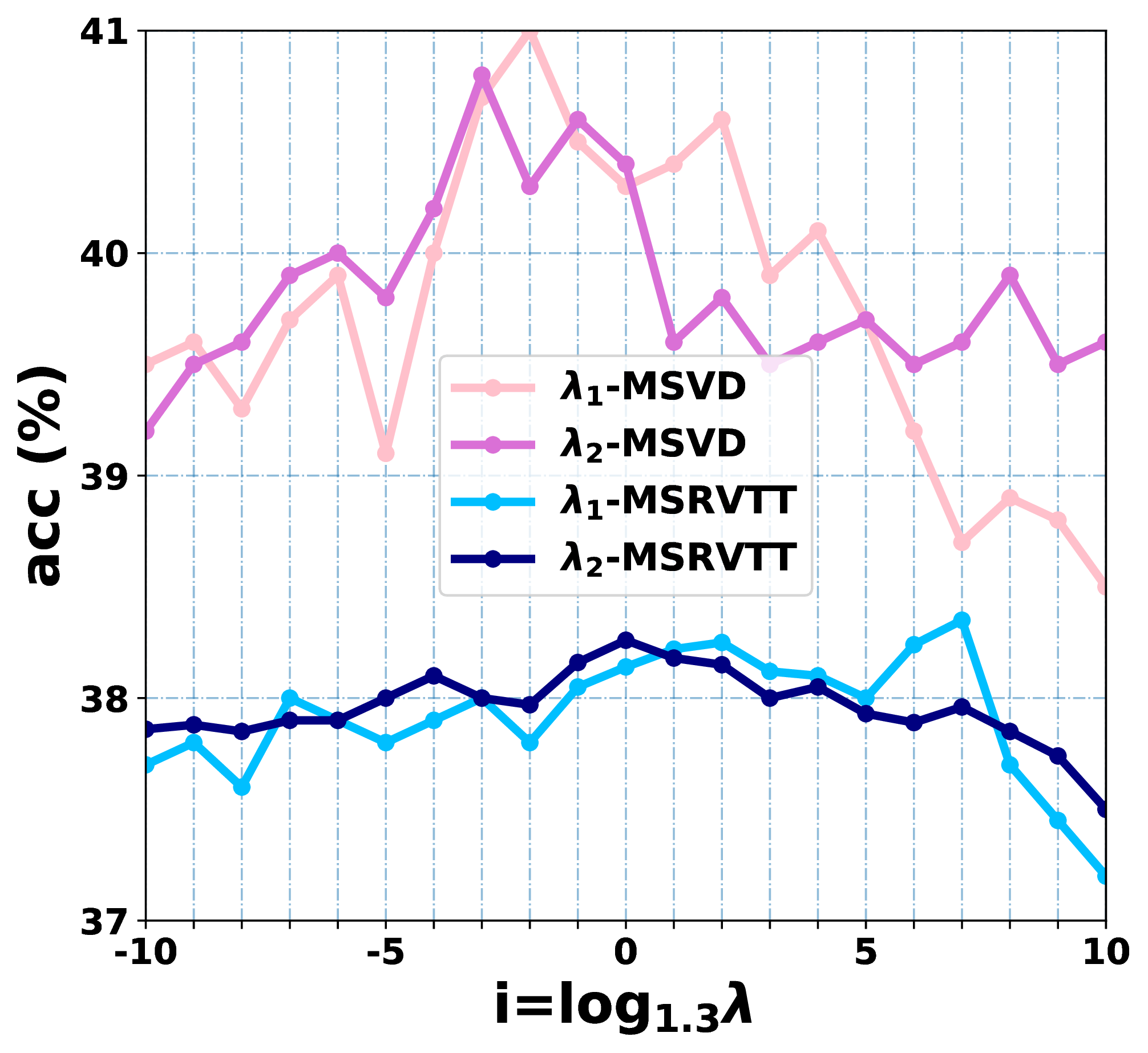}}
    \caption{(a) Study of feature setting.; (b) Study of $\lambda_1$ and $\lambda_2$.}
\end{figure}

To validate the sensitivity of IGV against the hyper-parameters, we conduct experiments with variations of $\lambda_1$ and $\lambda_2$ on two datasets. Without loss of generalization, we tune $\lambda_1$ ($\lambda_2$) as sample of $\left\{ 1.3^{i} \mid -10\le i\le 10, \; i \in\mathbb{Z}  \right\}$,  while keeping the $\lambda_2$ ($\lambda_1$) as 1. According to Figure \cref{w2-fig:lambda}, we have follow observations:

For MSVD-QA, we observe consistent peaks around 0.8 for both $\lambda_1$ and $\lambda_2$. Comparatively, fluctuation on MSRVTT-QA is more moderate, where tuning on $\lambda$ only causes a 1.5\% difference in their accuracy. It's noteworthy that IGV outperforms the baseline by a large margin (+3\%) under all tests, which indicates IGV's robustness against variation of hyper-parameters. Additionally, comparing to $\lambda_2$, IGV is more sensitive to $\lambda_1$. Typically, the performance suffers a drastic degradation for $\lambda_1$ larger than 5 on both datasets. Whereas $\lambda_2$ maintain above 39\% (MSVD-QA) and 37.5\% (MSRVTT-QA) for all tests.

\subsection{Qualitative analysis (RQ3)}\label{w2-section:rq3}
As mentioned in Section \cref{w2-sec:intro} , IGV is empowered with visual-explainability, and is apt to account for the right scene for its prediction.  Following this essence, we grasp the learning insight of IGV by inspecting some correct examples from the NExT-QA dataset and show the visualization in Figure \cref{w2-fig:case-study}. Concretely, each video comes with two questions that emphasize different parts of the video. We notice that, even for the same video, our grounding window is question-sensitive to enclose the explainable content with correct prediction. Nonetheless, we also observe results of \textit{insufficient-grounding} on the 
third row Q2, where the girl starts to bend down before the last two frames, even though the most informative last two frames are encompassed. In the last row, we show a failure case, the model ground wrong frames as the causal part, thus making an erroneous prediction.


%% file: work2/tab/sota-next.tex
\setlength{\tabcolsep}{9pt}
\begin{table}[t]
\small
  \centering
  \caption{Comparison of accuracy on NExT-QA test set.  The \textbf{best} and \underline{second-best} results are highlighted.}
    \vspace{-5pt}
    \begin{tabular}{l|cccc}
    \toprule
    Models & Causal     & Temp     & Descrip     & All \\
    \midrule
    \midrule
    Co-Mem \cite{gao2018motionappearance} & 45.85 & \underline{50.02} & 54.38 & 48.54 \\
    HCRN \cite{le2021hierarchical} & 47.07 & 49.27 & 54.02 & 48.82 \\
    HME  \cite{fan2019heterogeneous} & 46.76 & 48.89 & 57.37 & 49.16 \\
    HGA \cite{jiang2020reasoning}  & \underline{48.13} & 49.08 & \underline{57.79} & \underline{50.01} \\
    \midrule
    IGV(Ours)  & \textbf{48.56} & \textbf{51.67} & \textbf{59.64} & \textbf{51.34} \\
    Abs. Improve & +0.43 & +1.65 & +1.85 & +1.33 \\
    \bottomrule
    \end{tabular}
  \label{w2-tab:sota-next}%
\end{table}%

%% file: work2/tab/sota-ms.tex
\setlength{\tabcolsep}{5pt}
\begin{table}[t]
\small
  \centering
  \caption{Comparison of accuracy on MSVD-QA and MSRVTT-QA test set. "$\dagger$" indicates the result is re-implementation with the publicly available code}
  \resizebox{0.7\linewidth}{!}{
    \begin{tabular}{ll|cc}
    \toprule
   \multicolumn{2}{c|}{Models} & MSVD-QA  & MSRVTT-QA \\
    \midrule
    \midrule
    \multirow{3}{*}{Memory}
   &  AMU \cite{DBLP:conf/mm/XuZX0Z0Z17}   & 32.0    & 32.0  \cr
   &    HME \cite{fan2019heterogeneous}  & 33.7  & 33.0 \cr
   &    Co-Mem$\dagger$ \cite{gao2018motionappearance} & 34.6  & 35.3 \cr
    \midrule
    \multirow{2}{*}{GNN}
   & HGA$\dagger$ \cite{jiang2020reasoning}  & 35.4  & 36.1 \cr
   & B2A  \cite{park2021bridge} & 37.2  & \underline{36.9} \cr
    \midrule
    \multirow{2}{*}{Hierarchy} 
   & HCRN \cite{le2021hierarchical}  & 36.1  & 35.6 \cr
  &  HOSTR \cite{dang2021hierarchical} & \underline{39.4}  & 35.9 \cr
    \midrule
    \multirow{2}{*}{Causal view} 
    & IGV (Ours)  & \textbf{40.8} & \textbf{38.3} \cr
    & Abs. Improve & +1.4 & +1.4 \cr
    \bottomrule
    \end{tabular}}
  \label{w2-tab:sota-ms}%
\end{table}%

%% file: work2/tab/model-agnostic.tex
\setlength{\tabcolsep}{7pt}
\begin{table}[t]
\small
  \centering
  \caption{IGV strategy is applied to different SoTAs methods. "+IGV" denoted our strategy is incorporated.}
    \begin{tabular}{l|cc|cc}
    \toprule
    \multirow{2}[2]{*}{Models} & \multicolumn{2}{c}{MSVD-QA} & \multicolumn{2}{c}{MSRVTT-QA} \\
          & Baseline & +IGV & Baseline & +IGV \\
    \midrule
    \midrule
    Co-Mem \cite{gao2018motionappearance} & 34.6  & \textbf{37.7} & 35.3  & \textbf{37.3} \\
    HGA \cite{jiang2020reasoning}  & 35.4  & \textbf{38.8} & 36.1  & \textbf{37.5} \\
    Our Backbone  & 36.1  & \textbf{40.8} & 36.3  & \textbf{38.3} \\
    \bottomrule
    \end{tabular}
  \label{w2-tab:model-agnostic}%
\end{table}%

%% file: work2/tab/ablation-loss.tex
\setlength{\tabcolsep}{6.5pt}
\begin{table}[tbp]
\small
  \centering
  \caption{Study of IGV loss components}
   \vspace{-5pt}
    \resizebox{0.85\linewidth}{!}{
    \begin{tabular}{l|cc|cc}
    \toprule
    \multirow{2}[1]{*}{Variants} & \multicolumn{2}{c}{MSVD-QA} & \multicolumn{2}{c}{MSRVTT-QA} \\
          & Our Backbone   & Co-Mem \cite{gao2018motionappearance} & Our Backbone   & Co-Mem \cite{gao2018motionappearance} \\
    \midrule
    \midrule
    Baseline & 36.1  & 34.6  & 36.3  & 35.3 \\
    $\Lapl_{\hat{c}}$     & 36.0    & 33.3  & 36.7  & 36.0 \\
    $\Lapl_{\hat{c}}+\Lapl_{\hat{e}}$   & 37.4  & 36.1  & 37.8  & 36.8 \\
    $\Lapl_{\hat{c}}+\Lapl_{v^{*}}$   & 38.2  & 36.3  & 37.4  & 36.2 \\
    $\Lapl_{\hat{c}}+\Lapl_{\hat{e}}+\Lapl_{v^{*}}$ & \textbf{40.8} & \textbf{37.7} & \textbf{38.3} & \textbf{37.3} \\
    \bottomrule
    \end{tabular}
    }
  \label{w2-tab:ablation-loss}%
\end{table}%

%% file: work2/sec/5_conclusions.tex
\section{Conclusions}
\vspace{-5pt}
In this chapter, we pinpoint that the spurious visual-linguistic correlations in VideoQA are triggered by question-irrelevant scenes.
We propose a novel invariant grounding framework, IGV, to distinguish the causal scene and emphasize its causal effect on the answer.
With the grounding indicator and scene intervener, IGV captures the causal patterns that remain stable across different environments.
Extensive experiments verify the effectiveness of IGV on different backbone VideoQA models.



%% file: work3/main.tex
\chapter{Equivariant and Invariant Grounding for VideoQA}
\label{cha:eigv}

\input{ work3/sec/0_abstract}
\input{ work3/sec/1_intro_cam_ready}

\input{ work3/sec/6_related}
\input{ work3/sec/2_preliminaries}

\input{ work3/sec/3_reformulation_new}

\input{ work3/sec/4_method}
\input{ work3/sec/5_experiment}

\input{ work3/sec/7_conclusion}
s

%% file: work3/sec/0_abstract.tex
In addition to robustness, causal modeling also brings about visual explainability as an inherent benefit. In this chapter, we explore another causal VideoQA model to strengthen the explainability and reasoning robustness by answering the question ``What part of the video should the model look at to answer the question?'', which exhibits the visual-linguistic alignment during answering.

Instead of unreliable post-hoc explainability, this work focus on intrinsic interpretability to make the answering process transparent.
At its core is the grounding of the question-critical cues as the causal scene to yield the answers, while rolling out the question-irrelevant information as the environment scene.
Taking a causal look at VideoQA, 
we devise a self-interpretable framework, \underline{E}quivariant and \underline{I}nvariant \underline{G}rounding for Interpretable \underline{V}ideoQA (EIGV).
Specifically, the equivariant grounding encourages the answering to be sensitive to the semantic changes in the causal scene and question; in contrast, the invariant grounding enforces the answering to be insensitive to the changes in the environment scene.
By imposing them on the answering process, EIGV is able to distinguish the causal scene from the environment information, and explicitly present the visual-linguistic alignment.
Extensive experiments on three benchmark datasets justify the superiority of EIGV in terms of accuracy and visual interpretability over the leading baselines.

%% file: work3/sec/1_intro_cam_ready.tex
\section{Introduction}
\label{w3-sec:introduction}

\begin{figure}[t]
\centering
\includegraphics[scale=0.47]{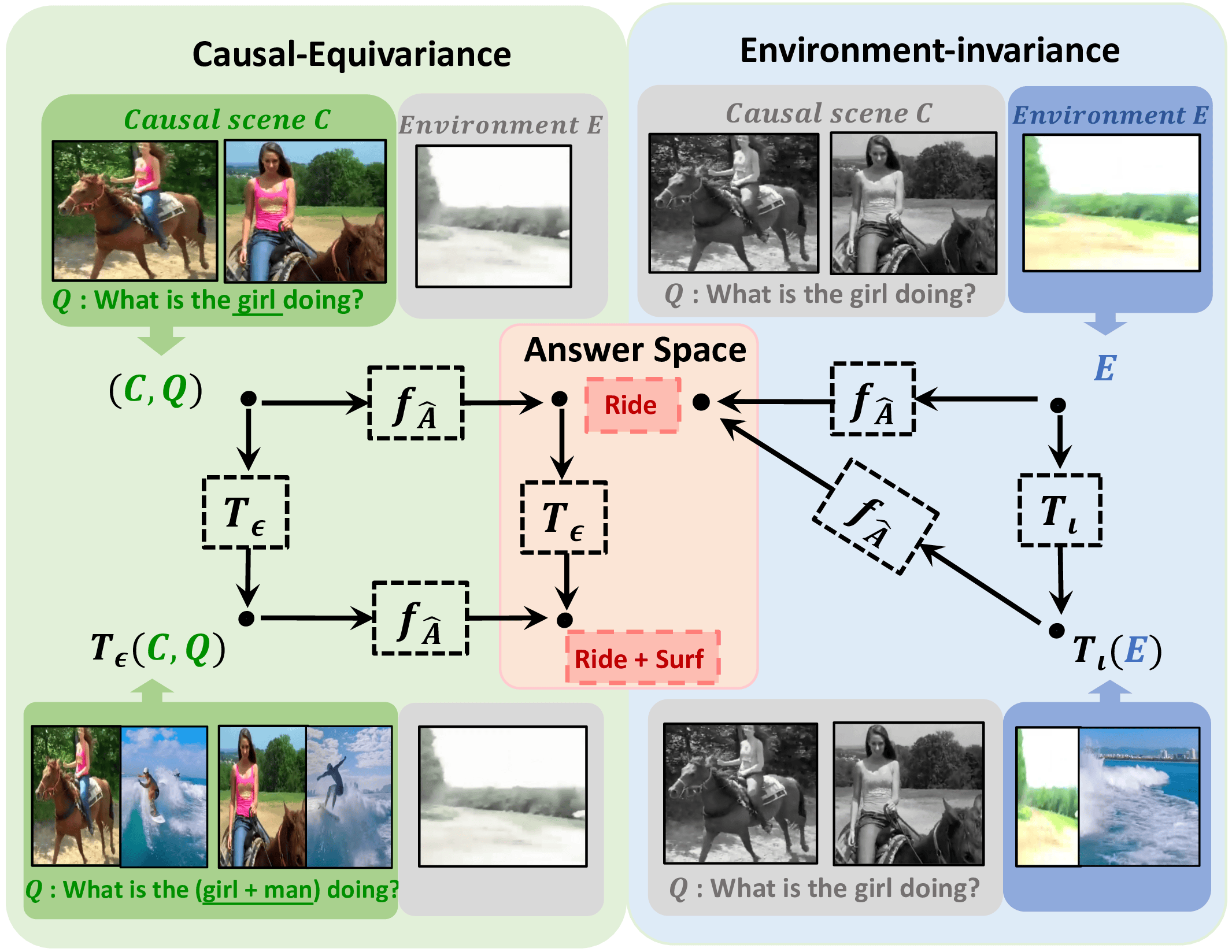}
\caption{
Illustration of equivariant and invariant grounding.
The causal-equivariant principle (left) asks that the semantic change $T_{\epsilon}$ applied to the causal scene $C$ and question $Q$ should be faithfully reflected in the answer change.
In contrast, the environment-invariant principle (right) outputs the same answer, regardless of changes $T_{\iota}$ on the environment scene $E$.
Here, $f_{\hat{A}}$ maps input to answer space.
}
\label{w3-fig:overview}
\end{figure}

Video Question Answering (VideoQA) \cite{zhong2022video} is a keystone in interactive AI, such as vision-language navigation and communication systems.
It aims to answer the natural language question based on the video content.
Striving for the architecture novelty, many studies have been conducted on modeling VideoQA's multi-modal nature, such as fostering the vision-language alignment \cite{jiang2020reasoning,park2021bridge} and revisiting the visual input structure \cite{le2021hierarchical,dang2021hierarchical}.
However, existing VideoQA models usually operate as black boxes, which fail to exhibit the working mechanism behind the predictions and hardly exhibit ``What knowledge should the model use to answer the question about the video?''.
As a result, the black-box nature causes concern for the model's reliability, especially in applications to safety and security.

The concern on the black-box nature calls for  better transparency of VideoQA models.
Here we focus on visual-explainability \cite{CSS,DBLP:conf/ijcai/RossHD17}, aiming to reveal ``Which part of the video should the model look at to answer the question?''.
It requires us to find a subset of visual scenes --- rationale --- that support the answering as evidence in way of human interpretation \cite{DBLP:conf/ijcai/RossHD17}.
Taking \cref{w3-fig:overview} as an example, when answering the question ``What is the girl doing?'', the rationale should focus on the ``girl-riding on-horse'' scene in the first two clips.
Towards this end, existing studies \cite{gao2018motionappearance,liu2021hair,park2021bridge} dwell mainly on the paradigm of \textbf{post-hoc explainability} \cite{DBLP:conf/iccv/SelvarajuCDVPB17}, which distributes the predictive answer of the target model to the input visual features via an additional explainer method.
They visualize the attention weights or gradient-like signals toward the visual features, and then identify a salient pattern as the rationale.
However, post-hoc explainability has several major limitations:
(1) It fails to make the target model intrinsically interpretable \cite{DBLP:conf/cvpr/YangZQ021,wang2021causal,DBLP:journals/natmi/Rudin19},  only approximating the decision-making process of the model.
As a result, the identified rationale cannot faithfully reveal how the model leverages the multi-modal information.
(2) Such visual inspections are fragile against input perturbations, since some artifacts can be easily captured as explanations instead of genuine knowledge from the data \cite{DBLP:conf/ijcai/LaugelLMRD19,slack2020fooling,heo2019fooling,ghorbani2019interpretation}.

The limitations of post-hoc explainability inspire us to explore the paradigm of \textbf{intrinsic interpretability} \cite{ghorbani2019interpretation,DBLP:journals/natmi/Rudin19}, which embeds a rationalization module into the model to make the decision-making process transparent.
Surprisingly, the intrinsic interpretability of VideoQA models is until-now lacking.
To fill the void, we draw on \textbf{causal theory} \cite{pearl2016causal,pearl2009causal} to formulate the interpretability task as disclosing ``Which part of the video is critical/causal to answering the question?''.
Concretely, we aim to identify the causal component of input video on-the-fly, which holds the question-response information and filters out the question-irrelevant cues.
Following this essence, one straightforward realization is to ground the input video into two segments:
(1) \textbf{causal scene}, which retains the question-critical visual content and sufficiently approaches the answer, thus naturally serving as the rationale;
and (2) \textbf{environment scene}, which holds the question-irrelevant visual content and can be seen as the rationale's complement.

%
However, discovering causal scene without the supervision of ground-truth rationale is challenging.
With a causal look at the reasoning process (\cf Section  \cref{w3-sec:causal-view}), we argue that the crux of intrinsic interpretability is to amplify the connection between the causal scene and the answer, while blocking the non-causal effect of the environment scene.
Following this line, we propose two principles to guide the grounding of the rationale:
\begin{itemize}[leftmargin=*]
    \item \textbf{Causal-Equivariance.}
    By ``equivariance'', we mean that answering should be sensitive to the semantic changes on the causal scene and question (termed E-intervention), \eg any change on the causal scene and question should be faithfully reflected on the predicted answer. For example, in \cref{w3-fig:overview}, the ``girl-riding on-horse'' and ``man-surfing in-ocean'' scenes are the oracle rationales of ``What is the girl doing?'' and ``What is the man doing?'', respectively. The intervention \cite{ivrd} applied on the input (\ie mixing the ``girl-riding on-horse'' and ``man-surfing in-ocean'' scene, and combining two questions as ``What is the girl doing? What is the man doing?'') should set off an equivariant change in the answer (\ie changing from ``Ride'' to ``Ride+Surf'').

    \item \textbf{Environment-Invariance.}
    By ``invariance'', we mean that answering should be insensitive to the changes in the environment scene (termed I-intervention), conditioning on the causal scene and question.
    Considering \cref{w3-fig:overview} again, the intervention applied to the environment (\ie mixing the ``meadow'' and ``ocean'' scenes) implies no impact towards answering ``What is the girl doing?'', reflecting a homogeneity in the answer space.
\end{itemize}

To impose these two principles for intrinsic interpretability, we propose a new framework, \underline{E}quivariant and \underline{I}nvariant \underline{G}rounding for Interpretable \underline{V}ideoQA (\textbf{EIGV}).
EIGV equips the VideoQA backbone model with three additional modules:
a grounding indicator, an intervener, and a disruptor.
First, the grounding indicator learns to attend the causal scene based on the input question, while leaving the rest as the environment.
Then, the intervener parameterizes the proposed principles to guide the grounding.
Specifically, towards the causal-equivariance principle, it conducts the E-intervention on the causal scene and question --- that is, mix them with the counterparts from another video-question pair --- and encourages the predictive answer to be anticipated accordingly.
Towards the environment-invariance principle, when leaving the causal scene and question untouched, it applies the intervention on the environment --- that is, mix it with the environmental stratification of a memory bank --- and enforces the predictive answer to be invariant.
Moreover, we build an unified sight of two principles via the lens of contrastive learning.
Concretely, on top of each intervened video-question pair, the disruptor constructs the positive views by disrupting the environment scene randomly, while creating the negative views by substituting the causal scene with random scenes.
Training with these two principles allows the backbone model to distinguish the causal scene from the environmental cues, and hinge on the critical visual-linguistic alignment.

Briefly put, our contributions are: 
\begin{itemize}[leftmargin=*]
\item \textbf{Enhancing Interpretability and Accuracy in VideoQA}: Our proposed EIGV framework is a model-agnostic VideoQA solution that stands out for its ability to distill causal visual-linguistic alignment, thus enabling answer generation in a self-interpretable manner. The key benefit here is the significant increase in the interpretability and accuracy of VideoQA models. EIGV provides a new way to approach answer generation, allowing for more transparent and trustworthy decision-making processes in AI systems.

\item \textbf{Innovative Approach to Visual Grounding}: By investigating the grounding rationale through the equivariant-invariant principle, our research introduces a novel perspective to visual grounding in VideoQA. This approach benefits the field by reducing the assumptions traditionally made about visual data processing. It offers a more nuanced understanding of how visual information is anchored and processed, leading to more precise and reliable VideoQA models.

\item \textbf{Empirical Superiority and Versatility} The validation of EIGV's effectiveness on three benchmark datasets (MSVD-QA, MSRVTT-QA, NExT-QA) highlights its benefits in terms of performance, where it consistently outperforms state-of-the-art models. This superiority demonstrates not just an increase in accuracy and efficiency but also showcases EIGV's versatility. Being model-agnostic, it can be applied across different VideoQA models, offering a widely adaptable solution that enhances the overall performance and applicability of VideoQA systems.
\end{itemize}

%% file: work3/sec/6_related.tex
\section{Related Work}
\label{w3-sec:related}

\smallskip
\noindent\textbf{Video Question Answering.}
Established to answer the question in dynamic visual content, VideoQA is bred through the task of ImageQA but has broadened its definition by assembling a temporal dimension. 
To make the task intriguing, the VideoQA benchmark has gone beyond the problem of description \cite{DBLP:conf/mm/XuZX0Z0Z17} and built several datasets to challenge temporal reasoning and even causal reflection \cite{next-qa}. 
As a result, VideoQA has experienced an aggressive expansion in the architecture design. 
Chronologically, early efforts tend to enact alignment through cross-modal attention \cite{zeng2016leveraging,li2019beyond} or enhanced memory design \cite{gao2018motionappearance, DBLP:conf/mm/XuZX0Z0Z17, fan2019heterogeneous}, while more recent works leverage the expressiveness of the graph neural network and perform the relation reasoning as node propagation \cite{jiang2020reasoning, mspan} or graph alignment \cite{park2021bridge}. 
In addition, current designs modify the representation of video and manipulate the temporal sequence from a hierarchical angle. Following this line, HCRN \cite{le2021hierarchical} first came out with the conditional relation module as building blocks that operate through different video intervals, whereas  HOSTR and PGAT made their advancement by incorporating visual content from different granularity. MSPAN, however, established cross-scale feature interaction on top of the hierarchy.
Despite effective, their intrinsic rationale has long been overlooked. To the best of our knowledge, EIGV is the first work that probes intrinsic interpretation. 

\smallskip
\noindent\textbf{Invariant Learning.}
Given a encoder $f(\cdot)$ and input $x$, a representation $f(x)$ is invariant to operation $T$, if $\forall x : f(G(x)) = f(x)$. 
In practice, this invariant property has a long history in presenting visual content (\eg HOG \cite{DBLP:conf/mmm/HuangTHTJ11}), which has recently been renovated by deep learning in form of risk minimization. As its most prevailing form, IRM \cite{arjovsky2020invariant} fosters this philosophy by posing an environment invariant prior and discovering the underlying causal pattern by reducing cross-environment variance.   
Different from previous studies that create environment via inductive re-grouping \cite{DBLP:conf/cvpr/AndersonWTB0S0G18} or adversarial inference \cite{DBLP:conf/icml/CreagerJZ21,wang2021causal,wang2021clicks}, our method conducts causal intervention that perturbs the original sample distribution to form a new one.
The most relevant work is \cite{IGV}, where an invariant framework is introduced as a model-agnostic framework. Similarly, EIGV also adopts this environment-invariant framework. However, EIGV gains better generalization ability by marrying causal-equivariance as a complementary learning principle, that helps to identify the causal scene of interest.

\smallskip
\noindent\textbf{Visual Interpretability.}
Machine interpretability can be achieved in various methods, such as clustering \cite{monnier2020dticlustering} or disentanglement \cite{shen2020interfacegan}. Our design can be vested in the category of attribution discovery, which investigates the contribution of different input elements toward the prediction. 
Based on whether the prediction and interpretation are yielded simultaneously, two categories are further defined: 1) post-hoc methods that generated the interpretation after prediction, such as backpropagation methods (\eg grad-CAM \cite{DBLP:conf/iccv/SelvarajuCDVPB17}). 2) self-interpretable method that cast prediction and interpretation at the same stage.
Unlike the post-hoc method that traces the interpretative clue from the output of the black-box, the self-interpretable model builds a transparency model via methods such as prototype generation \cite{DBLP:journals/corr/abs-1806-10574} or structural delineation \cite{DIR}. 
In fact, previous works tend to focus on static image. EIGV, however, approaches the video interpretation in a multi-modal situation. 

\vspace{-10pt}

\begin{table}[h]
  \centering
  \caption{Key Related Papers}
    \begin{tabular}{lp{10cm}} 
    \toprule
    Paper & Remark \\
    \midrule
    \cite{IGV} & This paper is the foundation of this work, which introduce the invariant learning to this VideoQA. On top of this, this work further develops equivariant grounding to enhance the grounding result.
    \\
    \midrule
    \cite{mixup} & Mix-up is a well-established data augmentation method. Instead of using mix-up as augmentation, we form mixup as a learning regularization to support the equivariance in this work.\\
    \bottomrule
    \end{tabular}%
\end{table}%

%% file: work3/sec/2_preliminaries.tex
\section{Preliminaries}
\label{w3-sec:preliminaries}
Since the task formulation and causal graph for VideoQA have been introduced in \cref{w2-sec:preliminary} and \cref{w2-sec:causal-view}, we briefly recap preliminary in this section.   

\vspace{5pt}
\noindent\textbf{Modeling}.
Given the video $V$, the VideoQA model $f_{\hat{A}}{(\cdot)}$ answers the question $Q$ by formulating the visual-linguistic alignment: 
\begin{gather}\label{w3-eq:conventional-modeling}\
    \hat{A} = f_{\hat{A}}(V,Q),
\end{gather}
where $\hat{A}$ is the predictive answer. Typically, $f_{\hat{A}}{(\cdot)}$ is a combination of a video-question encoder and a answer decoder. 
%
%
To optimize the video-question encoder and answer decoder, current VideoQA models usually adopt the scheme of empirical risk minimization (ERM) \cite{gao2018motionappearance,le2021hierarchical,jiang2020reasoning, pgat}, which measures and minimizes the risk between the ground-truth answer $A$ and predictive answer $\hat{A}$:
\begin{gather}\label{w3-equ:erm-loss}
    \min\mathcal{L}_{\text{ERM}}(\hat{A}, A).
\end{gather}

%% file: work3/sec/3_reformulation_new.tex


\begin{figure}[t]
\centering
\includegraphics[scale=0.5]{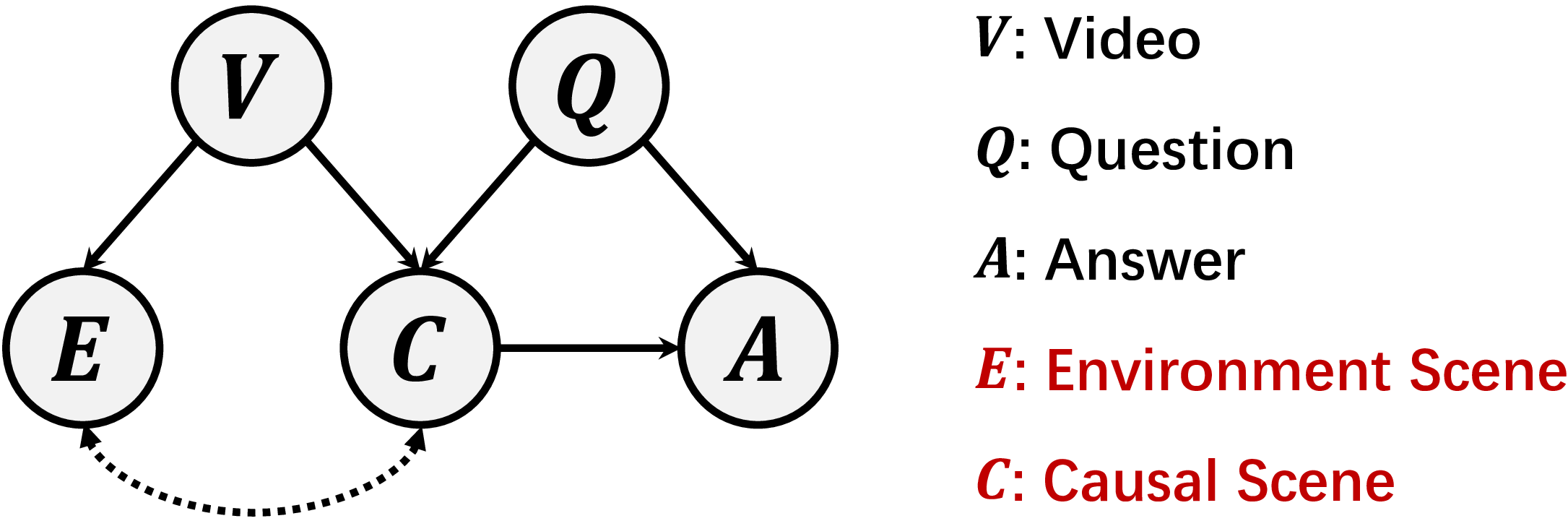}
\caption{Causal Graph of VideoQA}
\label{w3-fig:scm}
\end{figure}

\noindent\textbf{Causal Graph for VideoQ.}\label{w3-sec:causal-view}
\cref{w3-fig:scm} illustrates the causal graph, where each link depicts the cause-effect relationship between two variables:
\begin{itemize}[leftmargin=*]
    \item $Q\to C, E \gets V$. Given the question of interest $Q$, the video $V$ can be partitioned into two parts: the causal scene $C$ and the environment scene $E$. For example, to answering ``What is the girl doing?'' in \cref{w3-fig:overview}, $C$ should be the first two clips describing the ``girl-riding on-horse'' scene, while $E$ should be the last clip about the ``meadow'' scene. Moreover, the varying semantics of different questions will emphasize different $C$.
    \item $Q\rightarrow A \leftarrow C$. The visual knowledge in the causal scene $C$ and the linguistic semantics in the question $Q$ collaborate together to determine the answer $A$. Furthermore, this path, which presents the visual-linguistic alignment, internally interprets the reasoning.
    \item $E\dashleftarrow\dashrightarrow C$. The dashed arrow sketches additional probabilistic dependencies \cite{pearl2000causality} between $C$ and $E$, which typically arise from selection bias \cite{DBLP:conf/cvpr/TorralbaE11}. For example, the ``meadow'' scene is frequently collected as a common environment for the ``horse-riding'' scene. 
\end{itemize}

\noindent \textbf{Beyond ERM.}
With inspections on prior VideoQA studies, we investigate their inability to distinguish the causal and non-causal effects of scenes.
Specifically, in conventional VideoQA models, video and question are directly paired together to model their interaction and approach the golden answer, consequently.
Inevitably, taking video as a whole leaves the contributions of scenes untouched, thus failing to differentiate $C$ from $E$ and forgoing their function divergence towards the answer.
Worse still, ERM enforces these models to blindly capture the statistical correlations between the video-question pairs and answers.
As such, the visual-linguistic alignment hinges easily on the spurious correlations between $E$ and $A$, owing to the backdoor paths \cite{pearl2016causal}, which hinders the generalization of models \cite{niu2021counterfactual,DIR}.
Therefore, identifying the causal scene $C$ is critical to addressing these limitations.

%% file: work3/sec/4_method.tex
\section{Methodology}
\label{w3-sec:method}

\begin{figure*}[t]
\centering
\includegraphics[width=0.97\textwidth]{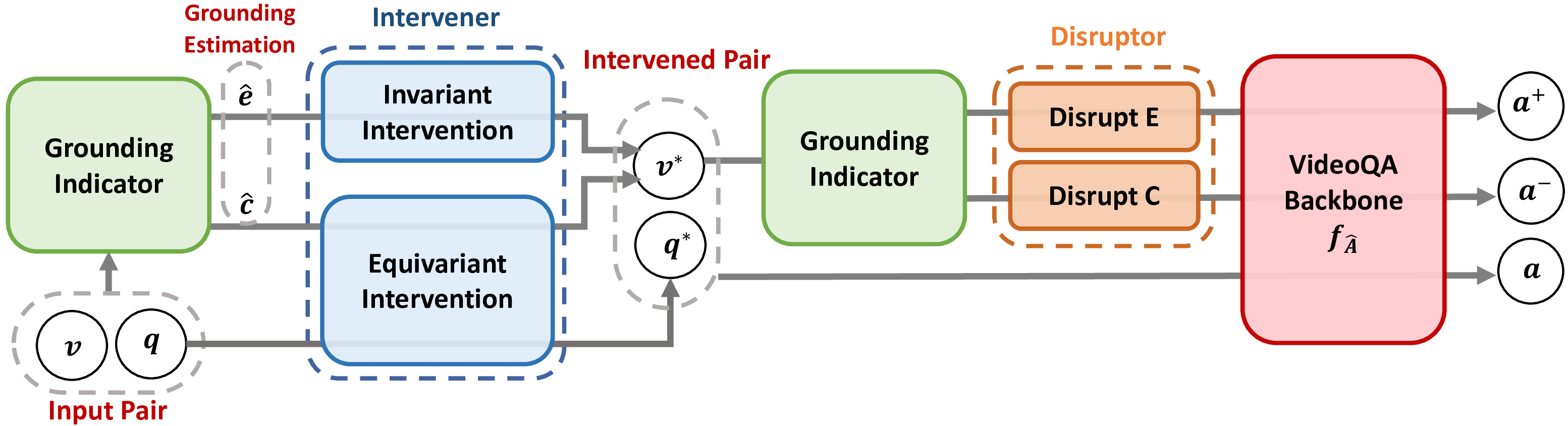}
\caption{Overview of EIGV. It comprises three additional modules on top of the conventional VideoQA backbone: 1) Grounding indicator, 2) Intervener, and 3) Disrupter. First, the grounding indicator learns the estimation of causal scene $\hat{c}$ and environment $\hat{e}$. Next, two interventions are imposed on the causal and non-causal factors to compose the intervened pair $(v^*,q^*)$. Finally, based on the re-grounded result, the disruptor creates contrastive samples, which are further feed into the VideoQA backbone.}
\label{w3-fig:main}
\end{figure*}

To ground the causal scene $C$ in the video $V$, we take a closer look at the VideoQA SCM (\ie \cref{w3-fig:scm}a), and emphasize the essential differences between $C$ and $E$.
Specifically, given the causal scene $C$ and question $Q$, the answer $A$ is determined, regardless of the variations in the environment scene $E$:
\begin{gather}\label{w3-equ:eq0}
      A\bot E \mid C,Q,
\end{gather}
where $\bot$ denotes the probabilistic independence. 

\vspace{5pt}
\noindent\textbf{Rationalization.} 
During training, the oracle grounding rationale $C$ is out of reach, while only the input $(V, Q)$ pair and training target $A$ are observed. 
Such an absence motivates VideoQA to embrace video grounding in its modeling. 
Specifically, in light of question $Q$, the estimated causal scene $\hat{C}$ is grounded from the massive $V$ to approach the oracle $C$ and then generate prediction $\hat{A}$ via $Q \rightarrow A \leftarrow C$. 
To systematize this relation, the causal-equivariance principle introduces an equivariant transformation $T_\epsilon$ to each of the parent variables (\ie $C$ and $Q$), and expects a proportionate change in the response variable (\ie $A$). On top of SCM, we formally present such notions as:
\begin{gather}
    T_\epsilon(\hat{A})=f_{\hat{A}}(T_\epsilon(\hat{C}),T_\epsilon(Q)). \label{w3-equ:equivariance}
\end{gather}
Meanwhile, environment-invariant principle formulated Equation \cref{w3-equ:eq0} in the sense that imposing an invariant-transformation $T_\iota$ on the estimated environment $\hat{E}$ should not trigger variation of answer $A$: 
\begin{gather}
      \hat{A}=f_{\hat{A}}(T_\iota(\hat{E}),Q)),\label{w3-equ:invariance}
\end{gather} 
To this end, we parameterize our learning framework, EIGV, as a combination of equivariant and invariant principles, which comprises three additional modules on top of the ERM-guided backbone: grounding indicator, intervener, and disruptor. In a nutshell, we display our EIGV framework in \cref{w3-fig:main}.

\vspace{5pt}
\noindent \textbf{Data representation.}
Following previous efforts \cite{jiang2020reasoning, mspan}, we encode video instance $v$ as a sequence of $K$ fixed visual clips, while question instance $q$ is encoded into a similar form with a fixed length of language tokens $L$.
Then, visual and linguistic features are applied with a linear layer and an LSTM \cite{10.1162/neco.1997.9.8.1735}, respectively, to align their dimension. As a result, we acquire the output of linear layer $ \Mat{v} \in \mathbb{R}^{k \times d}$ as the final video representation and the last hidden state of LSTM $\Mat{q} \in \mathbb{R}^{d}$ as the holistic question representation.


\subsection{Grounding Indicator}
Scene partition is fundamental to the rationale discovery, whose core is to estimate the value of $C$ and $E$ via a hard split on video $V$. Given an input sample $(v,q)$, the grounding indicator aims to access the causal scene and environment scene via their estimated value $\hat{c}$ and $\hat{e}$ according to question $Q$. 
Concretely, we first construct two cross-modal attention modules to indicate the probability of each visual clip of being causal scene ($\Mat{p}_{\hat{c}} \in\mathbb{R}^{K}$) 
and environment scene ($\Mat{p}_{\hat{e}}\in\mathbb{R}^{K}$):
\begin{align}
    \Mat{p}_{\hat{c}} &= \text{Softmax}(\text{FC}_{1}(\Mat{v})\cdot\text{FC}_{2}(\Mat{q})^\intercal),\\
    \Mat{p}_{\hat{t}} &= \text{Softmax}(\text{FC}_{3}(\Mat{v})\cdot\text{FC}_{4}(\Mat{q})^\intercal),
\end{align}
where $\text{FC}_{1},\text{FC}_{2},\text{FC}_{3},\text{FC}_{4}$ are fully connected layers that align cross-modal representations.
However, gathering messages via a soft mask still makes the visual information on different clips overlap. 
As discussed in Section 3.2, guided by ERM, the conventional attention mechanism is unable to block the influence of $\hat{e}$, thus undermining the veracity of $\hat{c}$.
%
As a correction, the grounding indicator makes a discrete selection over the clip-wise attention result to generate a disjoint group of the causal scene. We leverage Gumbel-Softmax \cite{DBLP:conf/iclr/JangGP17} to manage a differentiable selection on attentive probabilities and compute the indicator vector $\Mat{I}\in\mathbb{R}^{K\times 2}$ on the two attention scores  over each clip (\ie $\Mat{p}_{\hat{c,i}}$, $\Mat{p}_{\hat{e,i}}$, $i \in K$).  Formally, $\Mat{I}$ is derived as:
\begin{gather}
    \Mat{I} = \text{Gumbel-Softmax}([\Mat{p}_{\hat{c}};\Mat{p}_{\hat{e}}]), 
\end{gather}
where $[;]$ denote concatenation. The first and second column of $\Mat{I}$ (\ie $I_{0}$ and $I_{1}$) index the attribution of $\hat{c}$ and $\hat{e}$ over k clips, respectively. 
To this end, we estimate $\hat{c}$ and $\hat{e}$ as follows:
\begin{gather}
    \hat{c} = I_{0}\cdot v ,\quad \hat{e} = I_{1}\cdot v, \quad \st v=\hat{c}+\hat{e} .
\end{gather}


\subsection{Intervener}
\begin{figure}[t]
\centering
\includegraphics[scale=0.6]{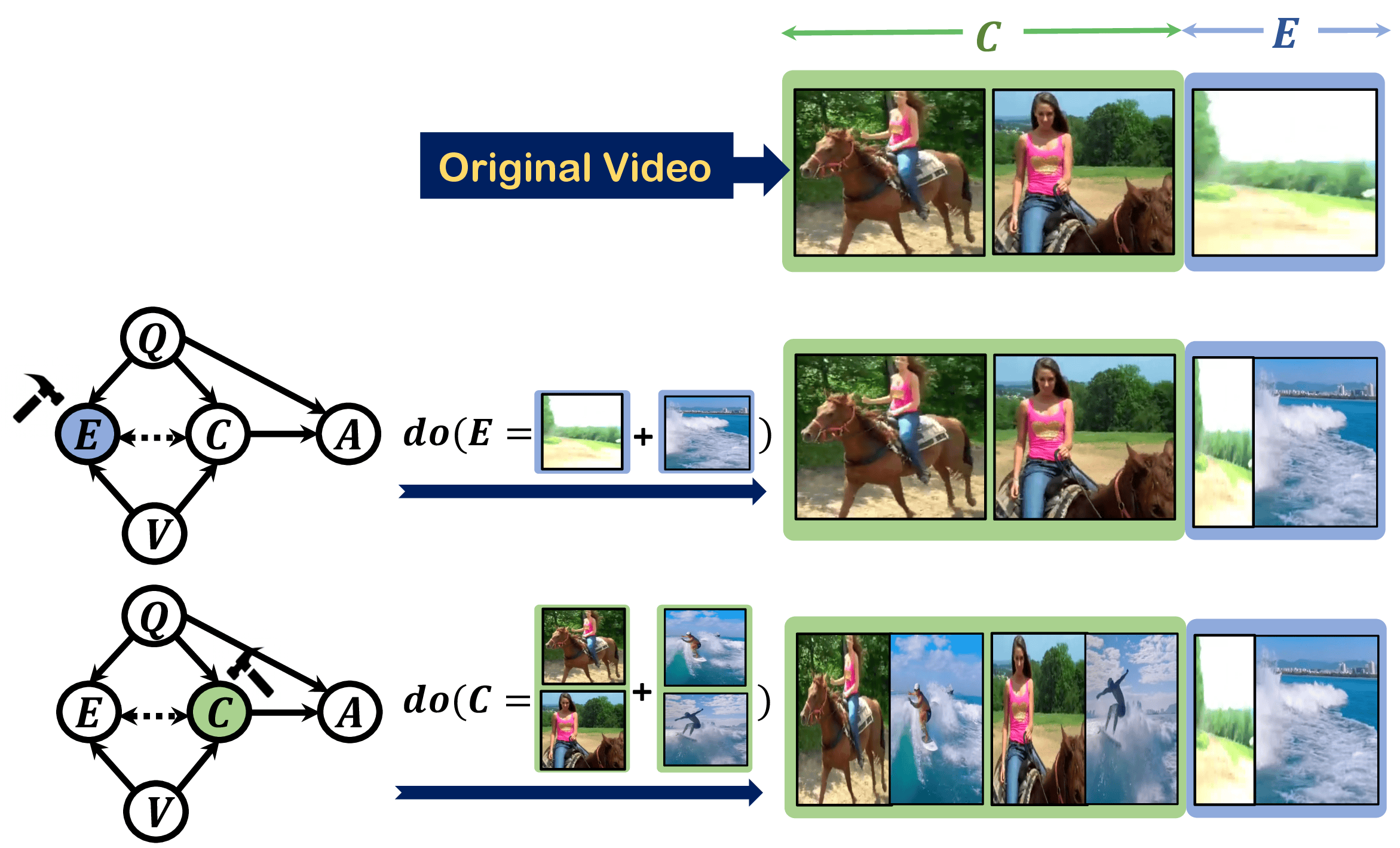}
\caption{We illustrate the invariant and equivariant interventions in the second and third rows, respectively. The effects on $Q$ and $A$ are omitted for illustration purposes.}
\vspace{-5pt}
\label{w3-fig:intervene}
\end{figure}

In absence of clip-level supervision, learning grounding indicators requires dedicated exploitation of the equivariance-invariance principle.
On this demand, we propose the intervener, which prompts the estimated rationale to the oracle by intervening $\hat{c}$ and $\hat{e}$. 
 \cref{w3-fig:intervene} describes the functionality of $do(\cdot)$ --- the intervention operator that successively manipulated SCM over $E$ and $C$. Concretely, two intervention operations are configured to realize the equivariant and invariant transformation defined in \cref{w3-equ:equivariance} and \cref{w3-equ:invariance}. 
%

To fulfill the causal-equivariant principle, we design the E-intervention on the causal scene $\hat{c}$, which applies a linear interpolation between two data points on their causal factors --- $C$, $Q$ and $A$.  
By casting the same mixing ratio $\lambda_0\sim\text{Beta}(\alpha,\alpha)$ on all causal factors, the equivariant intervener learns to capture 
the causal connection of $C, Q \to A$.
In particular, we attain the intervened causal scene $c^*\in \mathbb{R}^{K\times d}$, question $q^*\in \mathbb{R}^{d}$ and answer $a^*\in \mathbb{R}$ as follow:
\begin{gather}
    c^*=\lambda_0 \cdot \hat{c}+(1-\lambda_0) \cdot \hat{c}',\\
    q^*=\lambda_0 \cdot q+(1-\lambda_0) \cdot q',\\
    a^*=\lambda_0 \cdot a+(1-\lambda_0) \cdot a',
\end{gather}
where $\hat{c}'$, $q'$ and $a'$ are causal factors from a second sample.

To achieve the environment-invariant principle, we devise the I-intervention that adopts a similar mixing strategy to the environment scene $\hat{e}$. 
Notably, by drawing the mixing ratios $\lambda_1$ from a second distribution that is distinct from the equivariant one (\ie $\lambda_1\sim\text{U}(0,1)$), the invariant intervention learns to rule out the influence of environment scene on the answer, which essentially refines the ERM-guided scheme at our will. 
Formally, we arrive the intervened environment scene $e^*$ by:
\begin{gather}
    e^*=\lambda_1 \cdot \hat{e}+(1-\lambda_1) \cdot \hat{e}',
\end{gather}
where $\hat{e}'$ is the estimated environment scene of a second sample.

In practice, the equvariant and invariant intervention operations are performed in parallel on different parts of $v$, and the intervened video $v^* \in \mathbb{R}^{K\times d}$ is composed of $do(C=c^*)$ and $do(E=e^*)$:
\begin{gather}
    v^*=c^*+e^*.
\end{gather}


%
\subsection{Disruptor}

To fully exploit the privilege of the proposed principles, we employ contrastive learning as an auxiliary objective to establish a good representation that maintains the desired properties of $\hat{c}$ and $\hat{e}$. 
Specifically, we first compose a memory bank $\pi$ as a collection of visual clips from other training videos. Then, we apply the grounding indicator a second time on top of the intervened variables, where the re-grounded causal and environment scene are manipulated to set up the contrastive twins as follows:
\begin{itemize}[leftmargin=*]
\item In light of the environment-invariance principle, positive video is developed in the sense that changing the environment scene will not provoke disagreement in answer semantics.  Thus, the disruptor synthesizes a positive video $v^+$ by disrupting the $v^*$ on its environment part --- that is, replacing the environment scene with a random stratification sampled from the memory bank.\footnote{Note that the environment substitutes will not involve the question-relevant scenes, to avoid creating additional paths from $E$ to $A$.}  
\item Built upon the causal-equivariance principle, the negative counterpart $v^-$ is created by a similar disruption but on the causal scene of $v^*$, where substitution on the question-critical causal part should raise inconsistency in answer space.
%
Apart from the visual negatives, the disruptor also creates linguistic alternatives to enhance the distinctiveness of the vision-language alignment. Specifically, it disrupts the combination of the intervened input ($v^*, q^*$) and pairs the video with random sample question $q_r$ to create a second view of negative samples ($v^*, q_r$).
\end{itemize}
To this end, we attain the answer representation of anchor $\Mat{a}$ and its contrastive counterparts $\Mat{a}^+, \Mat{a}^-$ by feeding the paired positive and negative samples to backbone VideoQA model $f_{\hat{A}}$: 
\begin{gather}
    \Mat{a}=f_{\hat{A}}(v^*,q^*),\\
    \Mat{a}^+=f_{\hat{A}}(v^+,q^*),\\
    \Mat{a}^-=f_{\hat{A}}([(v^-,q^*);\,(v^*,q_r)]),
\end{gather}
where $[;]$ denotes concatenation.

Notably, EIGV is designed to be model-agnostic, which aims to promote any VideoQA backbone built on frame-level visual inputs.
%

\subsection{Optimization}
By far, we set up the intervened vision-language instance ($v^*,q^*$) for a pair of input ($v,q$), and further constitute its contrastive counterparts based on the estimated grounding result. 
To steer the learning process away from the conventional ERM pitfall, we establish two learning objectives on top of their output $a, a^+, a^-$:

\begin{itemize}[leftmargin=*]
    \vspace{5pt}
    \item \noindent \textbf{Contrastive loss.} To reflect the invariant property of the environment scene while maintaining the distinctiveness of the causal scene, we borrow the definition of InfoNCE \cite{van2018representation}, and construct the contrastive objective as follows:
    \begin{gather}
        \mathcal{L}_{CL}=-log(\frac{\text{exp}{(a^\top a^+)}}{\text{exp}{(a^\top a^+)}\,+\sum_{n}^{N}\text{exp}{(a^\top a_n^-)}}),
    \end{gather}
    where $N$ is the number of negative samples, $a_n^-$ denotes negative anwer generated by one of  negative samples.
    
    \vspace{5pt}
    \item \noindent \textbf{ERM loss.} 
    Estimating the rationale requires a robust causal connection from $V,Q$ to $A$. Thus, we imposed an entropy-based risk function $\text{XE}(\cdot)$ on $(v^*,q^*)$ to approach the intervened answer $a^*$:
    \begin{gather}
        \mathcal{L}_{ERM}=\text{XE}(f_{\hat{A}}(v^*,q^*), a^*),
    \end{gather}
\end{itemize}

As a result, the overall training objective of EIGV is the aggregation of the forgoing risks:
\begin{gather}
   \mathcal{L}_{\text{EIGV}} = \mathbb{E}_{(v,q,a)\in\mathcal{O}}\mathcal{L}_{ERM} + \beta\mathcal{L}_{CL},
\end{gather}
where $\mathcal{O}$ is the set of training instances $(v,q)$ alongside their ground-truth answer $a$; $\beta$ is the hyper-parameter that balances the strength of contrastive learning. The joint optimization disentangles the mischief of environment scene, thus fulfilling the desired interpretation by locating the causal pattern. During inference, EIGV generates the predication $\hat{a}$ without the intervener and disruptor involved, and gives the interpretation $\hat{c}$ as the partition result of the grounding indicator.




%% file: work3/sec/5_experiment.tex
\section{Experiment}
\label{w3-sec:experiment}

In this section, we show the experimental results to answer the following research questions.
\begin{itemize}[leftmargin=*]
\item \textbf{RQ1} How effective is EIGV in discovering the causal pattern and improving the model generalization across different settings?
\item \textbf{RQ2} How does the sub-module and feature setting contribute to the performance?
\item \textbf{RQ3} What pattern does EIGV capture in rationale discovery?
\end{itemize}

\subsection{Settings}

\vspace{5pt}
\noindent \textbf{Datasets.} We conduct experiments on three benchmark datasets that challenge the model's reasoning capacity from different aspects: 
\textbf{MSVD-QA}  \cite{DBLP:conf/mm/XuZX0Z0Z17} and \textbf{MSRVTT-QA} \cite{DBLP:conf/mm/XuZX0Z0Z17} mainly emphasize the recognition ability by asking the descriptive questions, where 50K and 243K question-answer pairs are automatically generated from the human-labeled video captions, respectively.
\textbf{NExT-QA} \cite{next-qa} pinpoints the causal and temporal relations among objects in the video. It contains 47.7K questions with answers in the form of multi-choice, which are manually annotated from 5.4K videos.

\vspace{5pt}
\noindent\textbf{Baseline.} We validate the effectiveness of EIGV across backbone VideoQA models of three kinds: 
1) \textbf{Memory-based} methods that foster a storage of input sequence via auxiliary memory design, such as AMU \cite{DBLP:conf/mm/XuZX0Z0Z17}, HME \cite{fan2019heterogeneous} and Co-Mem \cite{gao2018motionappearance}.
2) \textbf{Graph-based} methods that leverage the expressiveness of graph network to model the interaction between visual and language elements, which involves methods like L-GCN \cite{huang2020locationaware}, B2A  \cite{park2021bridge} and HGA \cite{jiang2020reasoning}. 
3) \textbf{Hierarchy-based} methods include HCRN \cite{le2021hierarchical}, PGAT \cite{pgat}, HOSTR \cite{dang2021hierarchical}, MSPAN \cite{mspan} and HQGA \cite{hqga}. In common, they exploit the multi-granularity nature of visual elements and realize the hierarchical reasoning via bottom-up architecture. 
In Specific, we test the generalization of EIGV by marrying our learning principles to three backbones of different categories: memory-based Co-Mem \cite{gao2018motionappearance}, graph-based HGA \cite{jiang2020reasoning} and hierarchy-based MSPAN \cite{mspan}. 

\vspace{5pt}
\noindent \textbf{Implementation Detail.} 
For input representation, we encode the video instance as a sequence of $K$=16 clips, where each clip is represented as a combination of appearance and motion features extracted from the pre-trained ResNet-152 and 3D ResNeXt-101. For the linguistic feature, we follow \cite{next-qa} and obtain the contextualized word representation using the fine-tuned BERT model. In the hyper-parameters setting, we set $d=512$ for cross-modal alignment, then train the model for 80 epochs with an initial learning rate of 5e-5.  During optimization, EIGV is trained with Adam optimizer and we decay the learning rate when validation stops improving for 5 epochs. The balance ratio $\beta$ is set to 0.75.

\subsection{Main Result (RQ1)}
\input{work3/tab/main}

\cref{w3-tab:main} shows the overall result of our method and the SoTAs on three benchmark datasets: MSVD-QA, MSRVTT-QA and NExT-QA. Our observations are summarized as follows:

\begin{itemize}[leftmargin=*]
\item Across all three benchmark datasets, the proposed EIGV outperforms SoTA by a distinct margin (+1.2\%$\sim$2.3\%). Such prevailing performance indicates the overall effectiveness of our design, which underpins the theoretical soundness of the equivariant and invariant principles. 

\item Narrowing the inspection to each of the three backbones, EIGV brings each backbone model a sharp gain across all benchmark datasets (+1.3\%$\sim$5.2\%), which evidences its model-agnostic property. 
Nevertheless, we notice that the improvements fluctuate across the backbones. As a comparison, on MSVD-QA and MSRVTT-QA benchmarks, EIGV acquires more favorable gains with backbone Co-Mem and HGA than it does with MSPAN. This is because the multi-granularity hierarchy empowers the MSPAN with robustness, especially to questions of the descriptive type. Therefore, it achieves stronger backbone performances on benchmarks that focus on the descriptive question (\ie MSVD-QA and MSRVTT-QA), which, in turn, account for the contribution of EIGV to some extent, thus makes improvement of MSPAN less remarkable.
In contrast, when it comes to the causal and temporal question (\ie NExT-QA) where the inherit advantage of MSPAN backbone vanishes,  EIGV shows equivalent improvements on all three backbones (+2.2\%$\sim$3.7\%). 
%

\item Comparing the average improvement across different benchmarks, we notice that EIGV achieves the best improvement on MSVD-QA (+2.3\%$\sim$5.2\%) while relatively moderate gains on MSRVTT-QA (+1.3\%$\sim$1.9\%) and NExT-QA (+2.2\%$\sim$3.7\%).
The reason for such discrepancy is that MSVD-QA is relatively small in size,
which constrains the reasoning ability of the backbone models by limiting their exposure to training instances.
As a comparison, MSVD-QA is five-time smaller than MSRVTT-QA in terms of QA pairs (43K vs 243K), and three-time smaller than NExT-QA in terms of video instances (1970 vs 5440).
However, such deficiency caters to the focal point of EIGV that develops better in a less generalized situation, thus leading to more preferable growth on MSVD-QA.
\end{itemize}

\input{work3/tab/ablation}

\subsection{In-Depth Study (RQ2)}

\begin{figure}[t]
\centering
\includegraphics[scale=0.55]{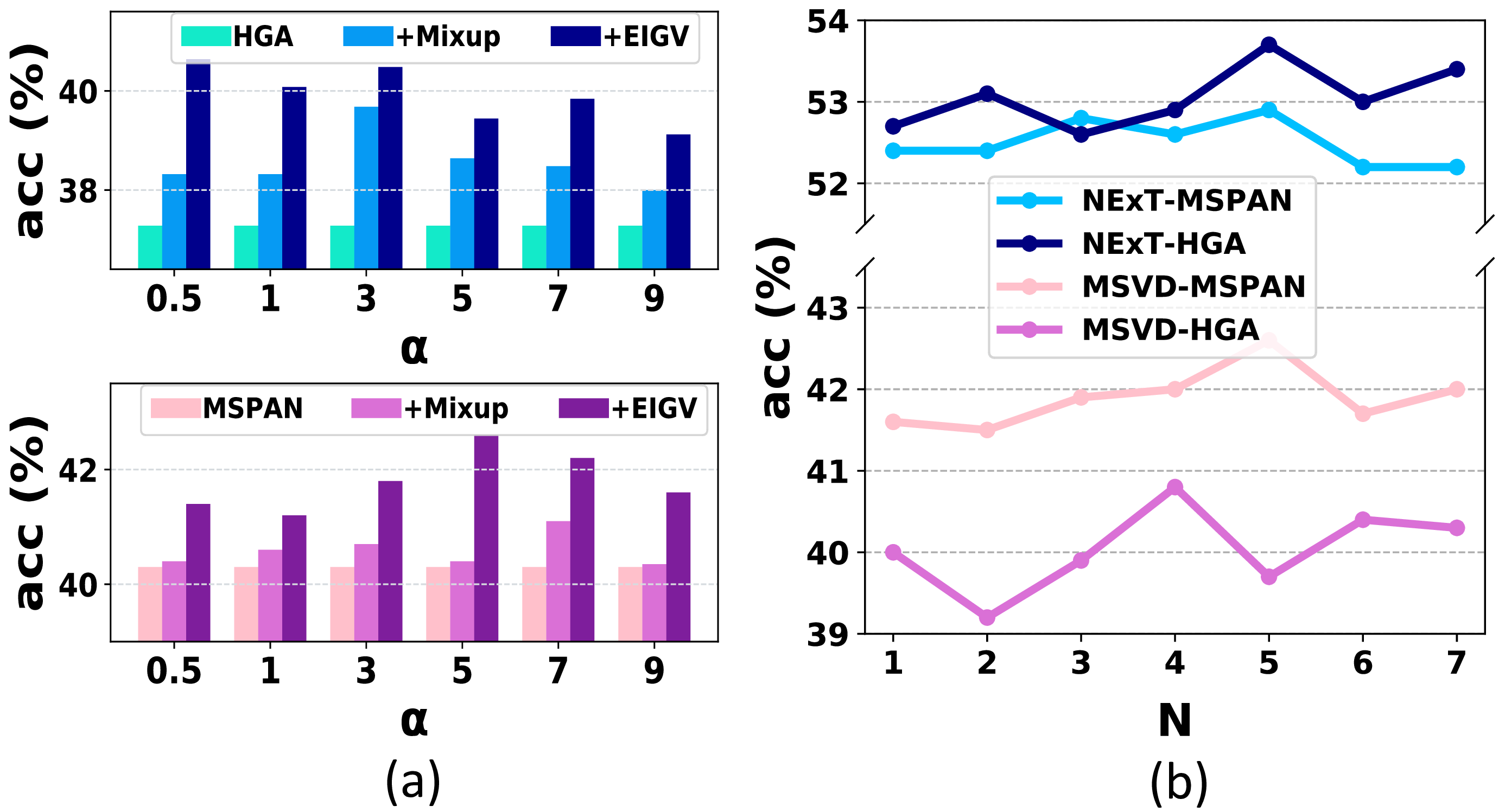}
\vspace{-7pt}
\caption{Hyperparameter analysis. (a) Study of $\alpha$ on MSVD-QA, which controls the equivariant mixing ratio by $\lambda_0\sim\text{Beta}(\alpha,\alpha)$. Performance of two EIGV enhanced models --- HGA (top) and MSPAN (bottom) are reported, alongside the SoTA backbone and mixup augmented performances.  
(b) Study on the impact of the negative sample number N, where EIGV with two backbones (\ie MSPAN and HGA) on two benchmark datasets (NExT-QA and MSVD-QA) are reported.}
\label{w3-fig:neg}
\end{figure}

\vspace{5pt}
\noindent{\textbf{What are the effect of EIGV's components?}}
To comprehensively understand the reasoning mechanism of EIGV, we poke its structure with careful scrutiny. Specifically, we explore the effectiveness of the proposed intervener and disruptor by analyzing their performance with different backbones on two benchmarks. We report the corresponding performances in \cref{w3-tab:ablation} and summarize our findings as follows:
\begin{itemize}[leftmargin=*]

\vspace{5pt}
\item \noindent \textbf{Effectiveness of Intervener.} \label{w3-exp:intervener}
We first testify the substantial efficacy of the intervener by comparing its permanence ( ``$+$Intervener'' ) to the backbone. This brings constant gains across different settings (+1.2\%$\sim$3\%), which demonstrates the stability of our design. Then, we compare the result of the intervener with the conventional mixup augmentation \cite{mixup}, which can be considered as a simplified case of the interventer that only applies the equivariant intervention to the entire training video. The result shows that our design outperforms the conventional mixup in all cases. This manifests that the benefit of invariant intervention is fundamental, and the functionality of invariance and equivariance principle are mutually reinforced.

\vspace{5pt}
\item \noindent \textbf{Effectiveness of Disruptor.} 
We validate the disruptor design by investigating its components --- the substitution on video (``$+$Disrupt-V'') and the permutation on question (``$+$Disrupt-Q''), respectively. 
Albeit moderate, improvement on (``$+$Disrupt-V'') shows that stressing causal scene can benefit visual robustness.
A similar trend also applies to ``$+$Disrupt-Q'' as well, the constant improvement in all settings shows that acknowledging artificial corrosion in ($v,q$) matching can strengthen vision-language alignment. 
Furthermore, the overall result on ``$+$Disrupt'' shows that the advancement of ``$+$Disrupt-V'' and ``$+$Disrupt-Q'' can be amplified by further integration. 
\end{itemize}

\input{work3/tab/unifeat}

\noindent{\textbf{What are the effects of different feature settings?}}
To answer this question, we perform uni-feature tests for the visual representation. Concretely, instead of combining the appearance and motion features together and then manipulating them as a whole, we run ablative experiments on each of them solely.  
As shown in \cref{w3-tab:unifeat}, under all circumstances, EIGV can improve models trained with appearance and motion features in equivalent magnitude, even though the appearance feature is demonstrated to be more visually informative in backbone comparison. This verifies that our improvement is ascribed to both feature modes rather than accessing only one of them.


\noindent{\textbf{What are the effects of hyper-parameters?}}
Justifying a reliable design requires a sensitivity test on its hyper-parameters. As shown in \cref{w3-fig:neg}, we probe the potency of EIGV by investigating the distribution of the equivariance mixing ratio and the number of negative samples. Our observations are as follows:

\begin{itemize}[leftmargin=*]
\item \cref{w3-fig:neg}a shows how EIGV performs compared to the SoTA backbone and the conventional mixup augmentation. Specifically, we adjust $\alpha$ to vary the distribution that the equivariant mixing ratio $\lambda_0$ draws from, and cross-check the performance of EIGV (``+EIGV'') against its counterparts (``SoTA Backbone'' and ``+HGA'') on two backbone models. Mixup, despite some improvement, its generalization is limited by the choice of the backbone. For MSPAN backbone, even the heavily tuned $\alpha$ fails to make a reasonable improvement. In contrast, EIGV successively outperforms mixup augmentation and backbone methods in every $\alpha$ setting, which recalls our finding in \cref{w3-exp:intervener} and justifies the necessity of the environment-invariance principle.      

\item \cref{w3-fig:neg}b demonstrates how the performance of EIGV varies as the number of negative sample increase. We notice that the predictive curves keep rising until $N$ reaches around five, which indicates that EIGV learns distinctive grounding rationale as more negative samples are considered. This is in line with the finding in the contrastive learning community that additional negative pairs bring about more desirable outcome \cite{he2019moco}. 
\end{itemize}

\subsection{Quantitative Study (RQ3)}

\begin{figure*}[t]
  \centering
  \includegraphics[width=1.0\textwidth]{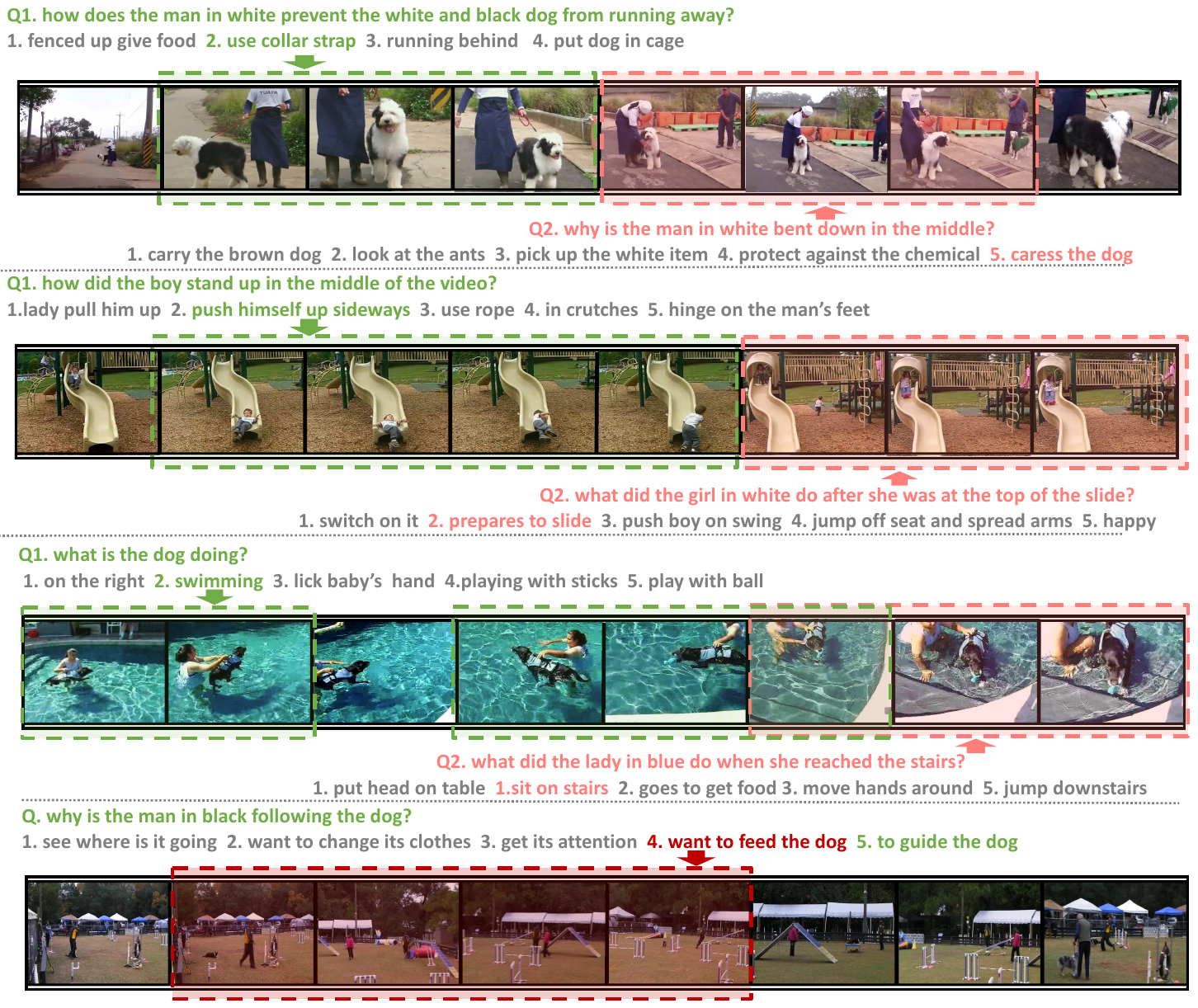}
 	\vspace{-13pt}
  \caption{Visualization of discovered grounding rationale. Each row comes with a video instance and two questions that target at different scene.  The \textcolor[rgb]{0,0.5,0}{green} and \textcolor[RGB]{225,152,150}{pink} windows indicate the rationales for the corresponding questions. The last row shows a failure case, where \textcolor[rgb]{0.5,0,0}{red} denotes the wrong prediction.}
  \label{w3-fig:case-study}
\end{figure*}

By nature, EIGV is equipped with intrinsic visual interpretability. To capture the learning insight of EIGV, we inspect the predictive answer of some video instances along with their grounded interpretations and show the visualization in \cref{w3-fig:case-study}, where each row provides a video instance and two questions that emphasize the visual content in different temporal spans. Notably, even for the same video instance, EIGV is able to accredit different scenes for the different questions. Nonetheless, we also observe the insufficient grounding result in Row 3 Q1, where the grounding result partially covers the dog swimming scene, while the whole video is answerable to the question. In addition, we also provide a failure case in the last row. Although the grounding result is reasonable, the model still can not understand why the man is following the dog, as a result, the model turns to the most likely answer according to the dataset prior, which is feeding the dog. To resolve this issue, the model should be provided with commonsense knowledge of dog racing competition,  a large language model might be able to mitigate this issue.

%% file: work3/tab/main.tex
\setlength{\tabcolsep}{5pt}
\begin{table}[t]
  \centering
  \caption{Comparison with SoTAs. Our results are highlighted.}
    \resizebox{0.8\linewidth}{!}{
    \begin{tabular}{cc|ccc}
    \toprule
          & Model & MSVD-QA & MSRVTT-QA & NExT-QA \\
    \midrule
    \midrule
    \multirow{9}[6]{*}{SoTAs} & AMU \cite{DBLP:conf/mm/XuZX0Z0Z17}   & 32.0    & 32.0    & - \\
          & HME \cite{fan2019heterogeneous} \  & 33.7  & 33.0    & 49.2 \\
\cmidrule{2-5}          & B2A  \cite{park2021bridge}   & 37.2  & 36.9  & - \\
          & L-GCN \cite{huang2020locationaware} & 34.3  & 33.7  & 49.5 \\
\cmidrule{2-5}          & HCRN \cite{le2021hierarchical}  & 36.1  & 35.6  & 48.9 \\
          & PGAT \cite{pgat} & 39.0    & 38.1  & - \\
          & HOSTR \cite{dang2021hierarchical} & 39.4  & 35.9  & - \\
          & HQGA \cite{hqga} & \underline{41.2}  & \underline{38.6}  & \underline{51.8} \\
\cmidrule{2-5} & IGV \cite{IGV} & 40.8  & 38.3  & 51.3 \\
    \midrule
    \midrule
    \multirow{6}[6]{*}{Ours} & Co-Mem \cite{gao2018motionappearance}& 34.6  & 35.3  & 48.5 \\
          & \white{66666}$+$ EIGV   & $\;$\;\; 39.8 $^{+5.2}$ &  \white{6666}37.2 $^{+1.9}$ &  \white{6666}50.7 $^{+2.2}$ \\
\cmidrule{2-5}          &  HGA \cite{jiang2020reasoning}   & 36.6  & 36.7  & 50.0 \\
          & \white{66666}$+$ EIGV   &  \white{6666}40.8 $^{+4.2}$&  \white{6666}38.5 $^{+1.8}$ &  \white{6666}\textbf{53.7} $^{+3.7}$ \\
\cmidrule{2-5}          & MSPAN \cite{mspan} & 40.3  & 38.0    & 50.7 \\
          & \white{66666}$+$ EIGV   &  \white{6666}\textbf{42.6} $^{+2.3}$ &  \white{6666}\textbf{39.3} $^{+1.3}$ &  \white{6666}52.9 $^{+2.2}$ \\
    \bottomrule
    \end{tabular}%
    }
    \vspace{-5pt}
  \label{w3-tab:main}%
\end{table}%

%% file: work3/tab/ablation.tex
\setlength{\tabcolsep}{10pt}
\begin{table}[t]
  \centering
  \small
  \caption{Evaluation on the effectiveness of sub-modules}
    \resizebox{0.8\linewidth}{!}{
    \begin{tabular}{c|cc|cc}
    \toprule
    \multirow{2}[2]{*}{Ablation} & \multicolumn{2}{c|}{MSVD-QA} & \multicolumn{2}{c}{NExT-QA} \\
          & MSPAN & HGA   & MSPAN & HGA \\
    \midrule
    \midrule
    SoTA Backbone & 40.3  & 36.6  & 50.7  & 50.0 \\
    \midrule
    $+$ Mixup \cite{mixup} & 41.0    & 38.3  & 52.0    & 51.8 \\
    $+$ Intervener & 41.5  & 39.6  & 52.5  & 52.7 \\
    \midrule
    $+$ Disruptor & 40.9  & 37.6  & 51.0    & 51.1 \\
    \white{66666} $+$ Disrupt-Q & 40.6  & 37.0    & 50.8  & 51.0 \\
    \white{66666} $+$ Disrupt-V & 40.7  & 37.3  & 51.0    & 50.8 \\
    \midrule
    EIGV & \textbf{42.6} & \textbf{40.8} & \textbf{52.9} & \textbf{53.7} \\
    \bottomrule
    \end{tabular}%
    }
  \label{w3-tab:ablation}%
\end{table}%

%% file: work3/tab/unifeat.tex
\begin{table}[t]
  \centering
  \caption{Study of feature setting. ``APP'' and ``MOT'' denotes using appearance and motion feature individually. }
    \resizebox{0.8\linewidth}{!}{
    \begin{tabular}{cc|cc|cc}
    \toprule
    \multicolumn{2}{c|}{\multirow{2}[2]{*}{Method}} & \multicolumn{2}{c|}{MSVD-QA} & \multicolumn{2}{c}{NExT-QA} \\
    \multicolumn{2}{c|}{} & MSPAN & HGA   & MSPAN & HGA \\
    \midrule
    \midrule
    \multirow{2}[2]{*}{APP} & SoTA Backbone & 40.1  & 35.0    & 49.7  & 48.3 \\
          & \white{66666}$+$ EIGV   & \textbf{41.0}    & \textbf{39.5}  & \textbf{52.0}    & \textbf{52.4} \\
    \midrule
    \multirow{2}[2]{*}{MOT} & SoTA Backbone & 37.8  & 33.6  & 49.4  & 47.5 \\
          & \white{66666}$+$ EIGV   & \textbf{39.3}  & \textbf{38.5}  & \textbf{51.1}  & \textbf{51.7} \\
    \bottomrule
    \end{tabular}%
    }
  \label{w3-tab:unifeat}%
\end{table}%

%% file: work3/sec/7_conclusion.tex
\section{conclusion}
In this chapter, we presented EIGV --- a model-agnostic explainer, that empowers the SoTA VideoQA model with intrinsic interpretability and robustness. In light of the causality, we formulated our learning principles --- causal-equivariance and environment-invariance by incorporating three constituents, the grounding indicator, the intervener, and the disruptor, which manage a robust rationale discovery. Experiments across three benchmarks validate EIGV's fulfillment in both interpretation and accuracy.


%% file: work4/main.tex
\chapter{Discovering Spatio-Temporal Rationales for Video Question Answering} 
\label{cha:transtr}

\input{work4/sec/0_abstract}

\input{work4/sec/1_intro}

\input{work4/sec/related}
\input{work4/sec/2_preliminary}

\input{work4/sec/3_method}
\input{work4/sec/4_exp}

\input{work4/sec/5_conclusion}

%% file: work4/sec/0_abstract.tex
Although \cref{cha:igv} and \cref{cha:eigv} have explored causal patterns at the frame-level, such rationales at object-level have been left untouched till now. In this chapter, we solve complex video question answering (VideoQA) which features long videos containing multiple objects and events at different time. To tackle the challenge, we highlight the importance of identifying question-critical temporal moments and spatial objects from the vast amount of video content. Towards this, we propose a \textbf{S}patio-\textbf{T}emporal \textbf{R}ationalization (STR), a differentiable selection module that adaptively collects question-critical moments and objects using cross-modal interaction.
The discovered video moments and objects then served as grounded rationales to support answer reasoning. Based on STR, we further propose TranSTR, a Transformer-style neural network architecture that takes STR as the core and additionally underscores a novel answer interaction mechanism to coordinate STR for answer decoding. Experiments on four datasets show that TranSTR achieves new state-of-the-art (SoTA). Especially, on NExT-QA and Causal-VidQA which feature complex VideoQA, it significantly surpasses the previous SoTA by 5.8\% and 6.8\%, respectively. We then conduct extensive studies to verify the importance of STR as well as the proposed answer interaction mechanism. With the success of TranSTR and our comprehensive analysis, we hope this work can spark more future efforts in complex VideoQA.  

%% file: work4/sec/1_intro.tex
\section{Introduction}\label{w4-sec:intro}
The great success of self-supervised pretraining with powerful transformer-style architectures \cite{devlin2018bert,deberta,DBLP:conf/icml/KumarIOIBGZPS16,fu2021violet,yang2021just,zellers2021merlot} has significantly boosted the performance of answering simple questions (\eg, ``what is the man doing'') on short videos (\eg, 3$\sim$15s) \cite{jang2017tgif,DBLP:conf/mm/XuZX0Z0Z17}. The advances thus point towards complex video question answering (VideoQA), that features long video containing multiple objects and events \cite{next-qa,causalvid,zhong2022video}. Compared with simple VideoQA, complex VideoQA poses several unique challenges: 

\begin{figure}
  \centering
  \begin{subfigure}{\linewidth}
  \includegraphics[width=0.98\linewidth]{ 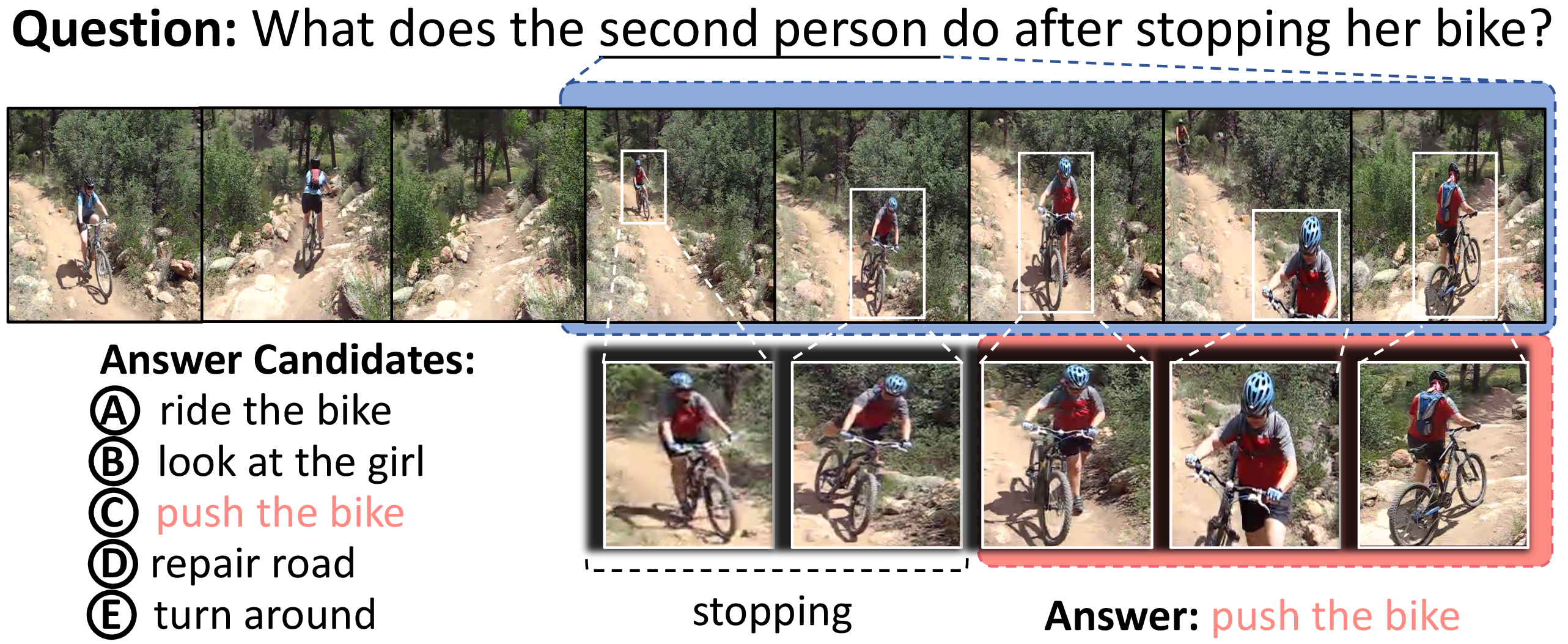}\par
  \caption{A example of long video (52s) with multiple objects, the question-related frames and objects are located to support the reasoning.}
  \label{w4-fig:1a} 
  \end{subfigure}

  \begin{subfigure}{0.48\linewidth}
  \includegraphics[width=\linewidth]{ 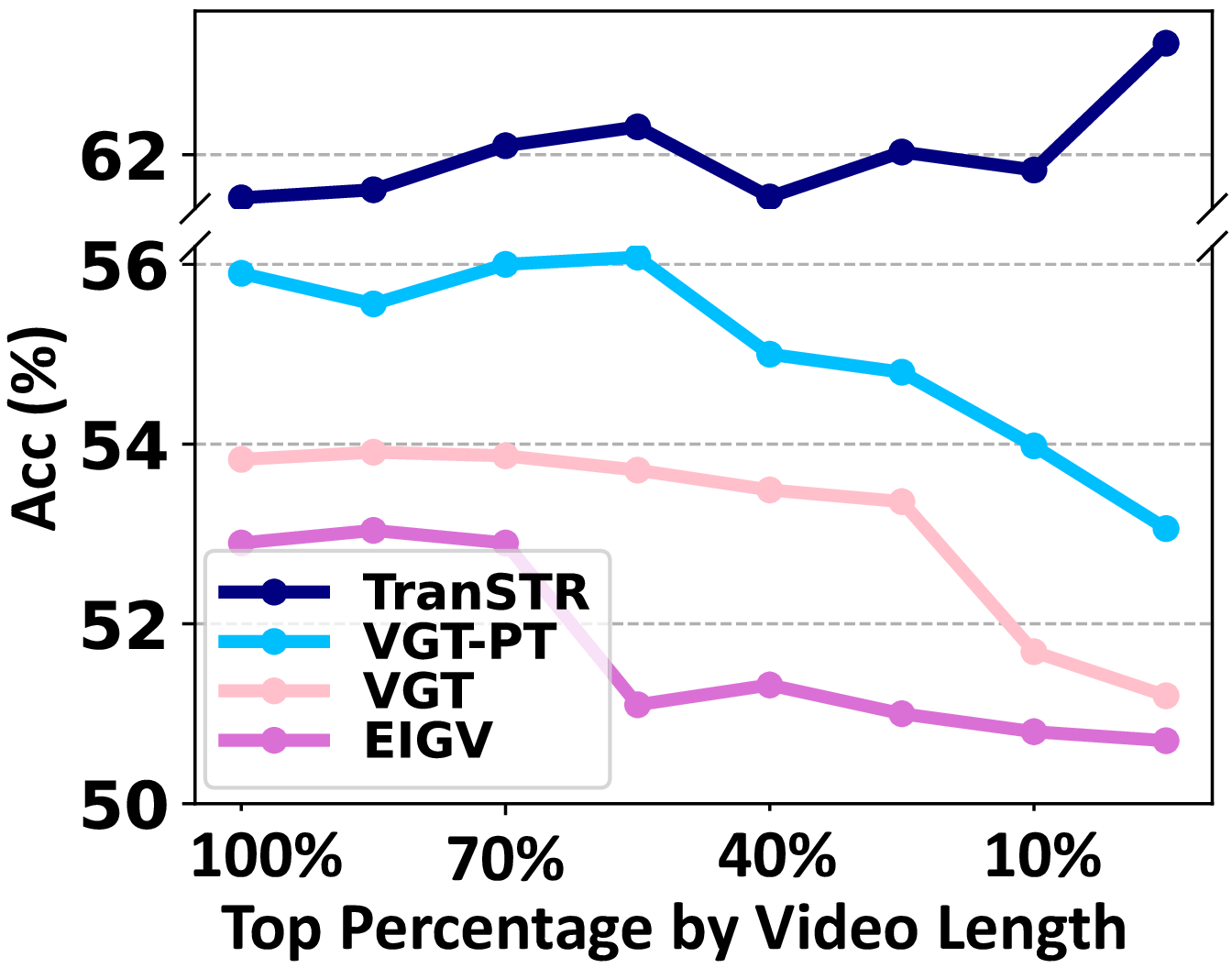}
  \caption{Accuracy by video length.}
  \label{w4-fig:1b} 
\end{subfigure}
\begin{subfigure}{0.47\linewidth}
  \includegraphics[width=\linewidth]{ 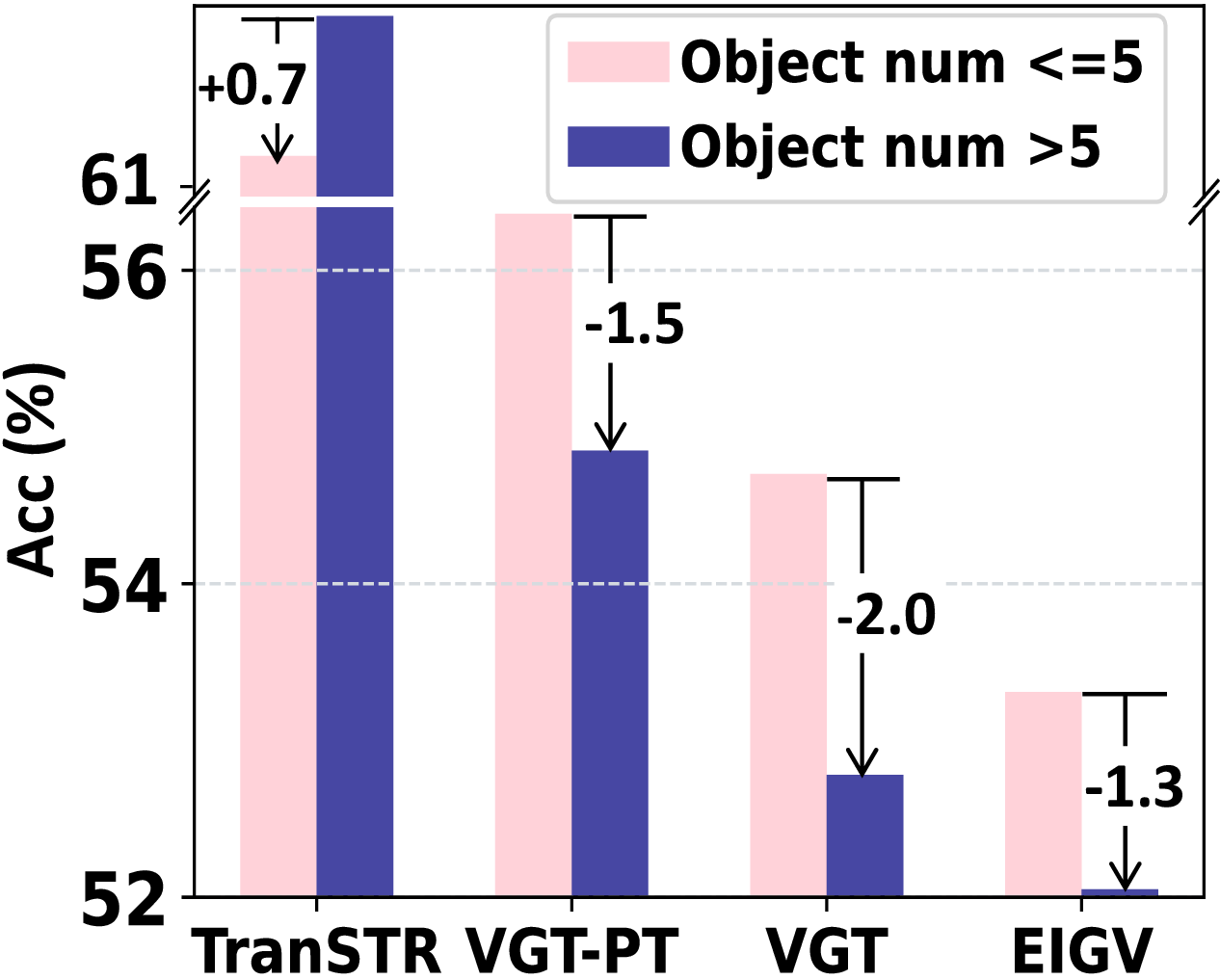}
  \caption{Accuracy by object number.}
  \label{w4-fig:1c}
\end{subfigure}
\caption{(a) Illustration of complex VideoQA, in which the videos are longer and the questions involve multiple objects and events at different time. (b) Prediction accuracy grouped by video length. We first sort all samples by video length, then select top x\% to calculate accuracy.
(c) Accuracy grouped by whether the video has more than 5 objects. All results are reported on NExT-QA test set \cite{next-qa}. Figure (b) and (c) show that our method TranSTR performs much better than the previous SoTAs for question answering of long videos with multiple objects.}
\label{w4-fig:1}
\end{figure}

1) \textbf{Longer videos with multiple objects interacting differently at different time.} The long video and rich visual content indispensably bring more background scenes that include  massive question-irrelevant video moments and objects. For example, to answer the question in \cref{w4-fig:1a}, only the interaction between object ``person'' and ``bike'' on the last three frames encloses the answer information, leaving the massive rest as background. These backgrounds, if not filtered properly, will overwhelm the critical scene and interfere with answer prediction. 
%
2) \textbf{Harder negative answer as distractors.}
Negative answers in complex VideoQA are typically tailored for each video instance. Due to the massive video content, the vast question-irrelevant scene provides an ideal foundation to build a hard negative candidate as distractor.
The hard negatives are very similar to the correct answer but correspond to a different video moment or object.
For example, the answer candidate ``A.ride the bike'' of \cref{w4-fig:1a}, though irrelevant to the question,  corresponds to a large part of the video. 
As a result, these distractors can seriously derail the prediction if not properly modeled.

In light of the challenges, current methods (both pretrained and task-specific architectures) hardly perform well on complex VideoQA. 
In \cref{w4-fig:1b} and \cref{w4-fig:1c}, we use the video length and number of objects\footnote{We acquire the object number using annotation of video relation detection dataset \cite{shang2021video}, which shares same source video as NExT-QA.} to indicate the complexity of the video questions. We can see that current methods suffer a drastic performance drop when video length increases or more objects are involved. The reason can be provided from two aspects:
%
%
\textbf{First}, confronting long video and multiple objects, pretrained methods suffer from a domain gap. Because they are typically pretrained with short videos and simple questions, \cite{DBLP:conf/icml/KumarIOIBGZPS16, lei2021less} 
where the answer can be easily captured via a static frame, without fine-grained reasoning over multiple objects in a long video. 
While recent task-specific methods exploit object-level representation for fine-grained reasoning \cite{hostr,pgat,hqga,vgt}, they exhibit limited generalization ability, as they handle different videos with only a fixed number of frames and objects, and cannot adapt to lengthy and varied visual content,
which rigidity undermines their adaptability to a wide range of video content.
\textbf{Second}, to model the answer candidate, prevailing designs \cite{jang2017tgif,gao2018motionappearance,fan2019heterogeneous,hga,hqga} append the candiate to the question and treat the formed question-answer sequence as a whole for cross-modal learning. However, this makes the answer candidate directly interact with the whole video content, which gives rise to a strong spurious correlation between the hard negative candidates (\eg ``A. ride the bike" in \cref{w4-fig:1a}) and the question-irrelevant scenes (\eg `` riding'' scene in first three frames), leading to a false positive prediction. 
%

In this regard, we propose TranSTR, a Transformer-style VideoQA architecture that coordinates a Spatio-Temporal Rationalization (STR) with a more reasonable video-text interaction pipeline for candidate answer modeling.
%
%
%
STR first temporally collects the critical frames from a long video, followed by a spatial selection of objects on the identified frames. By further fusing the selected visual objects and frames via light-weight reasoning module, we derive spatial and temporal rationales that exclusively support answering. 
%
In addition to STR, we circumvent the spurious correlation in the current modeling of answer candidates by formulating a more reasonable video-text interaction pipeline, where the question and answer candidates are separately (instead of being appended as a whole) fed to the model at different stages. 
Specifically, before rationale selection, only the question is interacted with the video to filter out the massive background content while keeping question-critical frames and objects. After that, a transformer-style answer decoder introduces the answer candidates to these critical elements to determine the correct answer.
Such a strategy prevents the interaction between the hard negative answers and the massive background scenes, thus enabling our STR to perform better on complex VideoQA (see TranSTR in \cref{w4-fig:1b} and \cref{w4-fig:1c}). 
It is worth noting that STR and the answering modeling are reciprocal. Without STR's selection, all visual content will still be exposed to answer candidates. Without our answer modeling, STR could identify the question-irrelevant frame and object as critical.
Thus, the success of TranSTR is attributed to the integration of both.

Our contributions are summarized as follows:

\begin{itemize}[leftmargin=*]
\item \textbf{Analysis of Complex VideoQA Challenges}: Our research conducts an in-depth analysis of the unique challenges and necessities inherent in complex VideoQA. A key finding is the crucial role of identifying and understanding spatio-temporal rationales, as well as the importance of avoiding spurious correlations in modeling candidate answers. This analysis benefits the field by providing a clearer understanding of what makes complex VideoQA distinct, thereby guiding more effective and targeted model development.

\item \textbf{Introduction of the TranSTR Framework}: We introduce TranSTR, a novel framework that incorporates a Spatio-Temporal Rationalization (STR) module along with an innovative candidate answer modeling strategy. The key benefit of TranSTR is its enhanced ability to accurately model and rationalize spatio-temporal data, leading to more precise answer generation in VideoQA tasks. Additionally, the effectiveness of our answer modeling strategy is independently verified, showing its potential to boost performance in existing VideoQA models. This highlights TranSTR's versatility and adaptability, making it a valuable tool for improving a wide range of VideoQA applications.

\item \textbf{Superior Performance on Benchmark Datasets}: Our extensive experiments demonstrate that TranSTR achieves State-of-the-Art (SoTA) performance across four major benchmark datasets, particularly excelling in datasets featuring complex VideoQA tasks (e.g., NExT-QA +5.8\%, CausalVid-QA +6.8\%). This outstanding performance not only proves the efficacy of TranSTR but also its significant impact in addressing the complexities of VideoQA. It sets new performance standards, especially in tackling the more challenging aspects of VideoQA, thereby contributing substantially to the advancement of the field.
\end{itemize}

%% file: work4/sec/related.tex
\section{Related Works}\label{w4-sec:related}

\noindent\textbf{Video Question Answering (VideoQA).}
Substantiated as a fundamental extension of ImageQA, VideoQA has enlarged its definition by adding a temporal extension. 
According to the pre-extracted feature granularity, existing methods either use the frame-level features or incorporate object features for fine-grain reasoning. 
In task-specific designs, Focus on simple questions and short videos, earlier efforts tend to model the video sequence as a visual graph using purely frame features. As the pioneer of graph-based structure, \cite{hga} and \cite{park2021bridge} build their typologies based on the heterogeneity of input modality,  while \cite{mspan} enables progressive relational reasoning between multi-scale graphs. 
Recently, the emergence of complex VideoQA benchmarks \cite{next-qa, causalvid} has prompted studies on long video with multiple visual entities. 
In this regard, Another line of research has prevailed by processing video as multi-level hierarchy. 
\cite{hcrn} first build a bottom-up pathway by assembling information first from frame-level, then merging to clip-level. The following works \cite{hostr, hqga} extend the hierarchy into the object-level, where a modular network is designed to connect objects on the same frame. Most recently, \cite{vgt} establish its improvement by enabling relation reasoning in a sense of object dynamics via temporal graph transformer.
Despite effectiveness, the current designs unanimously rely on a fixed number of frames and objects, which severely compromise their transferability across diverse video instances.
In sharp contrast, our method works in a fully adaptive manner to explicitly select frames and objects for reasoning over different circumstances, which demonstrates superior generalization ability.

\vspace{5pt}
\noindent\textbf{Rationalization.}
In pursuit of explainability, the recent development of DNN is encouraged to reveal the intuitive evidence of their prediction, \ie the rationales. 
As one of the prevailing practices, the rationalization has been extended from the NLP community\cite{DBLP:conf/kdd/Ribeiro0G16} to the Graph \cite{DIR} and Vision field \cite{DBLP:conf/cvpr/ZhangYMW19}. 
Recently, this development also stems from the multi-modal community.
\cite{DBLP:conf/cvpr/ParkHARSDR18} and \cite{DBLP:conf/cvpr/DuaKB21} proposes ImageQA-based tasks that inquire about additional textual evidence, \cite{causalvid} brings this idea to the videoQA.
Despite the progress, the recent solution focus on the rationale only at frame level, and they either require a rationale finder with heavy computation overhead \cite{IGV} or needs to be trained in a data-hungry contrastive manner \cite{EIGV}. TranSTR, however, identifies both critical frames and objects from an efficient cross-modal view. 
Also, distinct from the token reduction method in transformer literature, which trades accuracy for efficiency. Rationalization intends to improve performance \cite{rationalization-robustness}. The intuition behind is that, if a model can find the causal part, they have the potential to ignore the noise. 


\begin{table}[h]
  \centering
  \caption{Key Related Papers}
    \begin{tabular}{lp{10cm}} 
    \toprule
    Paper & Remark \\
    \midrule
    \cite{IGV,EIGV} & These previous works, explore the visual rationale at the frame level, while this work investigates a more fine-grain rationale at object level.
    \\
    \midrule
    cite{rationalization-robustness} & Existing efficient transformer work also explores token reduction method. But it generally trades accuracy for efficiency. While our rationalization intends to improve performance. \\
    \bottomrule
    \end{tabular}%
\end{table}%

%% file: work4/sec/2_preliminary.tex
\section{Preliminaries}
\label{w4-sec:preliminaries}

\vspace{5pt}
\noindent \textbf{Modeling.}
Given the video $V$ and the question $Q$, the VideoQA model $\phi(V,Q)$ aims to encapsulate the visual content and linguistic semantics and choose the predictive answer $\hat{A}$ from the answer candidates.
%
Typically, an entropy-based risk function $\mathcal{L}(\phi(V,Q), A)$ is applied to approach the ground-truth answer ${A}$.

\vspace{5pt}
\noindent \textbf{Data representation.}
We uniformly sample $T$ clips and keep the middle frame of each clip to represent a video.
Then, for each frame, we extract a frame feature $\Mat{f}_t$ via a pretrained image recognition backbone and $S$ object features $\Mat{o}_{t,s}$ using pretrained object detector, where $t,s$ denotes the $s$-th object on the $t$-th frame.
To represent the text, we encode the question as a sequence of $L$ tokens using a pretrained language model and obtain a textual representation $\Mat{q}_l$ for each of them. The visual backbones are frozen during training while the language backbone is finetuned end-to-end as in \cite{vgt}.
To project the representations into a common $d$-dimensional space, we apply a three linear mappings on $\Mat{f}_t$, $\Mat{o}_{t,s}$, and $\Mat{q}_l$, respectively, and thus acquire 
$\Mat{F}\!=\!\left\{ \Mat{f}_t \right\}_{t=1}^{T} \!\in\!\Space{R}^{T\times d}$, 
$\Mat{O}\!=\!\left\{ \Mat{o}_{t,s}\right\}_{t=1,s=1}^{T,S}\!\in\!\Space{R}^{T\times S \times d}$, 
and 
$\Mat{Q}\!=\!\left\{ \Mat{q}_l \right\}_{l=1}^{L}\!\in\!\Space{R}^{L\times d}$ to denote the frame, object, and question features, respectively.

%% file: work4/sec/3_method.tex
\section{Method}\label{w4-sec:method}


\begin{figure*}
\begin{minipage}{0.65\textwidth}
\begin{subfigure}{\textwidth}
\includegraphics[width=\linewidth]{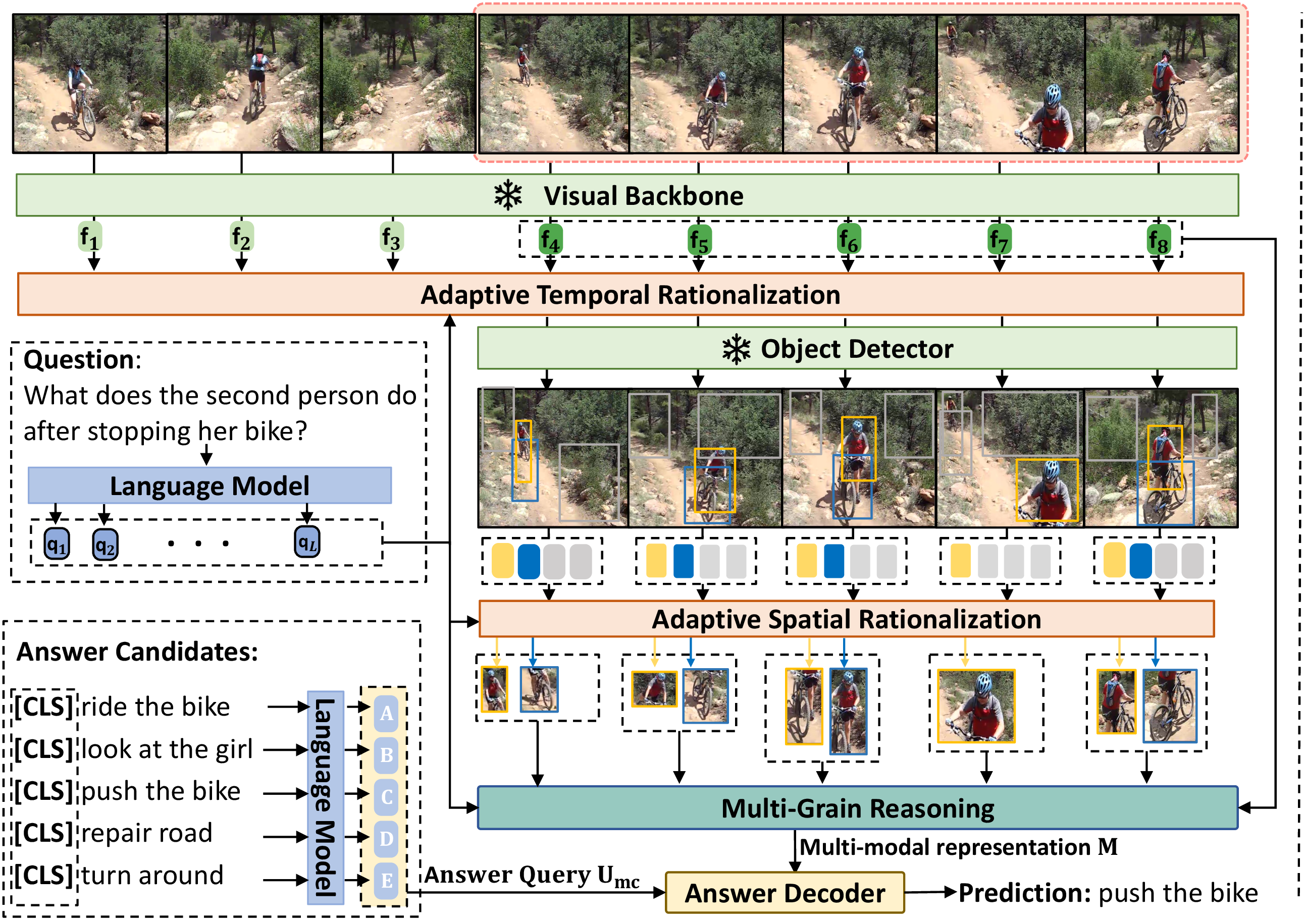}
\vspace{-10pt}
\caption{Overview of TranSTR} \label{w4-fig:2a}
\end{subfigure}
\end{minipage}
\begin{minipage}{0.34\textwidth}
\begin{subfigure}{\textwidth}
\includegraphics[width=1\linewidth]{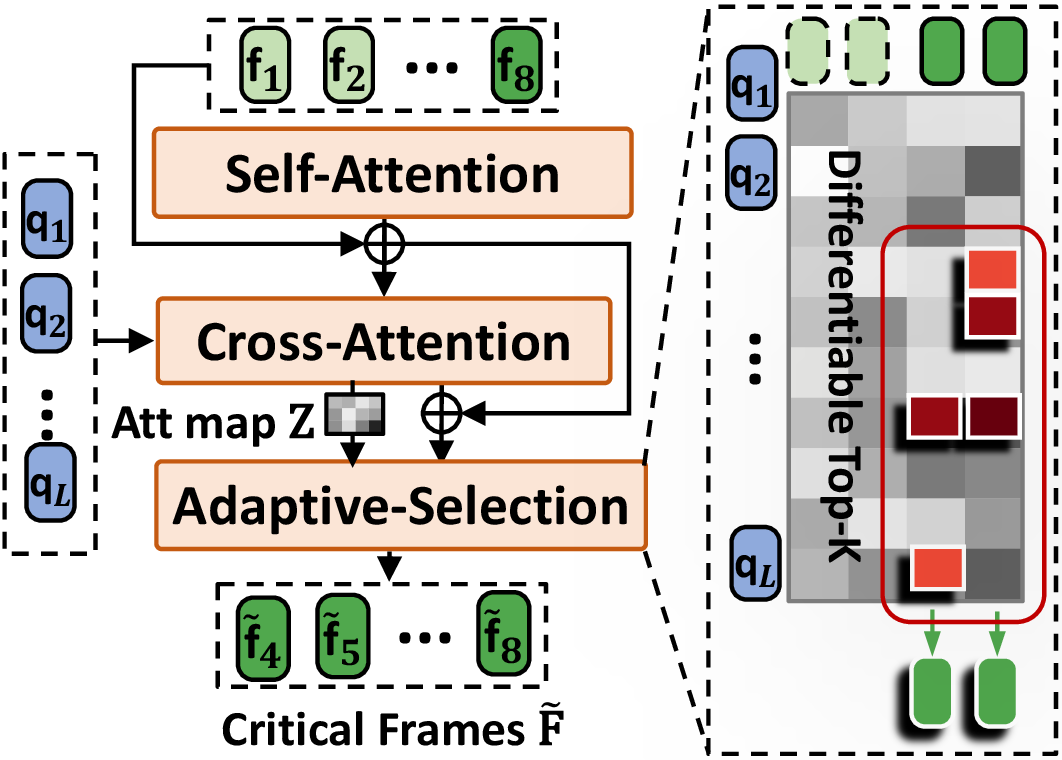}
\vspace{-15pt}
\caption{} \label{w4-fig:2b}
\end{subfigure}

\begin{subfigure}{\linewidth}
\includegraphics[width=1\linewidth,]{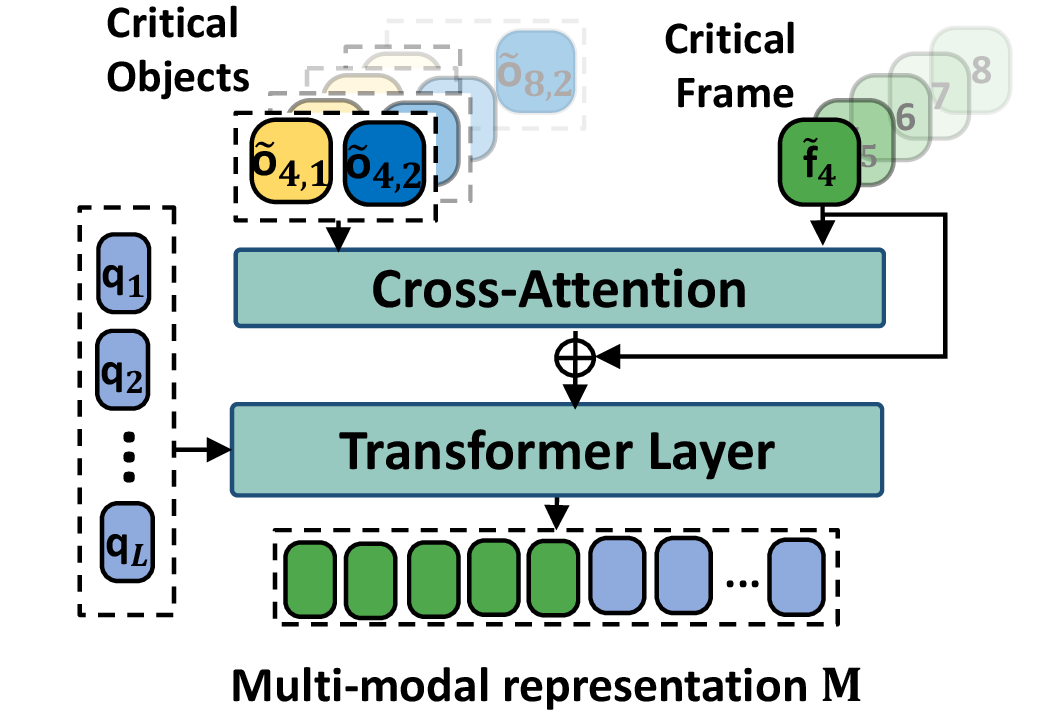}
\vspace{-15pt}
\caption{} \label{w4-fig:2c}
\end{subfigure}
\end{minipage}

\vspace{-5pt}
\caption{TranSTR (a) contains three components: the Spatio-Temporal Rationalization (b) which adaptively selects critical frames and objects, the Multi-Grain Reasoning (c) which forms multi-modal representation by integrating the critical frames and objects together with the question semantics, and the Answer Decoder which bring in answer candidates and make a prediction based on the multi-modal representation. 
%
Notably, STR follows a ``Temporal then Spatial'' rationalization process, where the two steps leverage a similar deisgn.  In (b), we illustrate this design using Adaptive Temporal Rationalization.}
\vspace{-0.5cm}
\end{figure*}

As shown in \cref{w4-fig:2a} TranSTR consists of three main components: Spatio-Temporal Rationalization (STR), Multi-Grain Reasoning (MGR), and Answer Decoder. 
First, STR follows a two-step selection process, where Adaptive Temporal Rationalization (TR) is first  performed for frame selection, followed by object selection via Adaptive Spatial Rationalization (SR).
Next, the selected frames and objects are then combined through MGR, which generates multi-modal representations by fusing with the question.  
Finally, based on the multi-modal representations, the answer decoder takes the combination of answer candidates as queries and predicts an answer.
In this section, we provide a detailed illustration of each module. 

\subsection{Spatio-Temporal Rationalization (STR)}

STR aims to find the question-critical frames and objects in a fully adaptive and differentiable manner, which comprises two components: an Adaptive Temporal-Rationalization (TR) that identifies the critical frames and an Adaptive Spatial-Rationalization (SR) that pinpoints the critical objects on the identified frames.

\subsubsection{Adaptive Temporal Rationalization (TR)}
To identify the critical frames from the video, TR takes the encoded video frames $\Mat{F}\!\in\!\space{R}^{T\times d}$ as input and adaptively selects question-critical frames from a cross-modal view.
As shown in \cref{w4-fig:2b}, a self-attention layer is first adopted to contextualize $\Mat{F}$. Then, a cross-attention is applied by taking contextualized frame feature $\Mat{F}'$ as query and question embedding $\Mat{Q}$ as key and value, which yields the frame tokens $\Mat{F}''\!\in\!\space{R}^{T\times d}$ and the cross-attention map $\Mat{Z}\!\in\!\space{R}^{T\times L}$.
\begin{gather}
    \Mat{F}'=\text{Self-Attention}(\Mat{F}) + \Mat{F}, \label{w4-eq:1}\\
    \Mat{F}'', \Mat{Z}=\text{Cross-Attention}(\Mat{F}', \Mat{Q}) + \Mat{F}'. \label{w4-eq:2}
\end{gather}
For brevity, we omit the superscript in $\Mat{F}''$ and use $\Mat{F}$ to denote the resulting frame tokens. 

Naturally, each value in the cross-attention map indicates the cross-modal interaction activeness between a frame and a question token.
To enable an adaptive frame selection that caters to different videos, we collect the $K_f$ interactions of the highest attention score from the cross-attention map $\Mat{Z}$, then gather their corresponding frame tokens $\tilde{\Mat{F}} \!\in\!\space{R}^{C\times d}$ as a subset of $\Mat{F}$, where $C$ is the number of critical frames. This process is formally given by:
\begin{gather} \label{w4-eq:3}
    \tilde{\Mat{F}}=\text{Adaptive-Selection}_K(\Mat{F}, \Mat{Z}) \quad \text{s.t.}\,K=K_f.
\end{gather}
Notably, by gathering 1-D tokens from the 2-D interaction view, we enable an adaptive collection of $\tilde{\Mat{F}}$ with much fewer tokens being selected as critical frames (\ie $C \!\ll\! T$ and $C \!\ll\! K_f$).
It is worth noting that the interaction selection via vanilla hard Top-K produces a discrete selection, making it inapplicable for end-to-end training. 
We address this issue by adopting a differentiable Top-K using the perturbed maximum method \cite{pertub}, 
which has empirically shown advantages over other differentiable technique (\cf \cref{w4-tab:ablation-loss})

\subsubsection{Adaptive Spatial Rationalization (SR)}
Given the embedding of the selected frames $\tilde{\Mat{F}}$, SR aims to pinpoint the question-critical objects in each frame. 
To achieve that, SR enable an adaptive object selection similar to \cref{w4-eq:1,w4-eq:2,w4-eq:3}.
Specifically, for the $t$-th critical frame $\tilde{\Mat{f}}_t$, we first feed SR with fixed $S$ object features detected on that frame, then collect top $K_o$ interactions from its cross-modal attention map with question embedding. Finally, by gathering their corresponding object tokens, we obtain the critical object feature $\tilde{\Mat{o}_t} \!\in\!\space{R}^{C_t\times d}$, where $C_t$ denotes the number of critical objects on $t$-th critical frame.
%
It is worth noting that, SR is applied independently to each frame, thus, different frames can adapt to different numbers of the critical objects $C_t$, even if we keep $K_o$ constant for all frames.

\subsection{Multi-Granularity Reasoning (MGR)}
MGR aims to enhance the frame-level representation with fine-grained object embedding, while modeling the video dynamic together with question semantics.
%
%
%
As shown in \cref{w4-fig:2c}, MGR first applies intra-frame aggregation via a cross-attention, which takes the frame feature of the $t$-th critical frame $\tilde{\Mat{f}}_{t} \!\ \in\! \space{R}^{1 \times d}$ as query, and all critical objects in $t$-th frame $\tilde{\Mat{o}}_{t} \! \in\! \space{R}^{{C_{t}} \times d} $ as key and value to generate an object-enhanced representation $\mathring{\Mat{f}}_{t}\! \in \! \space{R}^{1 \times d}$ for the $t$-th frame:
\begin{gather} \label{w4-eq:mga}
    \mathring{\Mat{f}}_{t}=\text{Cross-Attention}(\tilde{\Mat{f}}_{t}, \tilde{\Mat{o}}_{t}) + \tilde{\Mat{f}}_{t}.
\end{gather}
By doing so to all $C$ critical frames, we acquire $\mathring{\Mat{F}} \! \in \space{R}^{C \times d}$ as the object-enhanced frame representation for all critical frames.
%
Next, a transformer layer is adopted to establish cross-frame dynamics, which takes in the concatenation of $\mathring{\Mat{F}}$ and question tokens $\Mat{Q}$, and yields multi-modal representations $\Mat{M} \!\in\!\space{R}^{(C+L) \times d}$ for answer decoding:
\begin{gather}
    \Mat{M}=\text{Transformer-Layer}([\mathring{\Mat{F}};\Mat{Q}]),
\end{gather}
where $[;]$ denotes concatenation operation.

\subsection{Answer Decoding}
Existing methods \cite{vgt,IGV,EIGV} concatenate a question with answer candidates. As analyzed in Sec.~\cref{w4-sec:intro}, these methods suffer from a spurious correlation between negative candidates and question-irrelevant video scenes. To circumvent this issue in spatio-temporal rationalization, we employ a transformer-style decoder that takes as input the question-critical multi-modal representations $\Mat{M}$ and the representations of the candidate answers to determine the correct answer. We detail our implementations for multi-choice QA and open-ended QA in the next sub-sections. 

\vspace{-7pt}
\subsubsection{Multi-Choice QA} 
In Multi-Choice QA, answer candidates are given as $\left| A_{mc} \right|$ sentences or short phrases that are tailored for each video-question pair. Therefore, reasoning on Multi-Choice QA typically requires fine-grain inspection of the video content as well as the interaction between the video and the candidate answers.
To this end, we first prepend a $\left[ \text{CLS} \right]$ token to each answer candidate and feed the sequences to the same language model used for question encoding. Then, we gather the output of the `[CLS]' tokens for all encoded answer candidates, and form the answer query $\Mat{U}_{mc} \in \Space{R}^{\left| A_{mc} \right| \times d}$.
During decoding, we feed a transformer decoder with $\Mat{U}_{mc}$ as query to interact with the multi-modal representation $\Mat{M}$, which yields the decoded representation $\Mat{H}_{mc} \in \Space{R}^{\left| A_{mc} \right| \times d}$ as:
\begin{gather} \label{w4-eq:decoder-mc}
    \Mat{H}_{mc} = \text{Transformer-Decoder}(\Mat{U}_{mc}, \Mat{M}).
\end{gather}
Notably, since the correctness of a answer candidate is invariant to its position, answer query is free of position encoding. 
Finally, we apply a linear projection on $\Mat{H}_{mc}$ to get the answer prediction $\hat{A}_{mc} \in \Space{R}^{\left| A_{mc} \right|}$,
\begin{gather}
    \hat{A}_{mc}=\text{Linear}(\Mat{H}_{mc}).
\end{gather}

\vspace{-15pt}
\subsubsection{Open-Ended QA}
Open-Ended setting provides $\left| A_{oe} \right|$ simple-form answer candidates (typically a single word) that are shared among all question instances, which makes the whole candidates set it too large to be processed as Multi-Choice setting. (\ie $\left| A_{oe} \right| \gg  \left| A_{mc} \right|$).
Instead, we take inspiration from DETR \cite{detr}, and initialize a single learnable embedding $\Mat{U}_{oe} \in \Space{R}^{d}$ as answer query.
Analogous to Multi-Choice setting, we feed $\Mat{U}_{oe}$ to the transformer decoder together with $\Mat{M}$ and acquire the decoded representation $\Mat{H}_{oe} \in \Space{R}^{d}$ similar to \cref{w4-eq:decoder-mc}.
As a result, we obtain the prediction $\hat{A}_{oe} \in \Space{R}^{\left|A_{oe}\right|}$ by projecting $\Mat{H}_{oe}$ to the answer space $\Space{R}^{ \left|A_{oe}\right|}$ via a linear layer:
\begin{gather}
    \vspace{-2pt}
    \hat{A}_{oe}=\text{Linear}(\Mat{H}_{oe}).
        \vspace{-2pt}
\end{gather}

During training, we establish our objective on a cross-entropy loss.  For inference, the differentiable Top-K is replaced with vanilla hard Top-K for better efficiency. 

%% file: work4/sec/4_exp.tex
\section{Experiments}\label{w4-sec:exp}
\input{work4/tab/dataset}

\noindent\textbf{Datasets:}
%
Experiments were conducted on four benchmarks to evaluate TranSTR's performance from different aspects. Specifically, the recent NExT-QA \cite{next-qa} and Causal-VidQA \cite{causalvid} datasets challenge models with complex VideoQA tasks using a Multi-Choice setting, which aims to test their temporal reasoning ability with complex causal and commonsense relations. In addition, MSVD-QA \cite{DBLP:conf/mm/XuZX0Z0Z17} and MSRVTT-QA \cite{DBLP:conf/mm/XuZX0Z0Z17} employ an Open-Ended setting and emphasize the description of video objects, activities, and their attributes. Their statistics are presented in \cref{w4-tab:dataset}.

\smallskip
\noindent\textbf{Implementation Details:}
%
Following the convention established in \cite{vgt}, we sample each video as a sequence of $T$=16 frames, where each frame is encoded by a ViT-L \cite{vit} model that pre-trained on ImageNet-21k. To extract object-level features, we employ a Faster-RCNN \cite{faster-rcnn} model that pre-trained on the Visual Genome and detect $S$=20 objects in each frame.
For the textual encoding, we adopt a pretrained Deberta-base model \cite{deberta} to encode the question and answer. During training, we optimize the model using an Adam optimizer with a learning rate of 1e-5, and set the hidden dimension $d$ to 768. For the hyper-parameters, we set $K_f$=5 and $K_o$=12 for all datasets.

Next, we show our experimental results to answer the following questions:
\begin{itemize}[leftmargin=*]
\setlength\itemsep{-.20em}
    \item \textbf{Q1:} How is TranSTR compared with the SoTA?
    \item \textbf{Q2:} How effective are the proposed components?
    \item \textbf{Q3:} What learning pattern does the rationlizer capture?
\end{itemize}

\subsection{Main Result (Q1)}
In \cref{w4-tab:main} and \cref{w4-tab:causal_vid}, we show that TranSTR outperforms SoTAs on all question types. Our observations are as follows:

\smallskip
\noindent \textbf{QA setting.}
Comparing the performance of TranSTR with state-of-the-art methods on four datasets, we observe that TranSTR achieves a greater improvement on Multi-Choice (NExT +5.8\% and Causal-Vid +6.8\%) compared to Open-End QA (MSVD +3.5\% and MSRVTT +3.4\%). This can be explained from two aspects:
(1) Unlike Open-Ended datasets that contain simple questions and short videos, Multi-Choice datasets (NExT and Causal-Vid) focus on complex VideoQA, in which composite question sentences with long videos and multiple objects (see \cref{w4-tab:dataset}) makes the identifying and inspecting of critical scenes necessary. This aligns with TranSTR's design philosophy of removing redundancy and explicitly exposing critical elements. Therefore, TranSTR achieves a larger gain on complex VideoQA.
(2) In multi-choice QA, SoTA methods often append candidate answers to the question during encoding, which can create a spurious correlation between the negative answer and the question-irrelevant scene, leading to a false prediction. However, such an issue is less significant in open-ended QA, where each answer candidate is treated as a one-hot category without semantic meaning. Thus, the decoder of TranSTR brings extra benefits to the multi-choice setting, and result in larger gains compared to open-ended QA.

\smallskip
\noindent \textbf{Question-type.}
Based on the analysis of Multi-Choice datasets, we observe that the improvement in overall performance of TranSTR is largely due to the enhancement in answering composite questions (including Acc@C and Acc@T in NExT-QA, Acc@E and Acc@P and Acc@C in Causal-VidQA) that require deeper understanding such as causal relations and counterfactual thinking, compared to the descriptive question type (Acc@D:+1.8$\sim$2.7\%). This demonstrates TranSTR's outstanding reasoning ability for complex VideoQA tasks.
In particular, for Causal-VidQA, TranSTR shows a significant improvement in answering reason-based questions (Acc@P:AR +10.5\%, Acc@C:AR +8.3\%). Because questions of this type require the model to justify its prediction by selecting the correct evidence, which aligns with the concept of rationalization in TranSTR's design philosophy. Therefore, TranSTR's rationalization mechanism enables it to perform optimally in answering reason-based questions.
\input{work4/tab/main}
\input{work4/tab/causalvid}

\subsection{In-Depth Study (Q2)}
\noindent \textbf{Ablative Results.} 
\input{work4/tab/ablation}
We validate the key components of TranSTR by performing model ablation and discussing other implementation alternatives.
As shown in \cref{w4-tab:ablation-loss}, we first study the effectiveness of TranSTR by removing both STR and decoder (``w/o STR \& decoder''), which induces a severe performance decline on every question type. 
Then, we conduct experiments to study STR. As a detailed breakdown test, we notice that reasoning with all frames without temporal rationalization (w/o ``TR'') will cause a performance drop.
A similar declination is also observed when spatial rationalization is erased (w/o ``SR''), that is, all objects on the selected frame are used for reasoning. 
Such performance drops are expected, because a large proportion of frames only contain a question-irrelevant scene, and the pretrained object detector will inevitably introduce noisy objects. These question-irrelevant contents, if not properly ruled out, will make the background overwhelm the causal information, due to its spurious correlation with the answer distractor. 
As a result, we witness a more significant performance drop when both temporal and spatial rationalization are removed (w/o ``STR'').
Next, we validate the effectiveness of our decoder design by adopting a conventional implementation that concatenates each answer candidate with the question before feeding it to the model. This variant, remarked as ``w/o decoder'', also caused a substantial performance drop, which highlights the importance of our video-text interaction pipeline in eliminating the background-distractor correlation.
Comparing the performance of (w/o ``STA'') and (w/o ``deocder'') to (``w/o STR \& decoder''), we show that removing both STR and decoder induce a more severe decline, which demonstrates that STR can coordinate well with the proposed decoder and their benefits are mutually reinforcing.
We also evaluate the importance of our MGR module by replacing it with average pooling. This variant, denoted as ``w/o MGR'', uses an average pooling to gather all objects on each frame and adds the pooled representation to the corresponding critical frames. We observed a significant performance drop when compared to the original TranSTR, which confirms the necessity of MGR in aggregating the multi-grain evidence for reasoning.
To validate that the STR indeed learns to focus on the critical elements instead of making random choices, we replace our differentiable top-K module, with a random K selection. As a result, the performance of ``Random K" drops drastically, which verifies the proposed STR is fully trainable to capture the answer information.
In addition, we also verify our choice of the differentiable module by replacing our perturbed maximum method with the ``SinkHorn Top-K'' \cite{sinkhorn}, and the results validate our implementation.

\smallskip
\noindent \textbf{Analysis on Complex VideoQA.}
\input{work4/tab/vt_obj}
\input{work4/tab/decoder}

In \cref{w4-tab:vt_obj }, we compare the results of TranSTR on the simple and complex VideoQA, where the test set of NExT-QA \cite{next-qa} is split by the length and object number of the source video, respectively. 
By calculating the accuracy within each group, we notice that all existing methods, as well as the TranSTR baseline (TranSTR without STR and proposed decoder), suffer from a performance drop when the video length exceeds 80 seconds (diff: -1.3\%$\sim$-2\%) or the video contains more than 5 objects (diff: -0.5\%$\sim$-1.9\%)). 
In contrast, TranSTR has alleviated this issue by explicitly ruling out redundant frames and noisy objects, thus resulting in even better performance on complex samples.

Similar to \cref{w4-fig:1b}, we also investigate how the TranSTR performs on samples with different video lengths. In \cref{w4-fig:vid_len}, we sort all test samples based on their video length and use a subset with the top percentage of longest videos to calculate accuracy. For example, 10\% on the x-axis denotes the accuracy of samples with the top 10\% longest videos, and 100\% denotes all samples considered.
Although the performance of TranSTR and baseline are initially comparable, the advantage of TranSTR becomes more pronounced as videos become longer. As a result, when the subset narrows down to the 10\% longest videos, we observe a difference in the accuracy of over 4\%.

\smallskip
\noindent \textbf{Study of Decoder.}
Our video-text interaction pipeline is able to cater to any SoTA methods without compromising their structure. Thus, we apply the proposed decoder to three VideoQA backbones by separating the answer candidates from the question and feeding them to our answer decoder. As shown in \cref{w4-tab:decoder}, our decoder is able to consistently improve the performance of all backbones, validating the assumption that isolating the answer candidates from the video-text encoding can eliminate the spurious correlation between a negative answer and background scenes, resulting in favorable gains for the backbone models.
Moreover, our decoder design is also more efficient compared to the conventional implementation. In the traditional approach, a question needs to be fused with the video multiple times, with each time concatenating a different answer candidate. In contrast, our design only forwards the question once, as the answer candidates are introduced only after the video-question fusion, resulting in a much more efficient architecture.

\begin{figure}
	\begin{minipage}{0.48\linewidth}
		\centering
        \includegraphics[width=0.8\textwidth]{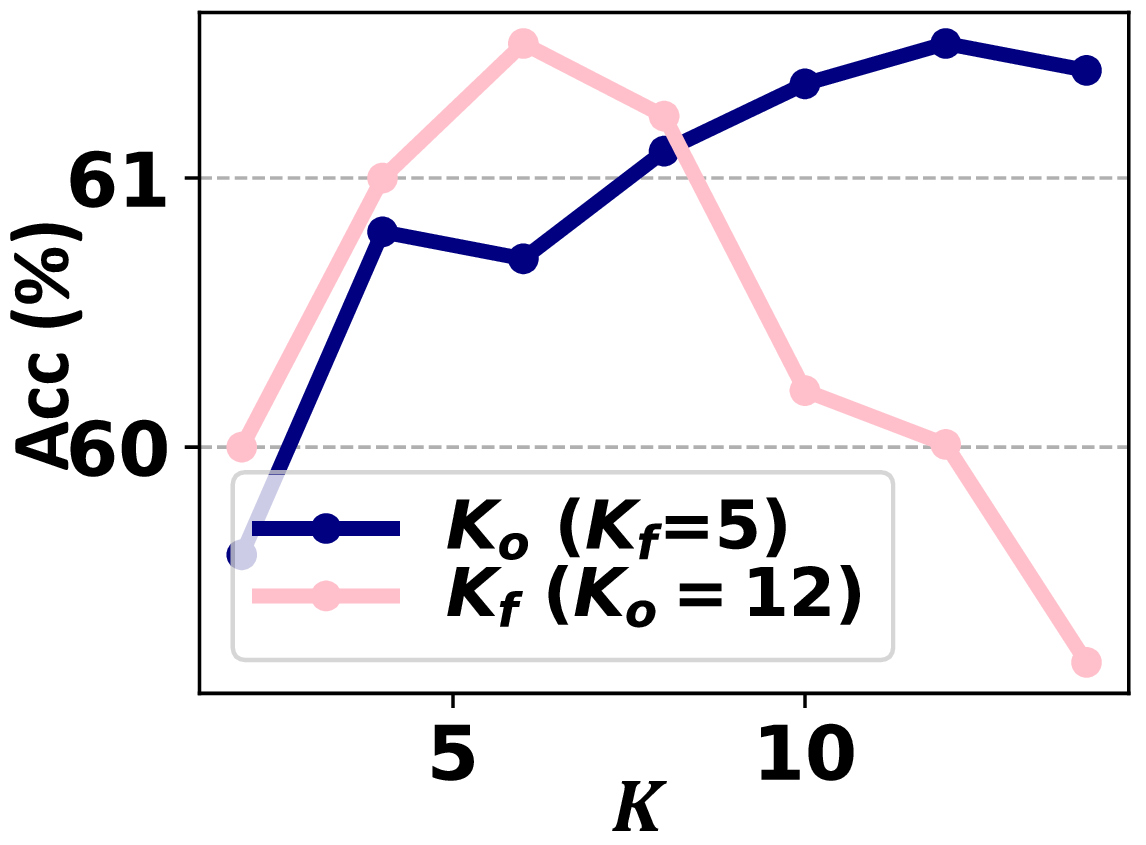}
        \captionof{figure}{Study of hyper-parameters.}
        \label{w4-fig:k}
	\end{minipage}\hfill
	\begin{minipage}{0.48\linewidth}
		\centering
        \includegraphics[width=0.8\textwidth]{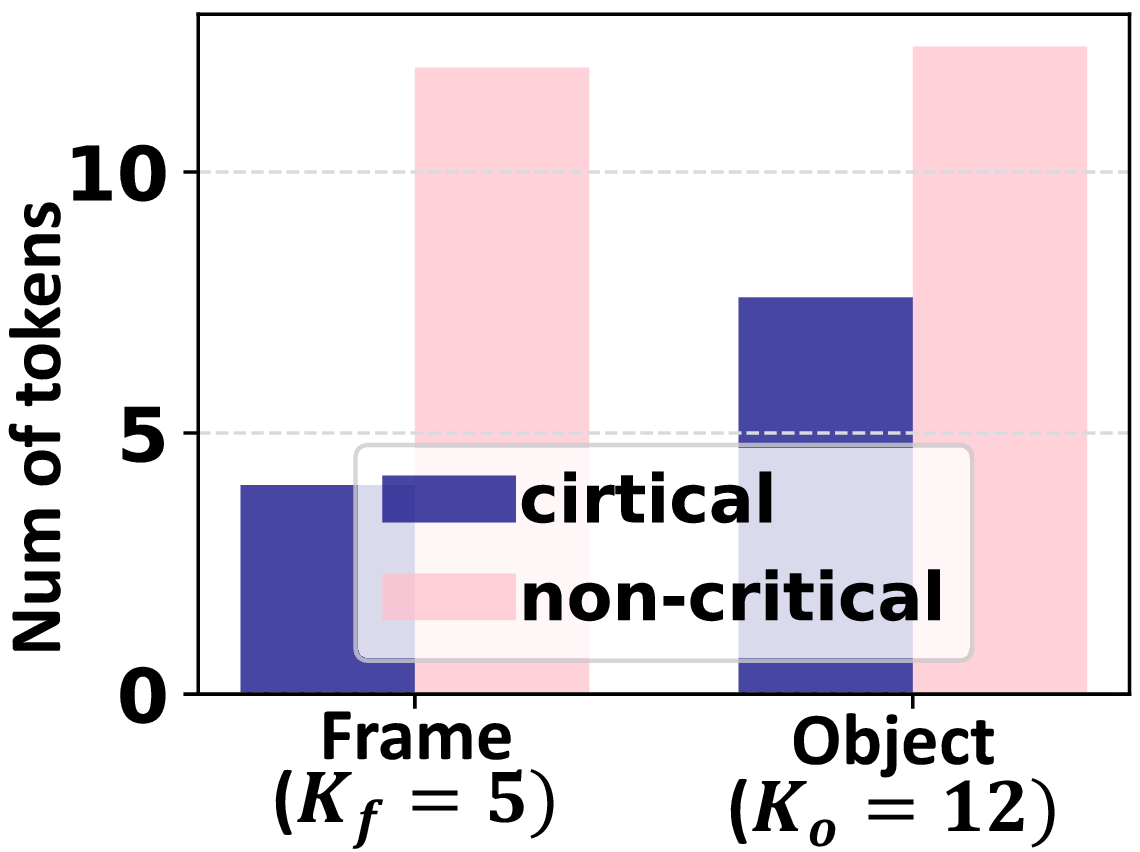}
        \captionof{figure}{Study of critical frames and objects.}            
	\label{w4-fig:c}
	\end{minipage}
\end{figure}

\begin{figure*}[t]
  \centering
\scalebox{1.0}{
  \includegraphics[width=1.0\textwidth]{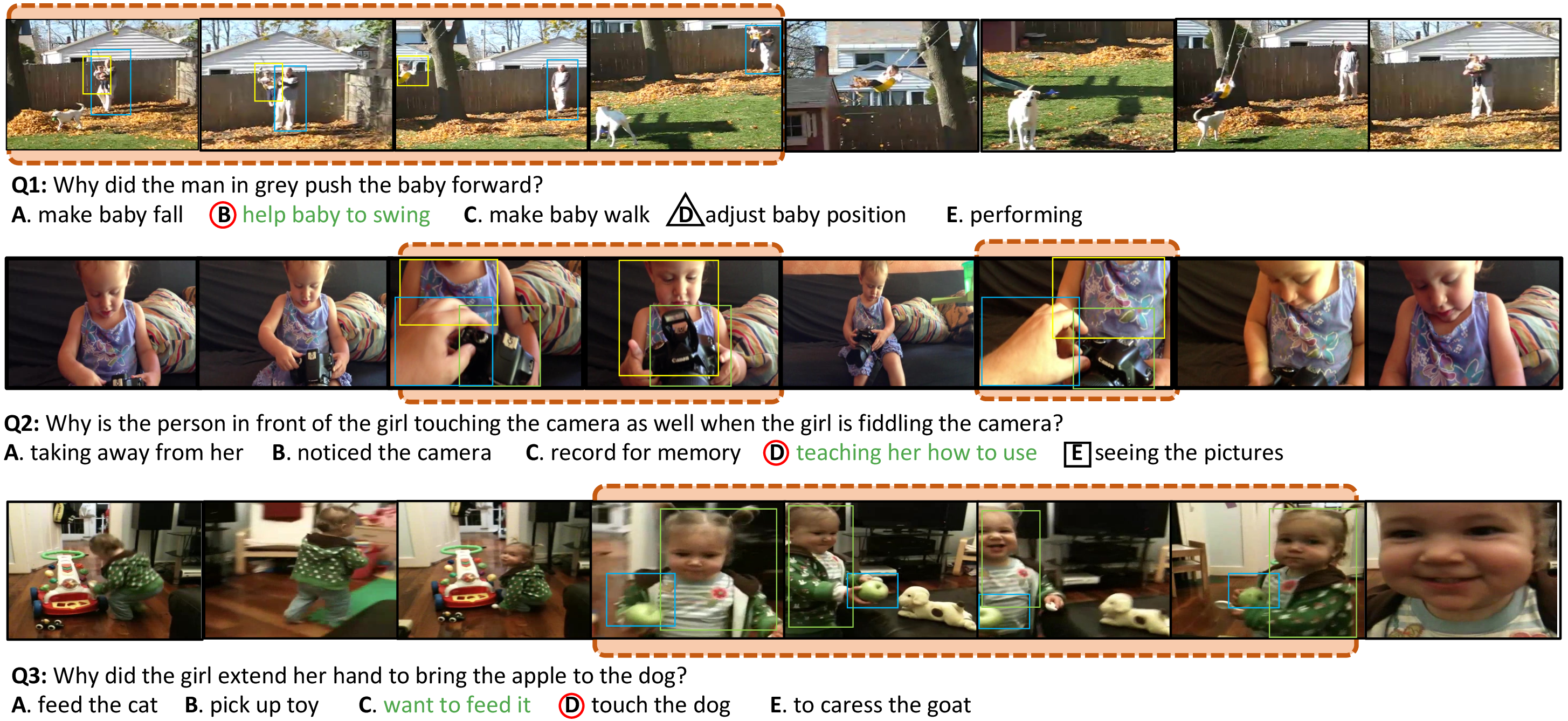}
  }
  	\vspace{-10pt}
  \caption{Case-Study on NExT-QA test set, the critical frames and objects are highlighted. Q1 and Q2 present the effect of the proposed STR and answer decoder, respectively. Q3 shows a failure case. The ground truth is colored in \green{green}. (
  \red{$\bigcirc$}:prediction of TranSTR,  $\bigtriangleup$: prediction of TranSTR w/o STR, $\Box$: prediction of TranSTR w/o decoder)}

  \label{w4-fig:case_study}
 	\vspace{-13pt}
\end{figure*}

\smallskip
\noindent \textbf{Study of Hyper-parameter.}
To validate the sensitivity of TranSTR to the number of collected interactions, we conduct experiments with variations of $K_f$ and $K_o$ on NExT-QA. Without loss of generality, we tune $K_f$ ($K_o$) while setting $K_o$ = 12 ($K_f$ = 5). According to \cref{w4-fig:k}, we observe that the performance of TranSTR varied mostly in the range of 61\% to 61.5\% under different combinations of hyperparameters, which demonstrated the effectiveness of TranSTR's adaptive design. However, we also notice a significant drop in some corner cases. When $K_f$ ($K_o$) is too small, the number of critical frames (objects) is limited, which hinders the performance as some visual evidence for answering is missing. Similarly, when $K_f$ is larger than 10, it introduces too much background, thus hurting the performance.

\subsection{Study of Critical Frames and Objects (Q3)}
\smallskip
\noindent \textbf{Quantitative study.} 
To grasp the learning insight of TranSTR, we inspect the number of frames and objects that are collected as critical by the adaptive rationalization. 
Concretely, we draw the number of frames $C$ and non-critical frames $T$-$C$ in \cref{w4-fig:k}. For the visual objects, since the number of critical objects $C_t$ varies according to frame content, we take the average of $C_t$ overall all critical frames, while leaving the rest objects as non-critical.
As a result, TranSTR can pinpoint a small group of tokens as critical tokens while leaving the rest as redundancy, which manifests the mass of question-irrelevant content in the original video, thus pointing the necessity of rationalization. 

\smallskip
\noindent \textbf{Qualitative study.}
To capture the learning pattern of TranSTR, we present some prediction results in \cref{w4-fig:case_study} along with the identified frames and objects.
In general, TranSTR can locate very few indicative elements as visual evidence. 
In question 1, we show the effectiveness of the STR. In temporal selection, it rules out the environment scene and targets the first four "swing" frames as critical. Next, in the spatial selection, it excludes non-causal objects (\ie "dog") and focuses on the relation between question-relevant objects (\ie "man" and "baby"). By aggregating the critical elements in frames and objects, TranSTR successfully reaches the gold answer. As a comparison, when the STR is removed, the massive background overwhelms the salient reasoning pattern and leads to a false prediction.
Question 2 demonstrates the effect of our answer decoder. We can see that TranSTR targets three critical frames that encompass the question-referred ``person", while selecting ``camera" and ``girl" as critical objects to correctly infer the person's intention. However, when the decoder is removed, the prediction falls into negative answer ``E.seeing the pictures". This is because implementation without our answer decoder inevitably suffers from a spurious correlation between the negative answer ``E.seeing the pictures" and frames where the girl is actually seeing pictures, even though these frames are irrelevant to the question.
Lastly, we present a failure case in question 3, where TranSTR fails to capture the subtle difference between ``feed'' and ``touch'', although the critical visual elements are located, which leads to a false prediction of the girl's intention.

%% file: work4/tab/dataset.tex
\setlength{\tabcolsep}{8pt}
\begin{table}[t!]
  \centering
  \small
  \caption{Dataset Statistics. MC and OE denote Multi-Choice and Open-Ended QA respectively.}
  \vspace{-0.1cm}
  \scalebox{0.95}{
    \begin{tabular}{lccccc}
    \toprule
    Dataset & Challenge & \#QA pair  & V-Len & Q-Len & QA \\
    \midrule
    NExT-QA & Causal \& Temporal & 48K   & 44s    &   11.6    & MC \\
    Causal-VidQA & Evidence \& Commonsense & 161K  & 9s     &   9.5    & MC \\
    MSVD-QA  & Description &   50K    & 10s    &    6.6   & OE \\
    MSRVTT-QA & Description & 244K  & 15s    &   7.4    & OE \\
    \bottomrule
    \end{tabular}
    }%
  \label{w4-tab:dataset}%
\end{table}%

%% file: work4/tab/main.tex
\setlength{\tabcolsep}{5pt}
\begin{table}[t!] 
  \centering
  \small
  \caption{Accuracy (\%) comparison on NExT-QA, MSVD-QA, and MSRVTT-QA. Acc@C, T, D, denote questions type of Causal, Temporal, and Descriptive in NExT-QA, respectively. The \textbf{best} and \underline{2nd best} results are highlighted.}
  \scalebox{1.0}{
    \begin{tabular}{l|c|ccc|c|c}
    \toprule
    \multirow{2}*{Methods} & 
    \multicolumn{4}{c|}{NExT-QA} & 
    \multirow{2}*{MSVD} & \multirow{2}*{MSRVTT} \\
    \cline{2-5} 
    ~  & Acc@All & Acc@C & Acc@T & Acc@D & ~ & ~ \\
    \midrule
    Co-Mem$\dagger$ \cite{gao2018motionappearance} & 48.5 & 45.9 & 50.0 & 54.4  & 34.6 & 35.3 \\
    HCRN \cite{hcrn}         & 48.9  & 47.1  & 49.3  & 54.0      & 36.1  & 35.6 \\
    HGA \cite{hga}         & 50.0 & 48.1  & 49.1  & 57.8      & 34.7  & 35.5 \\
    MSPAN \cite{mspan}           &  50.9  & 48.6    & 49.8     & 60.4       & 40.3  & 38.0 \\
    IGV \cite{IGV}   & 51.3 & 48.6  & 51.7  & 59.6   & 40.8 & 38.3 \\
    HQGA \cite{hqga}     & 51.8 & 49.0    & 52.3  & 59.4    & 41.2     & 38.6 \\
    EIGV \cite{EIGV}     & 52.9  & 51.2 & 51.5 & 61.0   & \underline{42.6}  & 39.3 \\
    VGT \cite{vgt}       & 53.7 & 51.6  & 51.9  & 63.7    & -     & \underline{39.7} \\

    VGT-PT \cite{vgt} &  \underline{55.7} & \underline{52.8}  & \underline{54.5}  & \underline{67.3}   & -     & - \\
    \midrule
    
    TranSTR  & \bf{61.5} & \bf{59.7}  & \bf{60.2}  & \bf{70.0}    & \bf{47.1}  & \bf{43.1} \\
    \textit{vs.} SoTA & +5.8 & +6.9 & +5.7 & +2.7  & +3.5 & +3.4 \\
    \bottomrule
    \end{tabular}}
  \label{w4-tab:main}%
\end{table}%

%% file: work4/tab/causalvid.tex
\setlength{\tabcolsep}{5pt}
\begin{table}[t!]
    \small
    \centering
    \caption{Accuracy (\%) comparison on Causal-VidQA. D: Description, E: Explanation, P: Prediction, C: Counterfactual. *: Reproduced result using official implementation.}
    \vspace{-0.5em}
    \scalebox{0.95}{
    \begin{tabular}{l|cccccccc|c}
    \toprule
    
    \multirow{2}*{Methods} & 
    \multirow{2}*{Acc@D} & 
    \multirow{2}*{Acc@E} & 
    \multicolumn{3}{c}{Acc@P} & \multicolumn{3}{c|}{Acc@C} & \multirow{2}*{Acc@All} \\
    \cline{4-9}
    ~ &~ & ~ & A & R & AR & A & R & AR &~ \\ 
    \midrule
    HCRN\cite{hcrn}  & 56.4 & 61.6 & 51.7 & 51.3 & \underline{32.6} & 51.6 & 53.4 & 32.7 & 48.1 \\
    HGA\cite{hga}  & 65.7 & 63.5 & 49.4 & 50.6 & 32.2 & 52.4 & 55.9 & 34.3 & 48.9 \\
    B2A\cite{park2021bridge}  & 66.2 & 62.9 & 49.0 & 50.2 & 31.2 & 53.3 & 56.3 & 35.2 & 49.1 \\
    VGT*\cite{vgt} & \underline{70.8} & \underline{70.3} & \underline{55.2} & \underline{56.9} & \underline{38.4} & \underline{61.0} & \underline{59.3} & \underline{42.0} & \underline{55.4}\\
    \midrule
    TranSTR & \bf{73.6} & \bf{75.8} & \bf{65.1} & \bf{65.0} & \bf{48.9} & \bf{68.6} & \bf{65.3} & \bf{50.3} & \bf{62.2}\\
     \textit{vs.} SoTA & +1.8 & +5.5 & +9.9 & +8.1 & +10.5 & +7.6 & +6.0 & +8.3 & +6.8 \\
    \bottomrule
    \end{tabular}
    }
    \label{w4-tab:causal_vid}
\end{table}

%% file: work4/tab/ablation.tex
\setlength{\tabcolsep}{12pt}
\begin{table}[t!]
  \centering
  \small
  \caption{Ablation Study}
    \scalebox{0.9}{
    \begin{tabular}{l|c|ccc}
    \toprule
    \multirow{2}*{Variants} & 
    \multicolumn{4}{c}{NExT-QA}\\
    \cline{2-5}
    ~  & Acc@All & Acc@C & Acc@T & Acc@D  \\
    \midrule
    TranSTR  & \bf{61.5} & \bf{59.7} & \bf{60.2} & \bf{70.0}  \\
    \midrule
    w/o STR \& decoder & 59.6 & 58.2  &  58.0  & 67.3  \\
    w/o STR & 60.3 & 59.1 & 58.3 & 67.9  \\
    $\;\;\;\;\;\;$ w/o TR & 60.8 & 59.4 &   58.6  &  69.5   \\
    $\;\;\;\;\;\;$ w/o SR & 60.7 & 59.6 & 58.2 & 69.0  \\
    w/o decoder & 60.1& 58.2   &   58.9   &  68.5  \\
    w/o MGR & 60.1 & 58.9 & 57.6 & 68.6  \\
    \midrule
    Random K & 54.6 & 53.6 &  51.6 & 64.0  \\
    SinkHorn Top-K & 61.0 & 59.4  &  59.7  &  68.7  \\


    \bottomrule
    \end{tabular}}%
  \label{w4-tab:ablation-loss}%
\end{table}%

%% file: work4/tab/vt_obj.tex
\setlength{\tabcolsep}{8pt}
\begin{table}[!t]
\small
\centering
\caption{Performance comparison, grouped by video length and object number. diff = Acc($\textgreater$ 80s) $-$ Acc( $\leq$80s)}.\label{w4-tab:vt_obj }
  \scalebox{0.95}{
\begin{tabular}{l|ccc|ccc|c}
\toprule
\multirow{2}{*}{Model} &\multicolumn{3}{c|}{Video Length} &\multicolumn{3}{c|}{Object Number} &\multirow{2}{*}{Total} \\
\cline{2-7}
& $\leq$ 80s & $\textgreater$ 80s & diff($\uparrow$) & $\leq$ 5 & $\textgreater$ 5 & diff($\uparrow$) & \\
\midrule
EIGV \cite{EIGV} &53.3 &51.3 &-2 &53.3 &52.1 &-1.2 &52.9 \\
VGT \cite{vgt} &54.4 &52.2 &-2.2 &54.7 &52.8 &-1.9 &53.7 \\
VGT-PT \cite{vgt}  &55.8 &54.5 &-1.3 &56.4 &54.9 &-1.5 &55.7 \\
\midrule
Baseline &59.8 &58.7 &-1.1 &60.0 &59.2 &-0.8 & 59.6 \\
TranSTR &61.4 &62.4 & \bf{+1} &61.2 &61.9 & \bf{+0.7} &61.5 \\
\bottomrule
\end{tabular}}
\end{table}

%% file: work4/tab/decoder.tex
          



\begin{figure}
	\begin{minipage}{0.45\linewidth}
		\centering
        \includegraphics[width=0.97\textwidth]{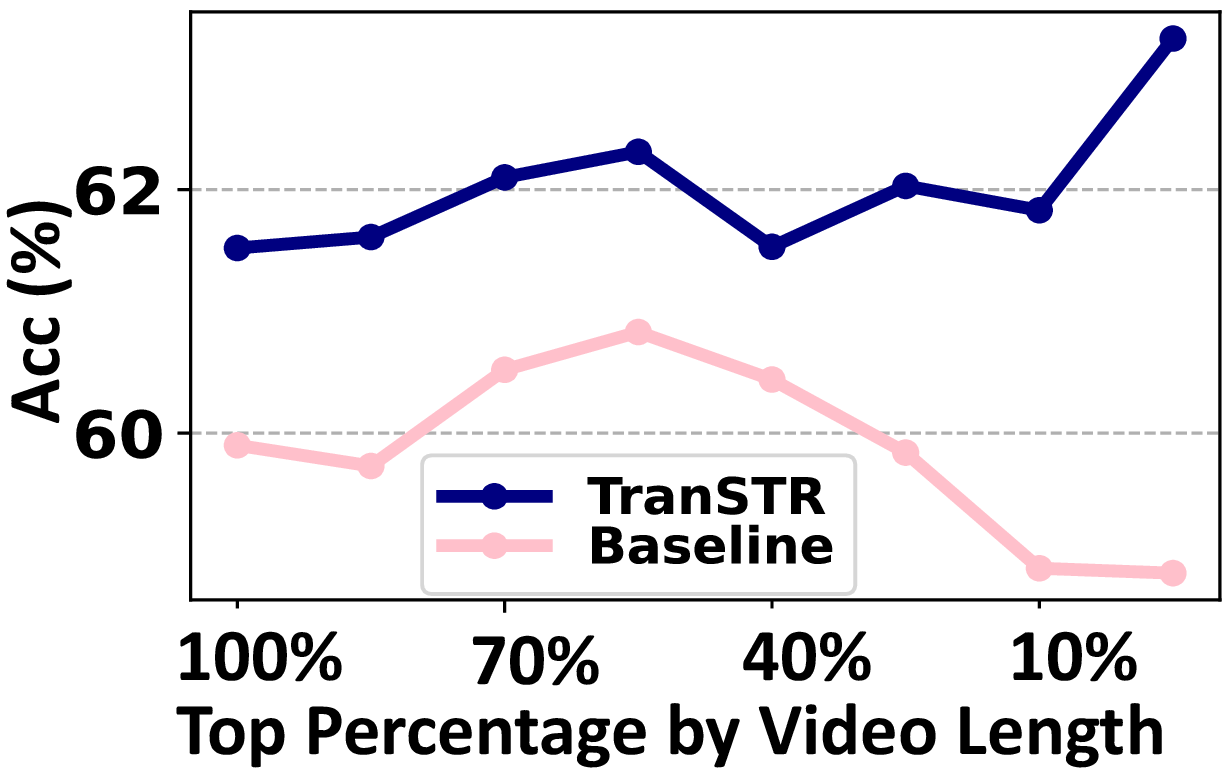} 
        \captionof{figure}{Acc by video length}
        \label{w4-fig:vid_len}
	\end{minipage}\hfill
	\begin{minipage}{0.4\linewidth}

		\centering
            \small
		\resizebox{0.95\textwidth}{!}{%
            
		    \begin{tabular}{l|cc}
        \toprule
        \multirow{2}*{Methods} & 
        \multicolumn{2}{c}{Decoder} \\
        ~ & w/o & w \\
        \midrule
        MSPAN \cite{mspan} &50.9 &52.7 \\
        EIGV \cite{EIGV} &51.3 &53.3 \\
        TranSTR &60.1 &61.5 \\
        \bottomrule
    \end{tabular}
                }
                \captionof{table}{Apply our decoder to SoTAs.}
                \label{w4-tab:decoder}
	\end{minipage}
\end{figure}




%% file: work4/sec/5_conclusion.tex
\section{Conclusion}
\vspace{-3pt}
For the first time, this chapter addresses complex VideoQA, where long multi-object video and hard answer distractors have crippled existing methods, owing to their incapacity in handling massive visual backgrounds and modeling hard distractor answers. We then proposed STR to adaptively trim off question-irrelevant scenes, and further develop a novel answer decoding scheme that coordinates with STR to overcome the spurious correlation resulted from distractor-background interaction.
Instantiating this pipeline with transformer architecture, we showed that our method, TranSTR, could achieve significant improvements over SoTAs, especially on complex VideoQA tasks. We hope our success can shed light on answering questions in the context of long videos with multiple objects.


%% file: conclusion.tex
\chapter{Conclusion and Future Work}
\label{cha:conclusion}
\section{Conclusion}
In this thesis, we have explored the domain of causal modeling in semantic video understanding by investigating casual patterns in two tasks, VideoQA and VidVRD. 
Following a shallow-to-deep manner, we began by studying the long-tail imbalance in VidVRD, where rare but informative relations were difficult to predict due to their scarcity in the dataset. To overcome this, we proposed Interventional Video Relation Detection (IVRD), a novel approach that leverages causal reasoning to discover the causal patterns of tail relations. By forming relation prototypes and focusing on the visual content of dynamic interactions between entities, IVRD significantly improved the detection of tail relation.

Then we moved beyond pair-wise relation and focused on the overall understanding of a video by answering a content-related question, \ie VideoQA. Specifically, we aim to mitigate the perturbed imbalance induced by the environmental scene, which often led to spurious correlations between the visual content and the answer. To tackle that, we introduced Invariant Grounding for VideoQA (IGV), a model-agnostic learning framework that identified the causal reasoning pattern by grounding the question-critical scene. IGV effectively shielded the answering process from the negative influence of spurious correlations, resulting in enhanced reasoning ability for several VideoQA backbones. On top of this, we introduced Equivariant Grounding for VideoQA (EIGV), which further improved robustness and visual explainability by incorporating equivariance, making the answering process sensitive to semantic changes in the causal scene and question.

Delving into the object-level causal pattern, we explored Spatio-Temporal Rationalization (STR) to tackle the challenge of low accuracy in complex VideoQA tasks, where videos were longer and involved multiple objects. By adaptively gathering question-critical moments and objects through cross-modal interaction, STR effectively identified question-irrelevant frames and objects as causal patterns, leading to improved predictions, especially in complex scenarios.

In conclusion, the key features of the VideoQA models discussed in this thesis are primarily designed to address specific challenges. The IGV and EIGV models are tailored to tackle the Out-Of-Distribution (OOD) problem, wherein the background scenes encountered are seldom observed during training. On the other hand, the TranSTR model is specifically designed for complex VideoQA tasks involving long videos and multiple objects interacting over time.

Throughout this thesis, we have demonstrated the effectiveness and importance of causal modeling in video understanding tasks. By leveraging causal reasoning, we have addressed long-tail and perturb imbalance in VidVRD and VideoQA, leading to more informative, accurate, and robust models. The proposed frameworks, IVRD, IGV, EIGV, and STR, have all contributed to advancing the field of semantic video understanding and providing valuable insights into the importance of causal reasoning. As video understanding continues to evolve, we believe that causal modeling will play a crucial role in pushing the boundaries of performance and generalization in this domain.

\section{Future Work}
Looking toward the future, there are several promising directions for further research and development in the domain of causal modeling for semantic video understanding, which includes its evaluation and extension to real-world applications and complex scenarios.

\begin{figure}[t]
\centering
\includegraphics[scale=0.7]{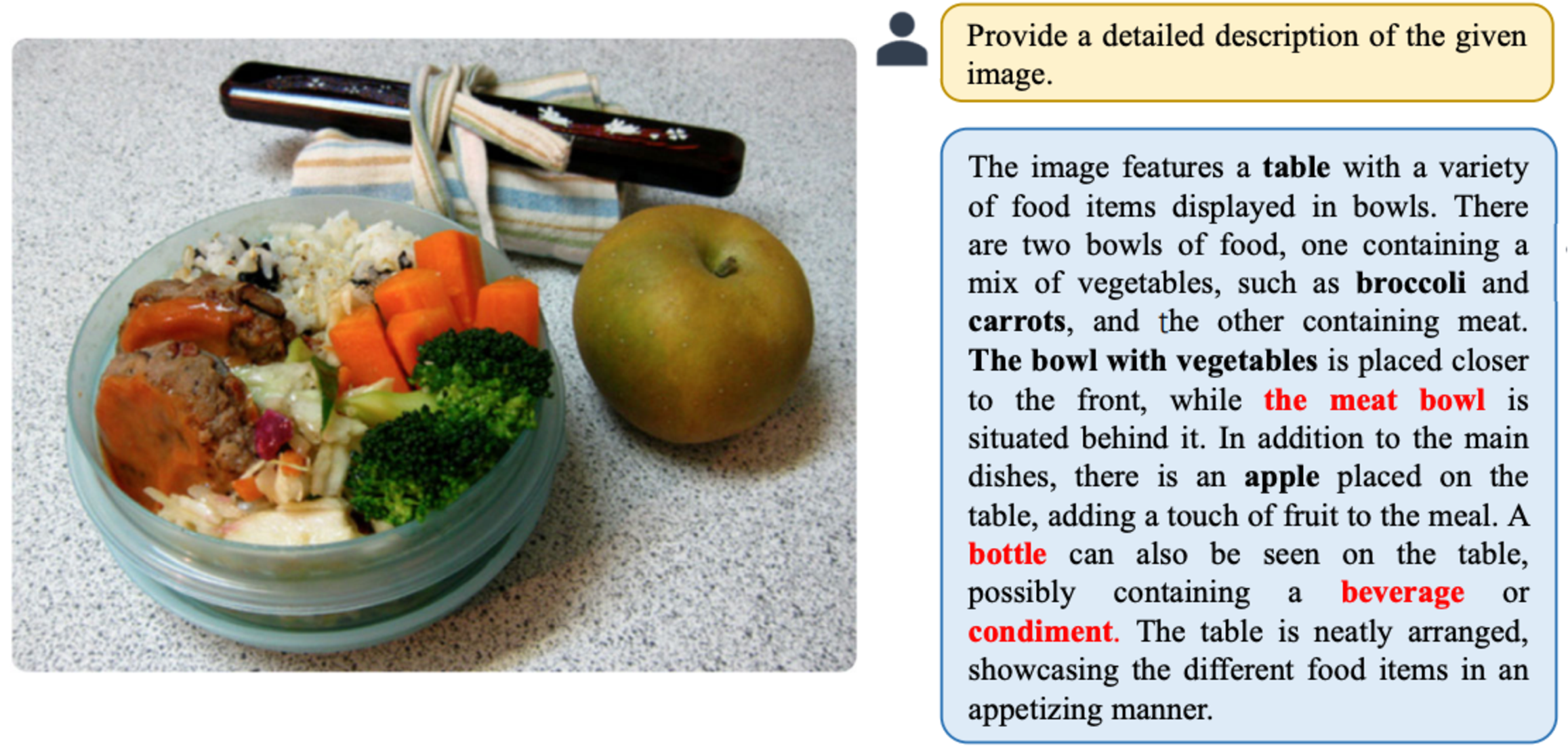}
\caption{Example of VLM hallucination discovered by POPE\cite{POPE}. Bold objects are ground-truth objects in the annotations and red objects are hallucinated objects.}
\label{rebuttal:hallucination}
\end{figure}

\smallskip
\noindent \textbf{Hallucibation in Vison-Language foundation model.}
Previous work has revealed that vision-language foundation models (VLM) suffer from hallucinations. In \cref{rebuttal:hallucination}, when asked to describe the image, VLM tends to generate
``bottle" and ``beverage" that do not exist in the image.  This phenomenon can be attributed to the frequent co-occurrence of "beverage" with "knife" in the training data, typically captured in dining scenes. Thus, upon detecting a "knife" in the scene, VLM may erroneously hallucinate the presence of a "beverage". 

In fact, this parallels the challenge we highlighted regarding "race on track" in \cref{cha:igv}. While the methodology introduced in \cref{cha:igv} could theoretically address this issue for VLM, the computational overhead for such practice is intolerable. Thus, my future work also includes how to alleviate the hallucination of VLM with a more efficient causal method.

\smallskip
%
\noindent \textbf{Grounded VideoQA.}
In order to advance the discovery and evaluation of causal patterns in the VideoQA domain, constructing a benchmark with human annotations on causal scenes is an essential step that I will look into. A good direction is to construct a benchmark that addresses the research question: "What part of the video is used by the current VideoQA model to answer the question" To achieve this goal, we propose an approach called visually grounded VideoQA, wherein the VideoQA models not only provide answers to questions but also produce relevant video moments that support their answers. This approach aims to enhance the interpretability and reliability of predictions by explicitly connecting the answers to specific video segments.
%
%

\noindent \textbf{3D Question Answering}
Moreover, I also see significant potential in exploring 3D Question Answering (3D-QA), which serves as a natural extension of video understanding and holds immense practical applications, especially in the context of home robots and other physical systems. In contrast to traditional video understanding, where the focus is on processing 2D visual data, 3D-QA enables robots to interact with the physical world in a more comprehensive manner, considering depth, spatial relations, and object interactions. However, this increased complexity also introduces new challenges in terms of understanding causal relationships accurately. Given that robots must make decisions and take actions based on the information gathered from the 3D environment, any responses that rely on spurious correlations can lead to detrimental consequences and deviate from their actions in the real world. Therefore, integrating causal modeling techniques into 3D-QA becomes essential to ensure that robots possess the necessary reasoning abilities to make informed and reliable decisions, thereby enhancing their safety, performance, and overall effectiveness in real-world interactions. By exploring causal patterns in 3D scenarios, we can equip robots with the capability to distinguish meaningful cause-and-effect relationships, leading to more responsible and efficient interactions in the physical environment.

%% file: work2/sec/X_supplementary.tex
\chapter{Appendix to Chapter~\ref{cha:igv}}
\label{app:igv}

\section{Example of context type}
\label{w2-app:critical-context-example}
\begin{figure}[t]
    \centering
    \includegraphics[width=0.96\linewidth]{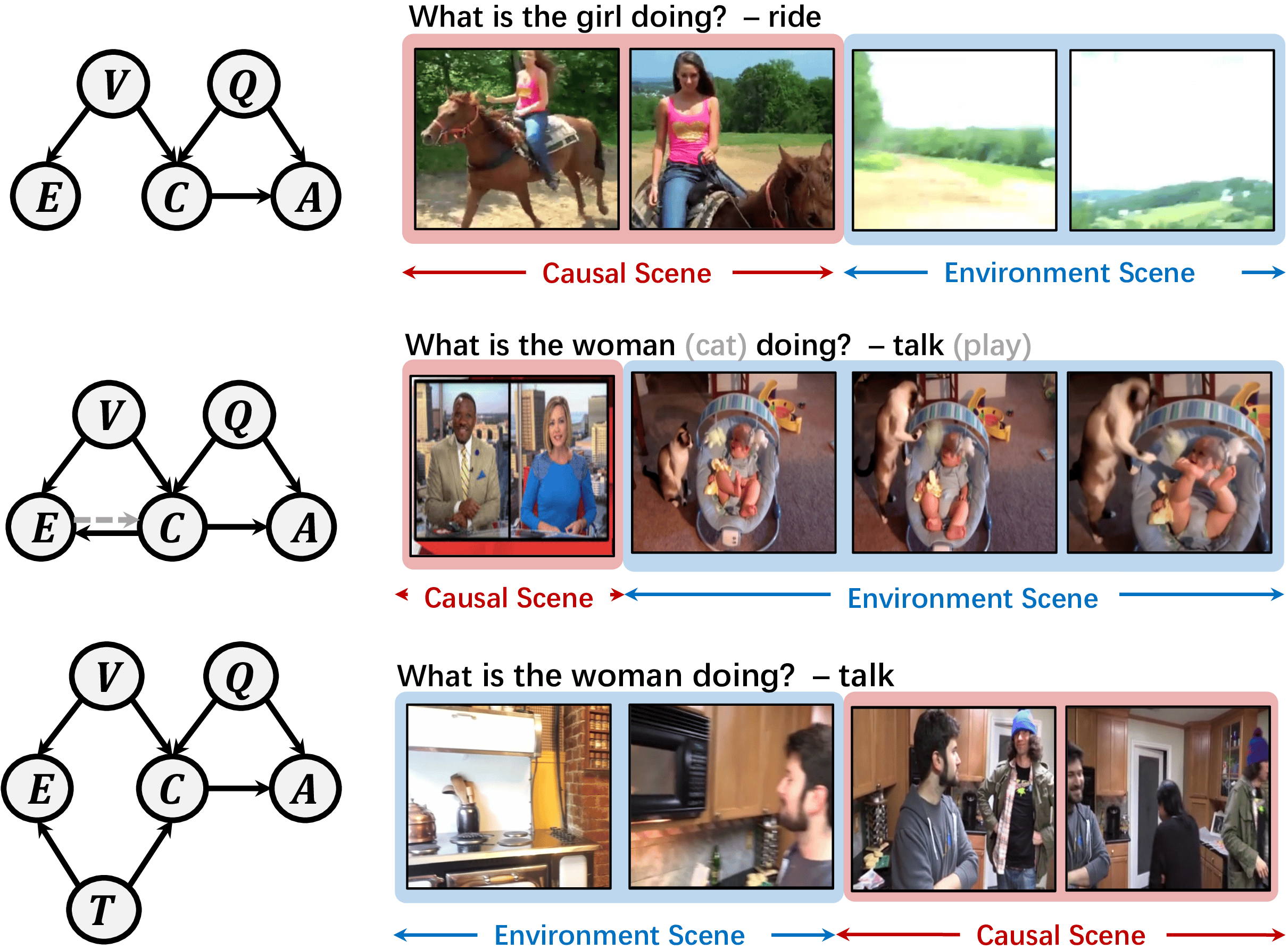}
    \caption{Illustration of environment type.}
    \label{w2-fig:complement type}
\end{figure}

As shown in Figure \cref{w2-fig:complement type}, we classify the relation between causal scene and its environment (\eg $T\dashleftarrow\dashrightarrow C$) into three types, where each row encompasses a causal graph (left) that depicts typical causal-environment relation demonstrated in the example (right):
\begin{itemize}[leftmargin=*]
\setlength\itemsep{-.20em}
    \item In the first row, $C$ and $T$ has no causal relation (\ie $T \bot C$). 
    \item The second row shows a scenario that $C$ is the direct cause of $T$ (\ie $C\rightarrow T$), or vise versa if the question is modified (\eg 'What is the cat doing?')  
    \item Similar to the example in Figure \cref{w2-fig:intro-example}, the third row demonstrates how shortcut deviate the prediction from the gold answer (\eg "talk") to false prediction (\eg "cook") via common cause $E$ (\eg visual concept "kitchen") since LMI between visual concept "kitchen" and candidate answer "cook" is much higher than it is with "talk".  
\end{itemize}

\section{Our backbone}
\label{w2-app:backbone}
Most VideoQA architectures from the state of the art are compatible with our IGV learning strategy. To testify, we design a simple and effective architecture inspired by \cite{jiang2020reasoning}. Specifically, $f_{\hat{A}}$ is presented as a combination of a visual-question mixer and an answer classifier. The mixer first encode $\hat{c}$:
\begin{gather}
    \Mat{v}_{g}^{\hat{c}}, \Mat{v}_{l}^{\hat{c}}=\text{LSTM}_{5}(\hat{c})
\end{gather}
where outputs $\Mat{v}_{g}^{\hat{c}} \in \mathbb{R}^{d}$, $\Mat{v}_{l}^{\hat{c}} \in \mathbb{R}^{N \times d}$ denote the global and local feature of $\hat{c}$ respectively. Then, based on the concatenation of local representation $\Mat{q}_{l}$ (\cf Equation \cref{w2-equ:vl-lstm}) and $\Mat{v}_{l}^{\hat{c}}$ , we construct an undirected heterogeneous graph that propagates information over each video shot and each question token. Typically, the adjacency matrix $\mathcal{G}_{\hat{c}} \in \mathbb{R}^{(L+N)\times (L+N)}$ is computed as the node-wise correlation scores in form of dot-product similarity, where $N\le K$ is the sequence length of casual scene. The output of the graph is assembled as holistic local factor $\Mat{s}^{\hat{c}}_l \in \mathbb{R}^{d}$ via a attention pooling operator.
More Formally, the process is as follows: 
\begin{gather}
    \Mat{x}_{\hat{c}}=[\Mat{v}_{l}^{\hat{c}};\Mat{q}_{l}], \;\; \mathcal{G}_{\hat{c}}=\sigma(\text{MLP}_{5}(\Mat{x}_{\hat{c}}))\cdot\Trans{\sigma(\text{MLP}_{6}(\Mat{x}_{\hat{c}}))}
\end{gather}
\begin{gather}
    \Mat{z}_{\hat{c}}=\text{GCN}(\Mat{x}_{\hat{c}}, \mathcal{G}_{\hat{c}})
\end{gather}
\begin{gather}
    \Mat{s}_{\hat{c}}^l=\text{Pooling}(\Mat{z}_{\hat{c}})
\end{gather}
where $\Mat{x}_{\hat{c}}$, $\Mat{z}_{\hat{c}}= \in \mathbb{R}^{(L+N)\times d}$ 
denote the input and output of graph reasoning, $\text{MLP}_5$ and $\text{MLP}_6$ denote is affine projection followed by ReLU activation $\sigma(\cdot )$. 
To capture the global information, our mixer integrates two global factors $\Mat{v}_{g}^{\hat{c}}$ and $\Mat{q}_{g}$ into holistic representation via BLOCK fusion \cite{BenYounes_2019_AAAI}:
\begin{gather}
    \Mat{s}_{\hat{c}}^g=\text{Block}(\Mat{v}_{g}^{\hat{c}},\Mat{q}_{g})
\end{gather}
Similarly, we obtain the final representation by applying the BLOCK again to global and local factor, which is further decoded into answer space with classifier $\Psi$:
\begin{gather}
    \Mat{s}_{\hat{c}}=\text{Block}(\Mat{s}_{\hat{c}}^g,\Mat{s}_{\hat{c}}^l)
\end{gather}
\begin{equation}
    \hat{y}_{\hat{c}}=\Psi(\Mat{s}_{\hat{c}})
\end{equation}
Analogously, we can obtain the predictive answer for $\hat{t}$ and $v^{*}$ via the shared backbone predictor. 

\section{Baselines}
\label{w2-app:baseline}
We compare our design against some existing work, which can be categorized into three categories: 
1) \textbf{Memory-based} methods that perform multi-step reasoning via updating the recurrent unit,  which refines the cross-modal representation iteratively. Specifically, AMU \cite{DBLP:conf/mm/XuZX0Z0Z17}, Co-Mem\cite{gao2018motionappearance} apply this module to encode the  visual representation, and HME \cite{fan2019heterogeneous} managed better exploitation for both modalities; 
2) \textbf{Graph-based} methods like HGA \cite{jiang2020reasoning} and B2A \cite{ park2021bridge} adopt graph reasoning on the clip-level, whose adjacent matrix is built on node-wise visual similarity. Comparatively, B2A additionally establishes a text graph through question parsing, and abridge two modalities via message passing;
3) \textbf{Hierarchical-based} methods HOSTR \cite{dang2021hierarchical} and HCRN \cite{le2021hierarchical} have similar hierarchical conditional architectures. Their discrepancy lies in the feature granularity, where HCRN grounds the temporal relation between frames, while HOSTR roots in object trajectories.

\section{Implementation details}
\label{w2-app:implementation}
All experiments are conducted on GPU NVIDIA Tesla V100 installed on Ubuntu 18.0.4. 
In terms of complexity, our algorithm matched equally with the corresponding baseline.  As a comparison, the default backbone model is trained for 2 hours till convergence on MSRVTT-QA, whereas IGV
takes 2.6 hours. For space complexity, since we use the same predictor for the causal, environment, and intervened prediction, IGV only takes 10\% more parameters than the default backbone model. 
